\documentclass{article}

\usepackage[final]{cpal_2024}

\usepackage[utf8]{inputenc} 
\usepackage[T1]{fontenc}    
\usepackage{hyperref}       
\usepackage{url}            
\usepackage{booktabs}       
\usepackage{amsfonts}       
\usepackage{amsthm}         
\usepackage{amsmath}
\usepackage{nicefrac}       
\usepackage{microtype}      
\usepackage{xcolor}         
\usepackage{tcolorbox}
\usepackage{multirow}
\usepackage{algorithm}
\usepackage{enumitem}
\usepackage{xspace}
\usepackage{wrapfig}
\usepackage{graphicx}
\usepackage{caption,subcaption}
\usepackage[export]{adjustbox}

\DeclareMathOperator*{\argmin}{argmin}
\newcommand{\our}{\texttt{UGoDIT}\xspace}

\title{\our: Unsupervised Group Deep Image Prior Via Transferable Weights}

\author{Shijun Liang\textsuperscript{1}
\thanks{The first two authors contributed equally.}~, ~  Ismail R. Alkhouri\textsuperscript{1,2}, ~Siddhant Gautam
\textsuperscript{1}, \\ ~Qing Qu\textsuperscript{2}, ~Saiprasad Ravishankar\textsuperscript{1} \\ 
  \textsuperscript{1}Michigan State University\\
  \textsuperscript{2}University of Michigan \\
}

\begin{document}

\maketitle

\begin{abstract}

Recent advances in data-centric deep generative models have led to significant progress in solving inverse imaging problems. However, these models (e.g., diffusion models (DMs)) typically require large amounts of fully sampled (clean) training data, which is often impractical in medical and scientific settings such as dynamic imaging.
On the other hand, training-data-free approaches like the Deep Image Prior (DIP) do not require clean ground-truth images but suffer from noise overfitting and can be computationally expensive as the network parameters need to be optimized for each measurement set independently. Moreover, DIP-based methods often overlook the potential of learning a prior using a small number of sub-sampled measurements (or degraded images) available during training. In this paper, we propose \our, an \textbf{U}nsupervised \textbf{G}r\textbf{o}up \textbf{DI}P via \textbf{T}ransferable weights, designed for the low-data regime where only a very small number, $M$, of sub-sampled measurement vectors are available during training. Our method learns a set of transferable weights by optimizing a shared encoder and $M$ disentangled decoders. At test time, we reconstruct the unseen degraded image using a DIP network, where part of the parameters are fixed to the learned weights, while the remaining are optimized to enforce measurement consistency. We evaluate \our on both medical (multi-coil MRI) and natural (super resolution and non-linear deblurring) image recovery tasks under various settings. Compared to recent standalone DIP methods, \our provides accelerated convergence and notable improvement in reconstruction quality. Furthermore, our method achieves performance competitive with SOTA DM-based and supervised approaches, despite not requiring large amounts of clean training data. 
\end{abstract}


\section{Introduction} \label{sec: intro}

Solving inverse problems are important across a wide range of practical applications, such as those in robotics \cite{manocha2002efficient}, forensics \cite{ren2018deep}, remote sensing \cite{levis2022gravitationally}, and geophysics \cite{bnilam2020parameter}. One class of such problems is Inverse Imaging Problems (IIPs), including super-resolution \cite{ledig2017photo}, image denoising~\cite{GOYAL2020220}, image deblurring~\cite{tran2021explore}, recovering missing portions of images~\cite{inpainting2022} (in-painting), and magnetic resonance imaging (MRI) reconstruction~\cite{5484183}. 

In recent years, an abundance of deep learning (DL) methods have been developed to solve IIPs \cite{mccann2017convolutional,alkhouri2024diffusion, ye2024decomposed, chung2022score, alkhouri2024sitcom}. Based on the availability of training data points, DL-based IIP methods can be categorized into (\textit{i}) data-less DL approaches, such as the ones based on the Deep Image Prior (DIP) \cite{ulyanov2018deep} and Implicit Neural Representation (INR) \cite{saragadam2023wire}, and (\textit{ii}) data-centric DL approaches, such as the ones based on supervised networks \cite{sriram2020end,aggarwal2018modl}, 
or generative models such as those based on diffusion models (DMs) \cite{chung2022score, alkhouri2024sitcom}. 

Data-centric methods rely on pre-trained models that typically require large-scale training datasets. In the context of scientific IIPs or medical IIPs such as MRI, in addition to being large-scale, this translates to the need for fully sampled measurements—i.e., clean images that serve as ground truth. Unlike natural images, however, acquiring large datasets of scientific or medical data such as fully sampled MRI scans is particularly challenging~\cite{garwood2017advanced, zbontar2018fastmri, chaudhari2020rapid}.
The underlying data may be inherently undersampled or corrupted with unknown noise.
Similar issues arise in other domains such as medical computed tomography (CT) \cite{shafique2024mri}. To mitigate this, previous DM-based methods for MRI and CT often fine-tune DMs pre-trained on natural images using a limited number of fully sampled medical images~\cite{chung2022score, ye2024decomposed, alkhouri2024sitcom}. This underscores the need for approaches that reduce reliance on large, clean (or fully sampled) datasets.

Training-data-free methods, such as DIP \cite{ulyanov2018deep}, which rely on untrained networks (e.g., U-Net \cite{ronneberger2015u}), avoid the need for training data altogether—fully sampled or otherwise. However, they often face two main challenges: (\textit{i}) vulnerability to noise overfitting (though recent variants, such as auto-encoding sequential DIP (aSeqDIP) \cite{aSeqDIP}, have shown improved robustness), and (\textit{ii}) high computational cost at inference time (as a full set of network parameters need to be optimized for each measurement independently), especially compared to the inference time of supervised methods. 
Although recent work, such as MetaDIP \cite{zhang2022metadipacceleratingdeepimage}, has explored transferring learned DIP weights to speed up inference, it still requires supervised training with fully sampled data points, and does not achieve competitive results when compared to standalone DIP methods as was demonstrated in \cite{li2023deep}.

The scarcity of fully-sampled measurements (i.e., clean images as ground truth) in certain applications, and the inability of most DIP methods to transfer learned weights serve as our motivation to explore the following question: \textcolor{black}{\textit{Is it possible to train a fully unsupervised model using only a few degraded images, such that the learned weights can later be used to reconstruct unseen measurements during inference?} }
 We term the regime when only very few training images are available as the ``low-data regime''
, and towards answering the above question, we make the following contributions:

\begin{itemize}[leftmargin=*]
    \item \textbf{Decoder-disentangled architecture for training}: Given a few measurement vectors (or degraded images), we propose the use of a shared encoder and multiple decoders for each measurement vector during training such that the inputs, shared encoder, and multiple decoders are optimized jointly using a measurement-consistent and auto-encoded objective function. 

    \item \textbf{Transferable weights at testing}: Given an unseen measurement vector, we transfer the pre-trained weights of the shared encoder to a single-encoder, single-decoder DIP network. Reconstruction is then performed by fixing the encoder and optimizing the decoder parameters using an input-adaptive, measurement-consistent objective function. Therefore, we term our method as \textbf{U}nsupervised \textbf{G}r\textbf{o}up \textbf{DI}P with \textbf{T}ransferable weights (\our). 

    \item \textbf{Extensive Evaluation}: We conduct extensive evaluations using multi-coil MRI, super resolution, and non-linear deblurring. We show that we outperform most recent advanced DIP methods while performing comparably with DM-based and supervised data-intensive methods all while requiring only a very few degraded images. Furthermore, our ablation studies show the impact of each component such as the proposed disentangled decoders. 
\end{itemize}

\section{Preliminaries \& related work} \label{sec: prelim and related}

Inverse Imaging Problems (IIPs) are defined as recovering an image $\mathbf{x}^* \in \mathbb{R}^n$ from measurement vector (or degraded image) $\mathbf{y}\in \mathbb{R}^m$ (with $m\leq n$), governed by the forward operator $\mathbf{A}$. For multi-coil MRI, the forward operator is given as $\mathbf{A} = \mathbf{M} \mathbf{F} \mathbf{S}$, for which $\mathbf{M}$ denotes coil-wise undersampling, $\mathbf{F}$ is the coil-by-coil Fourier transform, and $\mathbf{S}$ represents the sensitivity encoding with multiple coils\footnote{In MRI, the entries of $\mathbf{x}^*$ and $\mathbf{y}$, contain both real and imaginary numbers. However, to generalize the definition of the problem, we use real numbers.}. For natural image restoration tasks, the forward operator in super resolution (SR) is a down sampler. 
For non-linear deblurring (NDB), the non-linear forward operator is the neural network approximated kernel given in \cite{tran2021explore} and adopted in multiple works such as \cite{alkhouri2024sitcom,chung2022diffusion}. Next, we provide a brief overview of prior DL-based image reconstruction methods based on the availability of training data. 

\paragraph{High-data regime methods:} Training data-intensive methods—that is, methods that rely on a large number of training data points—can be categorized into generative, supervised, and unsupervised approaches. Among generative methods, recently, numerous IIP solvers have employed pre-trained diffusion models (DMs) as priors (e.g.,~\cite{chung2022score, alkhouri2024sitcom}; see \cite{diffusion_survey} for a recent survey). These methods typically modify the reverse sampling steps to enable sampling from a measurement-conditioned distribution. In supervised end-to-end (E2E) approaches, a variety of techniques have been explored, including unrolled networks such as the variational network (VarNet) in \cite{sriram2020end}. A common limitation of both generative and supervised methods is their reliance not only on large amounts of training data, but also on clean (i.e., fully sampled) ground truth images—an assumption that may not hold in 
many medical and scientific applications  
such as MRI reconstruction\footnote{We note that there have been recent efforts to train DMs on degraded images, such as the study in \cite{aali2024ambient}. However, these approaches still rely on large amounts of training data.}. 
Moreover, DM-based generative methods are generally slower at inference compared to supervised counterparts. Unsupervised (or self-supervised) methods have also been explored \cite{lehtinen2018noise2noise, NEURIPS2019_0ed94223, 9103213}. While they do not require clean images, they still depend on a large amount of training data and often fall short of the performance achieved by supervised or generative models. In this paper, we introduce a method that requires neither large-scale training data nor clean images (or fully sampled measurements) measurements.


\paragraph{Training-data-free methods:} The pioneering work of \cite{ulyanov2018deep} introduced the Deep Image Prior (DIP), a fully training-data-free approach for solving IIPs, demonstrating that an untrained convolutional neural network (CNN) can effectively recover low-frequency image structures using the implicit bias of the network. However, DIP is prone to noise overfitting due to the network's over-parameterization. As a result, numerous techniques have been proposed to mitigate this issue, including the use of regularization \cite{liang2024analysis, aSeqDIP}, early stopping \cite{wang2021early}, and network re-parameterization \cite{li2023deep, heckel2019deep}. We refer the reader to the recent tutorial in \cite{alkhouri2025understanding} for a comprehensive overview of DIP and its variants. Most DIP methods require optimizing a full set of weights for each measurement vector independently. In this paper, we present a DIP-based method that learns a subset of transferable weights from only a few degraded images, addressing this gap. 



\paragraph{Low-data regime methods:} Low-data regime refers to the case where the number of training images is extremely small for training standard supervised, generative, and/or unsupervised methods. Here, we discuss two methods that operate in the low-data regime and are closely related to our work. The first is MetaDIP \cite{zhang2022metadipacceleratingdeepimage}, where the authors leverage meta-learning to fine-tune a set of hyper-parameters and learn transferable weights for accelerated DIP reconstruction at test time. MetaDIP adopts a standard single-encoder, single-decoder U-Net architecture. There are three key differences between \our and MetaDIP:
(\textit{i}) MetaDIP requires clean images during training, while our method uses only degraded images (or sub-sampled measurement vectors);
(\textit{ii}) MetaDIP does not include architectural changes during training;
(\textit{iii}) their evaluation is limited to natural image restoration tasks, whereas our method also considers more complex, large-scale MRI reconstruction. The second method is STRAINER \cite{vyas2024learning}, which extends the Implicit Neural Representation (INR) framework \cite{sitzmann2020implicit,chen2019learning} to the low-data regime. Specifically, the authors propose using a fully connected network (or multi-layer perceptron, MLP) with multiple output sub-networks to learn transferable weights. At test time, a subset of these weights is used to initialize a subset of the MLP-based INR model for image recovery. The main differences between \our and STRAINER are:
(\textit{i}) STRAINER uses MLPs with spatial coordinate input embedding, while our method is based on CNNs and DIP;
(\textit{ii}) STRAINER uses clean (fully sampled) training data, whereas \our  uses degraded images (or sub-sampled measurements); 
(\textit{iii}) the learned weights in STRAINER are used only for \textit{initializing} part of the test-time INR network, whereas in our method, we keep the parameters of the pre-trained shared encoder unchanged during testing and only adapt the parameters of the test-time decoder.

\section{Proposed method} \label{sec: prop method}



Assume that we are given a set $Y$, consisting of $M$ under-sampled measurement vectors (or degraded images), where each $\mathbf{y}_i\in Y$, for $i\in [M]:=\{1,\dots,M\}$, is an $m$-dimensional vector. 
The degraded images in set $Y$ are obtained using the following forward models: $\mathbf{y}_i = \mathbf{A}_i\mathbf{x}_i^* + \boldsymbol{\eta}_i\:,$
where $\boldsymbol{\eta}_i$ represents the noise in the measurement domain, e.g., assumed sampled from a Gaussian distribution $\mathcal{N}(\mathbf{0},\sigma^2_{\mathbf{y}_i}\mathbf{I})$, where $\sigma_{\mathbf{y}_i}>0$ denotes the noise level of the $i$-th vector. In this setting, we assume access only to $\mathbf{y}_i$ and the corresponding $\mathbf{A}_i$ for each $i\in [M]$. We note that while we adopt a linear forward model for notational simplicity, our approach can be applied to nonlinear IIPs, as we will demonstrate in our experimental results.

Here, the number of available training images, $M$, is extremely small (e.g., $M<10$), making it infeasible to learn a data distribution to be used as a prior during inference. Under this under-sampled (or degraded), low-data regime, we consider the following question:
\begin{tcolorbox}[left=1.2pt,right=1.2pt,top=1.2pt,bottom=1.2pt]
\textcolor{black}{Given a CNN architecture and a set of $M$ measurement vectors, $\mathbf{y}_i \in Y$, with corresponding forward operators $\mathbf{A}_i$, can we learn a set of weights such that, for a unseen test-time measurement vector $\mathbf{y} \notin Y$ and its operator $\mathbf{A}$, we can reconstruct the image both \textit{robustly} and \textit{efficiently}?}
\end{tcolorbox}
Here, \textit{robust} reconstruction refers to achieving high-quality recovery while being resilient to noise overfitting—an issue commonly encountered in standalone DIP methods \cite{alkhouri2025understanding, liang2024analysis}\footnote{Standalone DIP refers to the standard setting where a single measurement vector is used, and the network is optimized from randomly initialized parameters.}. \textit{Efficient} reconstruction refers to faster convergence compared to standalone DIP.

Among existing DIP methods, autoencoding Sequential DIP (aSeqDIP) \cite{aSeqDIP} has demonstrated superior performance in terms of robustness and reconstruction quality. Meanwhile, Early Stopping DIP (ES-DIP) \cite{wang2021early} and Deep Random Projector (DRP) \cite{li2023deep} have shown improved efficiency in terms of faster run-time.



To enable the learning of transferable weights--which is the central goal of this paper-- we propose the use of a modified network architecture: a shared encoder jointly learned across all $M$ degraded images, along with a separate decoder for each individual measurement vector $\mathbf{y}_i$, resulting in $M$ disentangled decoders. To mitigate noise overfitting, we adopt an input-adaptive autoencoding objective, which we describe in the next subsection. At test time, given an unseen vector $\mathbf{y}$, we use a DIP network, consisting of the learned shared encoder and a randomly initialized decoder, and perform image reconstruction by optimizing only the decoder parameters. In other words, given $\mathbf{y}$ at test time, the pre-trained encoder is used to extract latent features and the decoder parameters are optimized to adapt to the unseen measurement vector.

We note that using a single encoder with multiple decoders has also been explored in other areas of machine learning, including multi-task and federated learning~\cite{bhattacharjee2022mult, zhou2025multi}. Another example is~\cite{Zhang_2024_CVPR}, which proposed a multi-decoder architecture to accelerate the training of diffusion models.



\subsection{Proposed training optimization and network architecture} \label{sec: prop method formulation}

Assume we are given a CNN network (e.g., a U-Net~\cite{ronneberger2015u}). Let $h : \mathbb{R}^n \rightarrow \mathbb{R}^l$, parameterized by $\phi$, denote the encoder function of the CNN--i.e., the \textbf{downsampler}, consisting of a stack of convolution and pooling layers that extracts and compresses features, with $l\ll n$. Similarly, let $g :  \mathbb{R}^l \rightarrow \mathbb{R}^n$, parameterized by $\psi$, be the decoder function--i.e., the \textbf{upsampler}, comprising a stack of transposed convolution and/or interpolation layers that reconstruct the high-resolution output from the compressed representation. 

Given some input $\mathbf{z} \in \mathbb{R}^n$, the network output is expressed as $g_\psi \circ h_\phi(\mathbf{z})$. Using this setting with some measurement vector $\mathbf{y}$ and its forward operator $\mathbf{A}$, reconstruction via vanilla DIP \cite{ulyanov2018deep} is formulated as in \eqref{eqn: vanilla DIP}, where $\hat{\mathbf{x}}$ denotes the reconstructed image.
\begin{equation}\label{eqn: vanilla DIP}
    \hat{\phi},\hat{\psi} = \argmin_{\phi,\psi} \|\mathbf{A}g_\psi \circ h_\phi(\mathbf{z}) - \mathbf{y}\|^2_2, \quad\quad \hat{\mathbf{x}} = g_{\hat{\psi}} \circ h_{\hat{\phi}}(\mathbf{z})\:,
\end{equation}
%
Given a set of $M$ degraded measurements of images, $\mathbf{y}_i \in Y$, each associated with a forward operator $\mathbf{A}_i$ for $i\in [M]$, the proposed \our method utilizes a shared encoder $h_\phi$ and a distinct decoder $g_{\psi_i}$ for each training image. This results in a total of $M$ disentangled decoders tailored to the individual measurement vector. 


Given that we have $M$ measurement vectors, choosing the input is not straightforward across a training set. Hence, we develop an input-adaptive approach. To this end, during training, we propose optimizing the weights of the shared-encoder, decoder-disentangled CNN architecture using the following objective: 
%
\begin{equation}\label{eqn: ours training opt}
    \hat{\phi},\hat{\psi}_i, \hat{\mathbf{z}}_i = \argmin_{\phi,\psi_i, \mathbf{z}_i, i\in [M]} \big\{ \sum_{i\in [M]}  \|\mathbf{A}_i g_{\psi_i} \circ h_\phi(\mathbf{z}_i) - \mathbf{y}_i\|^2_2 ~~ \text{s.t.} ~~ g_{\psi_i} \circ h_\phi(\mathbf{z}_i) = \mathbf{z}_i, ~ \forall i\in [M] \big\}\:. 
\end{equation}
%
To solve \eqref{eqn: ours training opt}, we initialize $\phi,\psi_i \sim \mathcal{N}(\mathbf{0},\sigma_{\textrm{ini}}^2\mathbf{I})$ and set $\mathbf{z}_i \leftarrow \mathbf{A}^H_i \mathbf{y}_i$ for all $i\in [M]$, where $\sigma_{\textrm{ini}}$ corresponds to the standard deviation for initialization. We then perform the following alternating updates for $K$ iterations (which corresponds to the number of network inputs updates): 
\begin{subequations}\label{eqn: ours training updates}
\begin{gather}
\phi,\psi_1,\dots,\psi_M \leftarrow \argmin_{\phi,\psi_i, i\in [M]} \big[ \sum_{i\in [M]}  \|\mathbf{A}_i g_{\psi_i} \circ h_\phi(\mathbf{z}_i) - \mathbf{y}_i\|^2_2 + \lambda \| g_{\psi_i} \circ h_\phi(\mathbf{z}_i) - \mathbf{z}_i\|^2_2 \big]\:,\\
\mathbf{z}_i \leftarrow g_{\psi_i} \circ h_\phi(\mathbf{z}_i), \forall i\in [M]\:,
\end{gather}
\end{subequations}
where $\lambda$ is a regularization parameter. Similar to aSeqDIP \cite{aSeqDIP}, we alternate between two phases: (\textit{i}) perform $N \ll K$ gradient updates to optimize the network parameters $\phi$ and $\psi_i$ using the optimization in \eqref{eqn: ours training updates}, and (\textit{ii}) update the network inputs $\mathbf{z}_i$ via forward passes. Specifically, each input $\mathbf{z}_i$ is updated by passing it through the shared encoder $h_\phi$ and the corresponding decoder $g_{\psi_i}$. 

In the training optimization of \eqref{eqn: ours training updates}, the first term enforces the output of decoder $i$ to be consistent with the measurements vector $\mathbf{y}_i$ governed by $\mathbf{A}_i$. The second term is an autoencoding term that implicitly enforces the equality constraints in \eqref{eqn: ours training opt}. It is used to mitigate noise overfitting by enforcing that the output of each decoder is approximately consistent with the updated input of that particular image $\mathbf{z}_i$ as the optimization progresses. 

In what follows, we provide a justification of how we relaxed \eqref{eqn: ours training opt} into the alternating optimization updates in \eqref{eqn: ours training updates}. 
\begin{algorithm}[t]
\small
\caption{\underline{\textbf{Training:}} \textbf{U}nsupervised \textbf{G}r\textbf{o}up \textbf{DI}P \textbf{T}ransferable via weights (\our)}
\textbf{Input}: Number of training images $M$, measurement vectors $\mathbf{y}_i$ and forward operators $\mathbf{A}_i$ for $i\in [M]$, number of input updates $K$, number of gradient updates $N$, regularization parameter $\lambda$, and learning rate $\beta$. \\
\vspace{1mm}
\textbf{Output}: Trained encoder weights $\hat{\phi}$. \\
\vspace{1mm}
\textbf{Initialization}: $\mathbf{z}_i \leftarrow \mathbf{A}_i^H \mathbf{y}_i$ ; $\phi, \psi_i \sim \mathcal{N}(\mathbf{0}, \sigma_{\textrm{ini}} \mathbf{I})$ for $i\in [M]$. 
\\
\vspace{1mm}
\small{1:} \textbf{For} $K$ iterations, \textbf{Do}  \\
\vspace{1mm}
\small{2:} \hspace{4mm} \textbf{For} $N$ iterations, \textbf{Do} (Shared encoder and $M$ decoders parameters update)\\
\vspace{1mm}
\small{3:} \hspace{8mm}$\psi_i \leftarrow \psi_i - \beta \nabla_{\psi_i} \Big[ \|\mathbf{A}_i g_{\psi_i} \circ h_\phi(\mathbf{z}_i) - \mathbf{y}_i\|^2_2 + \lambda \| g_{\psi_i} \circ h_\phi(\mathbf{z}_i) - \mathbf{z}_i\|^2_2 \Big], \forall i\in [M]$\\
\vspace{1mm}
\small{4:} \hspace{8mm}$\phi \leftarrow \phi - \beta \nabla_\phi \Big[ \sum_{i\in [M]}  \|\mathbf{A}_i g_{\psi_i} \circ h_\phi(\mathbf{z}_i) - \mathbf{y}_i\|^2_2 + \lambda \| g_{\psi_i} \circ h_\phi(\mathbf{z}_i) - \mathbf{z}_i\|^2_2 \Big]$.\\
\vspace{1mm}
\small{5:} \hspace{4mm}\textbf{Obtain} $\mathbf{z}_i \leftarrow g_{\psi_i} \circ h_\phi(\mathbf{z}_i)$. (Network input update for each training image $i\in [M]$)  \\
\vspace{1mm}
\small{6:} \textbf{Trained weights:} $\hat{\phi} \leftarrow \phi$ \\
\vspace{1mm}
\vspace{-3.5mm}
\label{alg: training}
\end{algorithm}
The general intuition stems from prior studies that highlight the impact of network inputs in DIP-based reconstruction \cite{zhao2020reference, aSeqDIP, liang2024analysis}, where the study in \cite{aSeqDIP} shows that reconstruction quality improves when the network input is closer to the ground truth. 
\begin{figure*}[t]
\centering
\includegraphics[width=14cm]{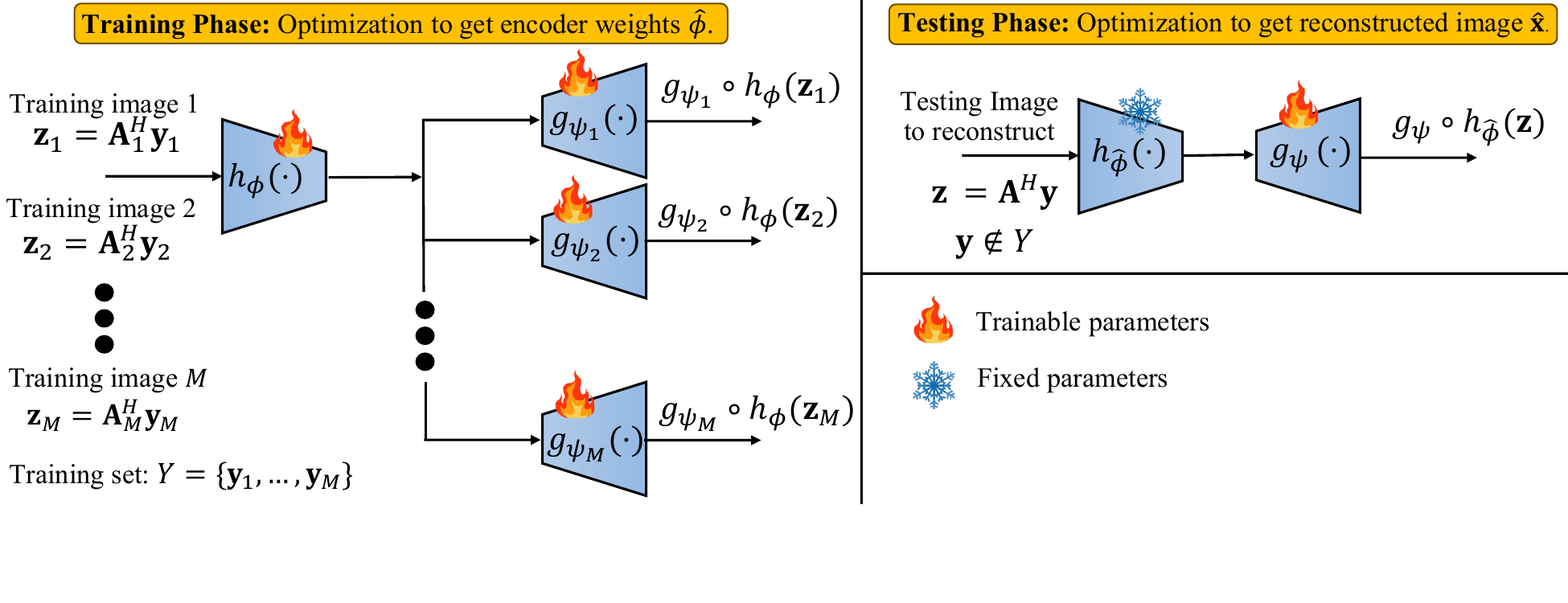}
\caption{\small{Illustrative block diagram \our. During training (\textit{left}), we learn the shared encoder by optimizing over $\phi, \psi_1,\dots, \psi_M$. At inference (\textit{right}), given some unseen measurement vector $\mathbf{y}$ and its forward operator $\mathbf{A}$, we reconstruct the image $\hat{\mathbf{x}}$ by optimizing over $\psi$.}}
\label{fig: BD}
\end{figure*}

Assume that the iterates in \eqref{eqn: ours training updates} converge to $\phi^*, \psi^*_i, \mathbf{z}^*_i, \forall i\in [M]$. Then, according to the second equation in \eqref{eqn: ours training updates}, for continuous mappings $g_{\psi_i}\circ h_\phi(\cdot), \forall i\in [M]$, we have $\mathbf{z}^*_i = g_{\psi^*_i}\circ h_{\phi^*}(\mathbf{z}^*_i)$. Substituting this into the first equation of \eqref{eqn: ours training updates}, in the limit, we get $\phi^*,\psi^*_i = \argmin_{\phi,\psi_i, i\in [M]} \big\{ \sum_{i\in [M]}  \|\mathbf{A}_i g_{\psi_i} \circ h_\phi(\mathbf{z}^*_i) - \mathbf{y}_i\|^2_2 \quad \text{s.t.} \quad g_{\psi_i} \circ h_\phi(\mathbf{z}^*_i) = \mathbf{z}^*_i, \quad \forall i\in [M] \big\}\:,$
which corresponds to the minimizer of \eqref{eqn: ours training opt} given a fixed $\mathbf{z}^*_i$. For each decoder, the limit points of \eqref{eqn: ours training updates} correspond to the solution of a constrained version of vanilla DIP in \eqref{eqn: vanilla DIP}. The constraint enforces additional prior that could alleviate overfitting. While it is not straightforward to use first-order gradient-based algorithms for solving \eqref{eqn: ours training opt} given the hard equality constraints, the limit points in the updates of \eqref{eqn: ours training updates} nevertheless minimize~\eqref{eqn: ours training opt}. Furthermore, the network input is updated by a sequential feed forward process without needing expensive gradient-based updates.


The training procedure of \our is given in Algorithm~\ref{alg: training}. In the initialization step, we assume the encoder and decoder weights are of equal size, though this is not necessary requirement. We emphasize that while steps 2 to 4 describe updating the network parameters using a fixed-step gradient descent, we use the ADAM optimizer \cite{kingma2014adam}, initialized with learning rate $\beta$. The computational requirements for training \our are determined by: (\textit{i}) the $NK$ gradient-based parameter updates, and (\textit{ii}) the number of function evaluations (network forward passes) necessary for updating $\mathbf{z}_i$, which is $MK$. See Figure~\ref{fig: BD} (\textit{left}) for an illustrative block diagram of the training phase.

\paragraph{Testing phase:} Provided with the trained weights of the shared encoder, i.e., parameters $\hat{\phi}$ from Algorithm~\ref{alg: training}, and unseen measurement vector $\mathbf{y}$ with its forward operator $\mathbf{A}$, at inference, we propose to use a new network mapping denoted by $h_{\phi}\circ g_{\psi}(\cdot)$, where $\phi$ is set to the trained weights $\hat{\phi}$. Then, $\psi$ is initialized using the standard way. More specifically, we set $\mathbf{z}\leftarrow \mathbf{A}^H \mathbf{y}$, $\phi\leftarrow \hat{\phi}$, and $\psi\sim \mathcal{N}(\mathbf{0},\sigma^2_{\textrm{ini}}\mathbf{I})$, and perform the following updates for $K$ iterations. 
\begin{subequations}\label{eqn: ours testing updates}
\begin{gather}
\psi \leftarrow \argmin_{\psi} \big[ \|\mathbf{A} g_{\psi} \circ h_{\hat{\phi}}(\mathbf{z}) - \mathbf{y}\|^2_2 + \lambda \|g_{\psi} \circ h_{\hat{\phi}}(\mathbf{z}) - \mathbf{z}\|^2_2 \big]\:,\\
\mathbf{z} \leftarrow g_{\psi} \circ h_{\hat{\phi}}(\mathbf{z})\:. 
\end{gather}
\end{subequations}
In \eqref{eqn: ours testing updates}, we only optimize over $\psi$ while the parameters of the pre-trained encoder are fixed (i.e., $\hat{\phi}$)\footnote{In Appendix~\ref{sec: ablation frzen encoder vs. initialized one}, we conduct a study to show the difference between freezing the encoder (our method) and only initializing it (as was done in STRAINER \cite{vyas2024learning}).}. Similar to the training algorithm, for every input update (i.e., second equation in \eqref{eqn: ours testing updates}), we use $N\ll K$ gradient steps for the optimization in \eqref{eqn: ours testing updates}. The intuition of using the steps in \eqref{eqn: ours testing updates} is the same as in aSeqDIP \cite{aSeqDIP}. However, we emphasize that our goal here is different as in \our, the aim is to learn transferable weights for better and accelerated reconstructions at testing time. Algorithm~\ref{alg: testing} presents the procedure for solving \eqref{eqn: ours testing updates}. The training and testing phases of \our are illustrated in Figure~\ref{fig: BD}. 


\begin{algorithm}[t]
\small
\caption{\underline{\textbf{Testing}:} \textbf{U}nsupervised \textbf{G}r\textbf{o}up \textbf{DI}P \textbf{T}ransferable via weights (\our-$M$)}
\textbf{Input}: Pre-trained encoder weights $\hat{\phi}$ from Algorithm~\ref{alg: training}, measurement vector $\mathbf{y}$, forward operator $\mathbf{A}$, number of input updates $K$, number of gradient updates $N$, regularization parameter $\lambda$, and learning rate $\beta$. \\
\vspace{1mm}
\textbf{Output}: Reconstructed image $\hat{\mathbf{x}}$. \\
\vspace{1mm}
\textbf{Initialization}: $\mathbf{z} \leftarrow \mathbf{A}^H \mathbf{y}$ ; $\psi \sim \mathcal{N}(\mathbf{0}, \sigma_{\textrm{ini}}\mathbf{I})$. 
\\
\vspace{1mm}
\small{1:} \textbf{For} $K$ iterations, \textbf{Do}  \\
\vspace{1mm}
\small{2:} \hspace{4mm} \textbf{For} $N$ iterations, \textbf{Do} (Decoder parameters update)\\
\vspace{1mm}
\small{3:} \hspace{8mm}$\psi \leftarrow \psi - \beta \nabla_{\psi} \Big[ \|\mathbf{A} g_{\psi} \circ h_{\hat{\phi}}(\mathbf{z}) - \mathbf{y}\|^2_2 + \lambda \| g_{\psi} \circ h_{\hat{\phi}}(\mathbf{z}) - \mathbf{z}\|^2_2 \Big]$\\
\vspace{1mm}
\small{4:} \hspace{4mm}\textbf{Obtain} $\mathbf{z} \leftarrow g_{\psi} \circ h_{\hat{\phi}}(\mathbf{z})$. (Network input update)  \\
\vspace{1mm}
\small{5:} \textbf{Reconstructed Image:} $\hat{\mathbf{x}} \leftarrow g_{\psi} \circ h_{\hat{\phi}}(\mathbf{z})$ \\
\vspace{1mm}
\vspace{-3.5mm}
\label{alg: testing}
\end{algorithm}


\subsection{Intuition behind using $M$ decoders in the training of \our}\label{sec: method intut}


In this subsection, we describe the rationale behind using $M$ disentangled decoders in the \our  training phase and how this design choice influences the quality of the learned encoder weights. Specifically, we aim to answer the following question: \textit{Why does learning with multiple decoders result in a better shared encoder for test-time reconstruction?} To build this intuition, we first examine the standard architecture comprising a single encoder and a single decoder--denoted by $h_{\phi_s}\circ g_{\psi_s}(\cdot)$--trained across $M$ degraded images. The parameters are initialized as $\phi_s,\psi_s\sim \mathcal{N}(\mathbf{0},\sigma^2_{\textrm{ini}}\mathbf{I})$ and the inputs as $\mathbf{z}_i \leftarrow \mathbf{A}^H_i \mathbf{y}_i, \forall i\in [M]$. Then, we perform the following alternating optimization (which extends \cite{aSeqDIP} to $M>1$ measurements vectors). 
\begin{subequations}\label{eqn: shared decoder training updates}
\begin{gather}
\phi_s,\psi_s \leftarrow \argmin_{\phi_s,\psi_s} \big[ \sum_{i\in [M]}  \|\mathbf{A}_i g_{\psi_s} \circ h_{\phi_s}(\mathbf{z}_i) - \mathbf{y}_i\|^2_2 + \lambda \| g_{\psi_s} \circ h_{\phi_s}(\mathbf{z}_i) - \mathbf{z}_i\|^2_2 \big]\:,\\
\mathbf{z}_i \leftarrow g_{\psi_s} \circ h_{\phi_s}(\mathbf{z}_i), \forall i\in [M]\:.
\end{gather}
\end{subequations}
Here, a single decoder $g_{\psi_s}$ is shared across all training images. While the inputs $\mathbf{z}_i$ are updated and initialized individually, they are all later optimized using the same encoder-decoder network. 
The gradient updates for this architecture are: $\psi_s \leftarrow \psi_s - \beta \nabla_{\psi_s} \big[ \sum_{i\in [M]}  \|\mathbf{A}_i g_{\psi_s} \circ h_{\phi_s}(\mathbf{z}_i) - \mathbf{y}_i\|^2_2 + \lambda \| g_{\psi_s} \circ h_{\phi_s}(\mathbf{z}_i) - \mathbf{z}_i\|^2_2 \big]\:,$ and $\phi_s \leftarrow \phi_s - \beta \nabla_{\phi_s} \big[ \sum_{i\in [M]}  \|\mathbf{A}_i g_{\psi_s} \circ h_{\phi_s}(\mathbf{z}_i) - \mathbf{y}_i\|^2_2 + \lambda \| g_{\psi_s} \circ h_{\phi_s}(\mathbf{z}_i) - \mathbf{z}_i\|^2_2 \big]\:.$

From these expressions, the encoder $h_{\phi_s}$ is trained with using back-propagated updates from a single decoder. In contrast, in \our, the encoder receives supervision from $M$ distinct decoders, each independently optimized to reconstruct a different training image (steps 3 and 4 in Algorithm~\ref{alg: training}). This over-parameterization allows each decoder to specialize in reconstructing its corresponding input, thereby improving the enforcement of measurement consistency across the training set. 


Moreover, because the encoder must support multiple decoders reconstructing different training samples, it is encouraged to extract more generalizable and robust features. This leads to the intuition: \textit{Encoders that yield more accurate reconstructions during training tend to reconstruct better during testing}. We empirically validate this claim in Section~\ref{sec: exp results} by comparing both training and test-time reconstructions using a shared decoder versus disentangled decoders (our method).

\section{Experimental results} \label{sec: exp results}








We consider three tasks: MRI reconstruction from undersampled measurements, super resolution (SR), and non-linear deblurring (NDB). For MRI, we use the knee portion of the fastMRI dataset \cite{zbontar2018fastmri}. The forward model is $\mathbf{y} \approx \mathbf{A} \mathbf{x}^*$. The multi-coil data is obtained using $15$ coils and is cropped to a resolution of $320 \times 320$ pixels. To simulate undersampling of the MRI k-space, we use a Cartesian sampling pattern with 4x and 8x acceleration factors (AF). Sensitivity maps for the coils are obtained using the BART toolbox~\citep{tamir2016generalized}. For the tasks of SR and NDB, we use the FFHQ dataset \citep{karras2019style}. For all three tasks, we test \our and baselines with $20$ randomly selected degraded images. Backbone architectures are adopted from \cite{ronneberger2015u} and \cite{newell2016stacked} for MRI and SR (and NDB), respectively. For evaluation metrics, we use the Peak Signal to Noise Ratio (PSNR), the Structural SIMilarity (SSIM) index \citep{wang2004image}, and run-time. For natural images, we also report LPIPS \cite{zhang2018unreasonable}. 
 \begin{figure}
\vspace{-0.5cm}
    \centering    \includegraphics[width=0.62\textwidth]{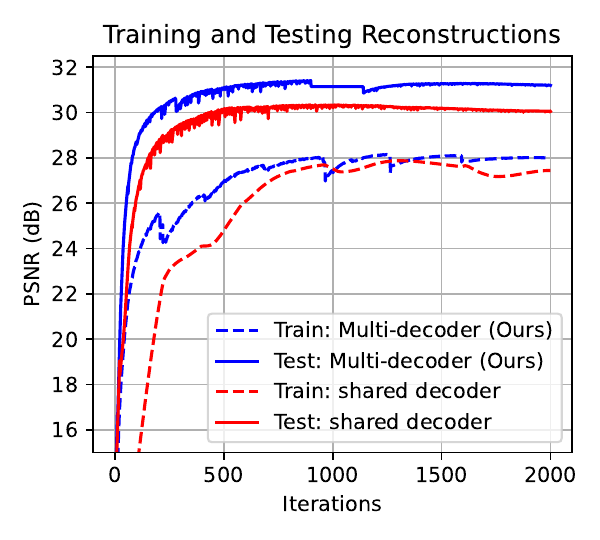}
    \caption{\small{{SR averaged PSNR curves of SR ($6$ for training and $20$ for testing) of \our vs. the shared-decoder case (i.e., \eqref{eqn: shared decoder training updates}).}}}
 
    \label{fig: impact of decoders}
\end{figure}
All the experiments are run on a single RTX5000 GPU machine. Further implementation details are provided in Appendix~\ref{sec: append imp details}. For training-data-free methods, we use three regularization-based recent methods, aSeqDIP \cite{aSeqDIP}, Self-Guided DIP \cite{liang2024analysis}, and Deep Random Projector \cite{li2023deep} (as a DIP acceleration method), and one early stopping (ES) method, ES-DIP \cite{wang2021early}. For data-driven MRI baselines, we use two recent SOTA DM-based methods, SITCOM-MRI \cite{alkhouri2024sitcom} and DDS \cite{ye2024decomposed}, and a very competitive supervised model, VarNet \cite{sriram2020end}. For natural images, we also use SITCOM \cite{alkhouri2024sitcom}, in addition to DPS \cite{chung2022diffusion} and STRAINER \cite{vyas2024learning}. STRAINER is a representative of low-data regime methods. The selection criteria of these baselines depend on methods' ability to achieve competitive reconstruction quality and high robustness to noise overfitting (for the case of DIP) for which several medical and natural image recovery tasks were considered (linear and non-linear). Baselines' implementation details is given in Appendix~\ref{sec: append imp details}. For \our, we present results with three models, trained with $4$, $5$, and $6$ degraded imaged (see Appendix~\ref{sec: append different number of training M} for results with $M>6$). Our code is available online\footnote{\tiny{\url{https://github.com/sjames40/UGoDIT}}}. We test \our and other data-driven methods with in-distribution test instances, i.e., images from the testing sets of fastMRI and FFHQ. Out-of-distribution evaluation is provided in Appendix~\ref{sec: appen ood eval}. Ablation studies on the number of layers and the number of training images are given in Appendix~\ref{sec: append different arch} and Appendix~\ref{sec: append different number of training M}, respectively. Step size is set to $\beta=0.0001$. The regularization parameter is set to $\lambda=2$, following the study in Appendix~\ref{sec: ablation lambda} (where we also show the robustness to noise overfitting). For $(N,K)$, we use $(2,2000)$, $(10,2000)$, and $(10,2000)$ for MRI, SR, and NDB, respectively, following the study in Appendix~\ref{sec: appen NK}.  




\paragraph{Impact of using $M$ decoders in \our:} Here, we examine the effect of the multi-decoder architecture proposed in the training of \our. Specifically, we assess how using $M>1$ disentangled decoders influences the quality of the shared encoder's learned weights during both training and testing for the super resolution task. To this end, we compare our method against a baseline where $M=6$ degraded are trained using a standard single-encoder, single-decoder architecture, as described in \eqref{eqn: shared decoder training updates}. At inference, both approaches follow the procedure in \eqref{eqn: ours testing updates} using their respective trained encoders. Figure~\ref{fig: impact of decoders} shows the average PSNR curves during training (dashed) and testing (solid with $20$ test images) of \our and the shared-decoder case. As seen, employing multiple decoders yields approximately over a $1$ dB to $1.5$ dB improvement in both training and testing. These findings empirically support the rationale discussed in Section~\ref{sec: method intut}. \textcolor{black}{Furthermore, we observe a gap of approximately 3 to 4 dB between the training and testing curves. We hypothesize that this arises because, during training, the network parameters are jointly optimized to reconstruct $M=6$ images, whereas the testing curve reflects the average performance over 20 separate inference-time optimizations.} Appendix~\ref{sec: appen learned features for uGoDiT} provides an additional study on the learned features of \our.

\begin{figure*}[t]
\centering
\includegraphics[width=14.1cm]{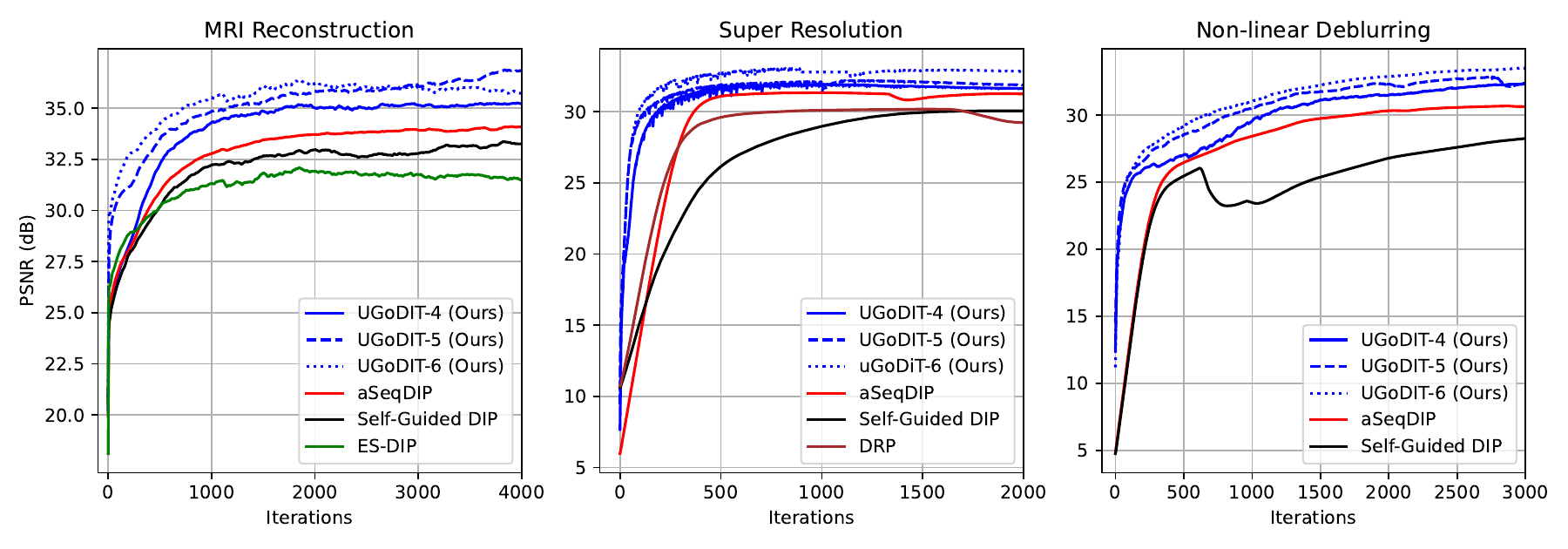}
\caption{\small{PSNR curves of \our-$M$ and baselines for the tasks of MRI (\textit{left}), SR (\textit{middle}), and NDB (\textit{right}) averaged over 20 test images. Iterations (x-axis) correspond to $NK$.}}
\label{fig: main comp with DIPs}
\end{figure*}
\begin{figure*}[t]
\centering
\includegraphics[width=14cm]{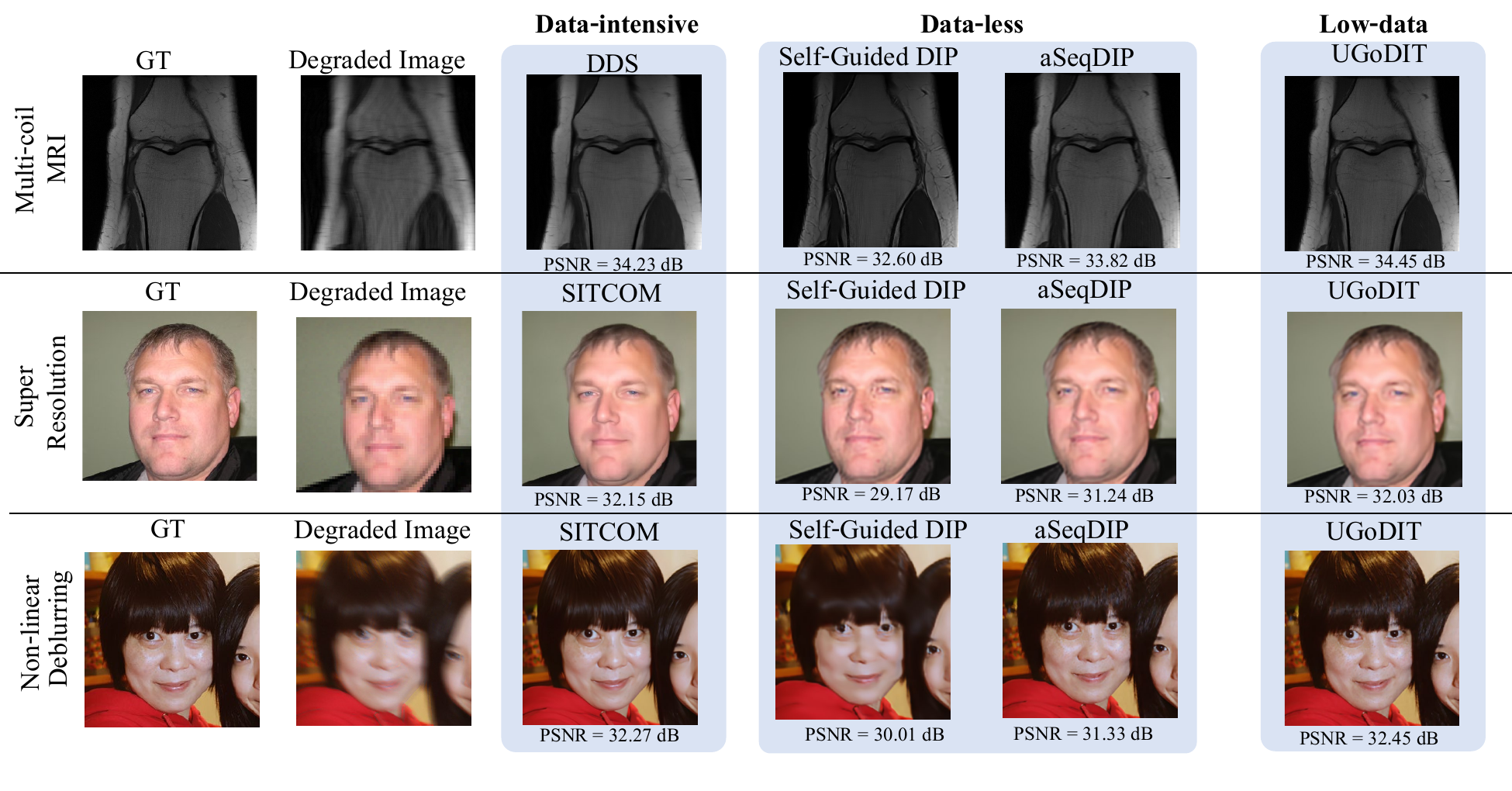}
\caption{\small{Reconstructed/recovered images using \our (last column) and baselines (columns 3 to 5). \textcolor{black}{As observed, \our return comparable PSNR results with data-intensive methods all without the need of large amount of clean images (or fully sampled measurement vectors).} See Appendix~\ref{sec: appen more visual} for more visualizations.}  } 
\label{fig: visual main}
\end{figure*}
\begin{table*}[t]
\small
\centering
\resizebox{0.88\textwidth}{!}{%
\begin{tabular}{c|cc|ccc|ccc}
\toprule
\multirow{2}{*}{\textbf{Method}} & {\textbf{Training}} & \multirow{2}{*}{\textbf{Type}} & \multicolumn{3}{c|}{\textbf{Acceleration Factor (AF)}: 4x} & \multicolumn{3}{c}{\textbf{Acceleration Factor (AF)}: 8x} \\
 & \textbf{images} $M$ & &  PSNR ($\uparrow$) & SSIM ($\uparrow$) &  run-time ($\downarrow$) & PSNR ($\uparrow$) & SSIM ($\uparrow$) &  run-time ($\downarrow$) \\
\midrule
VarNet \cite{sriram2020end} & 8000 & Fully sampled & 33.06\tiny{$\pm$0.59} & 0.854\tiny{$\pm$0.12} &
5.02\tiny{$\pm$2.12} & 32.03\tiny{$\pm$0.35} & 
0.829\tiny{$\pm$0.08} & 5.45\tiny{$\pm$2.11}  \\
DDS \cite{ye2024decomposed} & 19460 & Fully sampled & 34.95\tiny{$\pm$0.74} & 0.954\tiny{$\pm$0.10} & 25.20\tiny{$\pm$5.22} & 31.72\tiny{$\pm$1.88} & 0.876\tiny{$\pm$0.06} & 31.02\tiny{$\pm$2.01}  \\
SITCOM-MRI \cite{alkhouri2024sitcom} & 19460 & Fully sampled & 36.33\tiny{$\pm$0.37} & 0.962\tiny{$\pm$0.08} & 76.24\tiny{$\pm$6.25} & 32.78\tiny{$\pm$0.64} & 0.892\tiny{$\pm$0.02} & 82.45\tiny{$\pm$4.20}  \\


\midrule
\our (Ours)  & 4 & Sub-sampled & 35.45\tiny{$\pm$0.34} & 0.957\tiny{$\pm$0.08} & 45.20\tiny{$\pm$4.20} & 32.04\tiny{$\pm$0.45} & 0.878\tiny{$\pm$0.06} & 47.42\tiny{$\pm$5.64}  \\
\our (Ours)  & 5 & Sub-sampled & 35.74\tiny{$\pm$0.41} & 0.959\tiny{$\pm$0.09} & 48.12\tiny{$\pm$5.42} & 32.36\tiny{$\pm$0.51} & 0.881\tiny{$\pm$0.06} & 48.12\tiny{$\pm$3.24}  \\
\our (Ours)  & 6 & Sub-sampled & 36.05\tiny{$\pm$0.34} & 0.960\tiny{$\pm$0.08} & 48.42\tiny{$\pm$5.40} & 32.54\tiny{$\pm$0.45} & 0.883\tiny{$\pm$0.06} & 49.33\tiny{$\pm$7.44}  \\
\bottomrule
\end{tabular}}
\vspace{-0.2cm}
\caption{\small{MRI reconstruction average results (with run-time in seconds) across two AFs of \our vs. data-centric methods. In SITCOM-MRI and DDS, the DM is trained on 973 MRI volumes each with 20 images. Column 3 corresponds to the ``Type'' of the training images. Values past $\pm$ represent the standard deviation. }}
\label{tab: main res MRI with data methods}
\end{table*}
\begin{table*}[t]
\small
\centering
\resizebox{1\textwidth}{!}{%
\begin{tabular}{c|cc|cccc|cccc}
\toprule
\multirow{2}{*}{\textbf{Method}} & {\textbf{Training}} & \multirow{2}{*}{\textbf{Type}} & \multicolumn{4}{c|}{\textbf{Super Resolution} (SR)} & \multicolumn{4}{c}{\textbf{Non-linear Deblurring} (NDB)} \\
 & \textbf{images} $M$  & &  PSNR ($\uparrow$) & SSIM ($\uparrow$) & LPIPS ($\downarrow$) & run-time ($\downarrow$) &  PSNR ($\uparrow$) & SSIM ($\uparrow$) & LPIPS ($\downarrow$) & run-time ($\downarrow$) \\
\midrule
DPS \cite{chung2022diffusion} &  49000 & Clean & 24.44\tiny{$\pm$0.56} & 0.801\tiny{$\pm$0.032} & 0.26\tiny{$\pm$0.022}  & 75.60\tiny{$\pm$15.20} &  
 23.42\tiny{$\pm$2.15} & 0.757\tiny{$\pm$0.042} & 0.279\tiny{$\pm$0.067}  & 93.00\tiny{$\pm$26.40}  \\
SITCOM \cite{alkhouri2024sitcom} &  49000 & Clean &30.68\tiny{$\pm$1.02} & 0.867\tiny{$\pm$0.045} 
& 0.142\tiny{$\pm$0.056}  & 27.00\tiny{$\pm$4.80} 
& 30.12\tiny{$\pm$0.68} & 0.903\tiny{$\pm$0.042} 
& 0.145\tiny{$\pm$0.037}  & 33.45\tiny{$\pm$9.40} \\

\midrule
STRAINER \cite{vyas2024learning} & 4 & Clean & 29.03\tiny{$\pm$1.30} & 0.841\tiny{$\pm$0.12} & 0.189\tiny{$\pm$0.045} & 32.12\tiny{$\pm$6.20} & 27.20\tiny{$\pm$1.30} & 0.812\tiny{$\pm$0.21} & 0.201\tiny{$\pm$0.055} & 33.40\tiny{$\pm$7.24}  \\
STRAINER \cite{vyas2024learning} & 5 & Clean & 29.24\tiny{$\pm$0.89} & 0.846\tiny{$\pm$0.15} & 0.181\tiny{$\pm$0.056} & 33.45\tiny{$\pm$5.78} & 27.39\tiny{$\pm$0.86} & 0.817\tiny{$\pm$0.24} & 0.197\tiny{$\pm$0.035} & 32.60\tiny{$\pm$7.24}  \\
STRAINER \cite{vyas2024learning} & 6 & Clean & 29.45\tiny{$\pm$0.56} & 0.849\tiny{$\pm$0.08} & 0.177\tiny{$\pm$0.041} & 31.12\tiny{$\pm$6.20} & 27.41\tiny{$\pm$0.56} & 0.819\tiny{$\pm$0.043} & 0.189\tiny{$\pm$0.048} & 30.50\tiny{$\pm$4.67}  \\

\midrule
\our (Ours) & 4 & Degraded & 30.55\tiny{$\pm$0.67} & 0.859\tiny{$\pm$0.045} & 0.154\tiny{$\pm$0.067} & 29.25\tiny{$\pm$5.54} & 29.89\tiny{$\pm$0.52} & 0.897\tiny{$\pm$0.078} & 0.167\tiny{$\pm$0.045} & 28.88\tiny{$\pm$4.67}  \\
\our (Ours) & 5 & Degraded & 30.62\tiny{$\pm$0.38} & 0.863\tiny{$\pm$0.067} & 0.149\tiny{$\pm$0.035} & 30.15\tiny{$\pm$6.74} & 30.09\tiny{$\pm$0.37} & 0.899\tiny{$\pm$0.057} & 0.156\tiny{$\pm$0.035} & 29.29\tiny{$\pm$3.78}  \\

\our (Ours)& 6 & Degraded & 30.75\tiny{$\pm$0.58} & 0.871\tiny{$\pm$0.054} & 0.157\tiny{$\pm$0.058} & 30.41\tiny{$\pm$5.54} & 30.19\tiny{$\pm$0.46} & 0.906\tiny{$\pm$0.067} & 0.147\tiny{$\pm$0.035} & 30.18\tiny{$\pm$3.89}  \\

\bottomrule
\end{tabular}}
\caption{\small{Restoration results of \our vs. data-centric methods. Column 3 corresponds to the ``Type'' of the training images. Following \cite{chung2022diffusion, alkhouri2024sitcom}, we use $\sigma_{\mathbf{y}_i} = 0.05, \forall, i \in [M]$ and $M\in \{4,5,6\}$. \our uses the degraded versions of the clean training images used for STRAINER. Values past $\pm$ represent standard deviation.}}
\label{tab: main res natural images with data methods}
\end{table*}

\paragraph{Comparison with DIP-based methods:} Here, we aim to answer the question: \textit{Given a set of degraded test images, does using \our offer any advantages over running a DIP-based method independently on each image?}  Figure~\ref{fig: main comp with DIPs} presents the average PSNR curves, comparing \our (trained with 4, 5, and 6 degraded images) with four recent SOTA DIP baseline methods. In terms of PSNR improvements, \our achieves at least a 2 dB gain over aSeqDIP for MRI. For SR, we observe a marginal improvement—around 1 dB or slightly less—compared to aSeqDIP, but more than 2 dB over DRP and Self-Guided DIP. For NDB, \our consistently outperforms the second-best method (aSeqDIP) by at least 1 dB. Regarding acceleration, \our demonstrates a clear advantage in all three tasks, as measured by the number of iterations required to reach target PSNR values. For instance, in MRI, \our reaches 32.5 dB in 500 iterations or fewer, whereas aSeqDIP requires over 750 iterations to achieve the same PSNR. This trend also holds for SR and NDB. For example, in SR, \our-6 requires only about 125 steps to reach 30 dB, while aSeqDIP needs more than 250, and DRP nearly 600. Another noteworthy observation is that increasing the number of training images used by \our generally leads to better reconstruction quality. Collectively, these results highlight that \our provides improvements in both reconstruction performance and convergence speed.

\paragraph{Comparison with data-driven methods:} Here, we compare \our with data-driven methods. Table~\ref{tab: main res MRI with data methods} (resp. Table~\ref{tab: main res natural images with data methods}) presents the results for MRI (resp. SR and NDB). For MRI, all three variants of \our outperform DDS and VarNet in terms of PSNR, although they are slower. Compared to SITCOM-MRI, \our achieves slightly lower PSNR but is faster in terms of run-time. For SR and NDB, \our achieves comparable PSNR performance to SITCOM. Compared to STRAINER, \our provides improvements of over 1 dB for SR and 2 dB for NDB, given similar run-times—importantly, without requiring clean images for training. Overall, these results demonstrate that \our can achieve competitive performance with SOTA data-driven methods, all without relying on large quantities of fully sampled (i.e., clean) training images. See also the visualizations Figure~\ref{fig: visual main}. 

\section{Conclusion} \label{sec: conc}

This paper introduced \our, a fully unsupervised framework for solving inverse imaging problems in the low-data regime, where only a very small number of degraded images are available during training. By learning a shared encoder with transferable weights across multiple decoders, \our effectively bridges the gap between the recent SOTA data-intensive generative models and training-data-free deep mage prior (DIP) methods. Unlike DIP approaches that optimize a separate network per image, \our leverages the shared encoder to accelerate convergence and improve reconstruction quality. Empirical evaluations across MRI, super-resolution, and non-linear deblurring tasks show that \our achieves results comparable to SOTA supervised and diffusion-based methods, all without requiring access to a large number of clean ground-truth images. These results highlight the potential of \our as a data-efficient solution for inverse problems. Limitations and future works are discussed in Appendix~\ref{sec: limitation and future}.

\bibliography{refs}

\begin{thebibliography}{50}
\providecommand{\natexlab}[1]{#1}
\providecommand{\url}[1]{\texttt{#1}}
\expandafter\ifx\csname urlstyle\endcsname\relax
  \providecommand{\doi}[1]{doi: #1}\else
  \providecommand{\doi}{doi: \begingroup \urlstyle{rm}\Url}\fi

\bibitem[Manocha and Canny(2002)]{manocha2002efficient}
Dinesh Manocha and John~F Canny.
\newblock Efficient inverse kinematics for general 6r manipulators.
\newblock \emph{IEEE transactions on robotics and automation}, 10\penalty0 (5):\penalty0 648--657, 2002.

\bibitem[Ren et~al.(2018)Ren, Chen, Li, Chen, and Li]{ren2018deep}
Shaofei Ren, Guorong Chen, Tiange Li, Qijun Chen, and Shaofan Li.
\newblock A deep learning-based computational algorithm for identifying damage load condition: an artificial intelligence inverse problem solution for failure analysis.
\newblock \emph{Computer Modeling in Engineering \& Sciences}, 117\penalty0 (3):\penalty0 287--307, 2018.

\bibitem[Levis et~al.(2022)Levis, Srinivasan, Chael, Ng, and Bouman]{levis2022gravitationally}
Aviad Levis, Pratul~P Srinivasan, Andrew~A Chael, Ren Ng, and Katherine~L Bouman.
\newblock Gravitationally lensed black hole emission tomography.
\newblock In \emph{Proceedings of the IEEE/CVF Conference on Computer Vision and Pattern Recognition}, pages 19841--19850, 2022.

\bibitem[BniLam and Al-Khoury(2020)]{bnilam2020parameter}
Noori BniLam and Rafid Al-Khoury.
\newblock Parameter identification algorithm for ground source heat pump systems.
\newblock \emph{Applied energy}, 264:\penalty0 114712, 2020.

\bibitem[Ledig et~al.(2017)Ledig, Theis, Husz{\'a}r, Caballero, Cunningham, Acosta, Aitken, Tejani, Totz, Wang, et~al.]{ledig2017photo}
Christian Ledig, Lucas Theis, Ferenc Husz{\'a}r, Jose Caballero, Andrew Cunningham, Alejandro Acosta, Andrew Aitken, Alykhan Tejani, Johannes Totz, Zehan Wang, et~al.
\newblock Photo-realistic single image super-resolution using a generative adversarial network.
\newblock In \emph{Proceedings of the IEEE conference on computer vision and pattern recognition}, pages 4681--4690, 2017.

\bibitem[Goyal et~al.(2020)Goyal, Dogra, Agrawal, Sohi, and Sharma]{GOYAL2020220}
B.~Goyal, A.~Dogra, S.~Agrawal, B.S. Sohi, and A.~Sharma.
\newblock Image denoising review: From classical to state-of-the-art approaches.
\newblock \emph{Information Fusion}, 55:\penalty0 220--244, 2020.
\newblock ISSN 1566-2535.
\newblock \doi{https://doi.org/10.1016/j.inffus.2019.09.003}.
\newblock URL \url{https://www.sciencedirect.com/science/article/pii/S1566253519301861}.

\bibitem[Tran et~al.(2021)Tran, Tran, Phung, and Hoai]{tran2021explore}
Phong Tran, Anh~Tuan Tran, Quynh Phung, and Minh Hoai.
\newblock Explore image deblurring via encoded blur kernel space.
\newblock In \emph{Proceedings of the IEEE/CVF conference on computer vision and pattern recognition}, pages 11956--11965, 2021.

\bibitem[Elharrouss et~al.(2020)Elharrouss, Almaadeed, Al-Maadeed, and Akbari]{inpainting2022}
O.~Elharrouss, N.~Almaadeed, S.~Al-Maadeed, and Y.~Akbari.
\newblock Image inpainting: A review.
\newblock \emph{Neural Processing Letters}, 51:\penalty0 2007--2028, 2020.

\bibitem[Fessler(2010)]{5484183}
J.~A. Fessler.
\newblock {Model-Based Image Reconstruction for MRI}.
\newblock \emph{IEEE Signal Processing Magazine}, 27\penalty0 (4):\penalty0 81--89, 2010.
\newblock \doi{10.1109/MSP.2010.936726}.

\bibitem[McCann et~al.(2017)McCann, Jin, and Unser]{mccann2017convolutional}
Michael~T McCann, Kyong~Hwan Jin, and Michael Unser.
\newblock Convolutional neural networks for inverse problems in imaging: A review.
\newblock \emph{IEEE Signal Processing Magazine}, 34\penalty0 (6):\penalty0 85--95, 2017.

\bibitem[Alkhouri et~al.(2024{\natexlab{a}})Alkhouri, Liang, Wang, Qu, and Ravishankar]{alkhouri2024diffusion}
Ismail Alkhouri, Shijun Liang, Rongrong Wang, Qing Qu, and Saiprasad Ravishankar.
\newblock Diffusion-based adversarial purification for robust deep mri reconstruction.
\newblock In \emph{IEEE International Conference on Acoustics, Speech and Signal Processing (ICASSP)}, pages 12841--12845. IEEE, 2024{\natexlab{a}}.

\bibitem[Chung et~al.(2024)Chung, Lee, and Ye]{ye2024decomposed}
Hyungjin Chung, Suhyeon Lee, and Jong~Chul Ye.
\newblock Decomposed diffusion sampler for accelerating large-scale inverse problems.
\newblock In \emph{ICLR}, 2024.

\bibitem[Chung and Ye(2022)]{chung2022score}
Hyungjin Chung and Jong~Chul Ye.
\newblock Score-based diffusion models for accelerated mri.
\newblock \emph{Medical image analysis}, 80:\penalty0 102479, 2022.

\bibitem[Alkhouri et~al.(2025{\natexlab{a}})Alkhouri, Liang, Huang, Dai, Qu, Ravishankar, and Wang]{alkhouri2024sitcom}
Ismail Alkhouri, Shijun Liang, Cheng-Han Huang, Jimmy Dai, Qing Qu, Saiprasad Ravishankar, and Rongrong Wang.
\newblock Sitcom: Step-wise triple-consistent diffusion sampling for inverse problems.
\newblock \emph{ICML}, 2025{\natexlab{a}}.

\bibitem[Ulyanov et~al.(2018)Ulyanov, Vedaldi, and Lempitsky]{ulyanov2018deep}
Dmitry Ulyanov, Andrea Vedaldi, and Victor Lempitsky.
\newblock Deep image prior.
\newblock In \emph{Proceedings of the IEEE conference on computer vision and pattern recognition}, pages 9446--9454, 2018.

\bibitem[Saragadam et~al.(2023)Saragadam, LeJeune, Tan, Balakrishnan, Veeraraghavan, and Baraniuk]{saragadam2023wire}
Vishwanath Saragadam, Daniel LeJeune, Jasper Tan, Guha Balakrishnan, Ashok Veeraraghavan, and Richard~G Baraniuk.
\newblock Wire: Wavelet implicit neural representations.
\newblock In \emph{Proceedings of the IEEE/CVF Conference on Computer Vision and Pattern Recognition}, pages 18507--18516, 2023.

\bibitem[Sriram et~al.(2020)Sriram, Zbontar, Murrell, Defazio, Zitnick, Yakubova, Knoll, and Johnson]{sriram2020end}
Anuroop Sriram, Jure Zbontar, Tullie Murrell, Aaron Defazio, C~Lawrence Zitnick, Nafissa Yakubova, Florian Knoll, and Patricia Johnson.
\newblock End-to-end variational networks for accelerated mri reconstruction.
\newblock In \emph{Medical Image Computing and Computer Assisted Intervention--MICCAI 2020: 23rd International Conference, Lima, Peru, October 4--8, 2020, Proceedings, Part II 23}, pages 64--73. Springer, 2020.

\bibitem[Aggarwal et~al.(2018)Aggarwal, Mani, and Jacob]{aggarwal2018modl}
H.~K. Aggarwal, M.~P. Mani, and M.~Jacob.
\newblock Modl: Model-based deep learning architecture for inverse problems.
\newblock \emph{IEEE transactions on medical imaging}, 38\penalty0 (2):\penalty0 394--405, 2018.

\bibitem[Garwood et~al.(2017)Garwood, Recht, and White]{garwood2017advanced}
Elisabeth~R Garwood, Michael~P Recht, and Lawrence~M White.
\newblock Advanced imaging techniques in the knee: benefits and limitations of new rapid acquisition strategies for routine knee {MRI}.
\newblock \emph{American Journal of Roentgenology}, 209\penalty0 (3):\penalty0 552--560, 2017.

\bibitem[Zbontar et~al.(2018)Zbontar, Knoll, Sriram, Murrell, Huang, Muckley, Defazio, Stern, Johnson, Bruno, et~al.]{zbontar2018fastmri}
J.~Zbontar, F.~Knoll, A.~Sriram, T.~Murrell, Z.~Huang, M.~J. Muckley, A.~Defazio, R.~Stern, P.~Johnson, M.~Bruno, et~al.
\newblock {fastMRI: An open dataset and benchmarks for accelerated MRI}.
\newblock \emph{arXiv preprint arXiv:1811.08839}, 2018.

\bibitem[Chaudhari et~al.(2020)Chaudhari, Kogan, Pedoia, Majumdar, Gold, and Hargreaves]{chaudhari2020rapid}
Akshay~S Chaudhari, Feliks Kogan, Valentina Pedoia, Sharmila Majumdar, Garry~E Gold, and Brian~A Hargreaves.
\newblock Rapid knee {MRI} acquisition and analysis techniques for imaging osteoarthritis.
\newblock \emph{Journal of Magnetic Resonance Imaging}, 52\penalty0 (5):\penalty0 1321--1339, 2020.

\bibitem[Shafique et~al.(2025)Shafique, Liu, Schniter, and Ahmad]{shafique2024mri}
Muhammad Shafique, Sizhuo Liu, Philip Schniter, and Rizwan Ahmad.
\newblock Mri recovery with self-calibrated denoisers without fully-sampled data.
\newblock \emph{Magnetic Resonance Materials in Physics, Biology and Medicine}, 38:\penalty0 53--66, 2025.
\newblock \doi{10.1007/s10334-024-01207-1}.

\bibitem[Ronneberger et~al.(2015)Ronneberger, Fischer, and Brox]{ronneberger2015u}
Olaf Ronneberger, Philipp Fischer, and Thomas Brox.
\newblock U-net: Convolutional networks for biomedical image segmentation.
\newblock In \emph{Medical image computing and computer-assisted intervention--MICCAI 2015: 18th international conference, Munich, Germany, October 5-9, 2015, proceedings, part III 18}, pages 234--241. Springer, 2015.

\bibitem[Alkhouri et~al.(2024{\natexlab{b}})Alkhouri, Liang, Bell, Qu, Wang, and Ravishankar]{aSeqDIP}
Ismail~R. Alkhouri, Shijun Liang, Evan Bell, Qing Qu, Rongrong Wang, and Saiprasad Ravishankar.
\newblock Image reconstruction via autoencoding sequential deep image prior.
\newblock In \emph{Advances in Neural Information Processing Systems}, volume~37, pages 18988--19012, 2024{\natexlab{b}}.

\bibitem[Zhang et~al.(2022)Zhang, Xie, Gor, Chen, Zhou, and Metzler]{zhang2022metadipacceleratingdeepimage}
Kevin Zhang, Mingyang Xie, Maharshi Gor, Yi-Ting Chen, Yvonne Zhou, and Christopher~A. Metzler.
\newblock Metadip: Accelerating deep image prior with meta learning, 2022.
\newblock URL \url{https://arxiv.org/abs/2209.08452}.

\bibitem[Li et~al.(2023)Li, Wang, Zhuang, and Sun]{li2023deep}
Taihui Li, Hengkang Wang, Zhong Zhuang, and Ju~Sun.
\newblock Deep random projector: Accelerated deep image prior.
\newblock In \emph{Proceedings of the IEEE/CVF Conference on Computer Vision and Pattern Recognition}, pages 18176--18185, 2023.

\bibitem[Chung et~al.(2023)Chung, Kim, Mccann, Klasky, and Ye]{chung2022diffusion}
Hyungjin Chung, Jeongsol Kim, Michael~Thompson Mccann, Marc~Louis Klasky, and Jong~Chul Ye.
\newblock Diffusion posterior sampling for general noisy inverse problems.
\newblock In \emph{The Eleventh International Conference on Learning Representations}, 2023.

\bibitem[Daras et~al.(2024)Daras, Chung, Lai, Mitsufuji, Milanfar, Dimakis, Ye, and Delbracio]{diffusion_survey}
Giannis Daras, Hyungjin Chung, Chieh-Hsin Lai, Yuki Mitsufuji, Peyman Milanfar, Alexandros~G. Dimakis, Chul Ye, and Mauricio Delbracio.
\newblock A survey on diffusion models for inverse problems.
\newblock 2024.
\newblock URL \url{https://giannisdaras.github.io/publications/diffusion_survey.pdf}.

\bibitem[Aali et~al.(2025)Aali, Daras, Levac, Kumar, Dimakis, and Tamir]{aali2024ambient}
Asad Aali, Giannis Daras, Brett Levac, Sidharth Kumar, Alexandros~G Dimakis, and Jonathan~I Tamir.
\newblock Ambient diffusion posterior sampling: Solving inverse problems with diffusion models trained on corrupted data.
\newblock \emph{ICLR}, 2025.

\bibitem[Lehtinen et~al.(2018)Lehtinen, Munkberg, Hasselgren, Laine, Karras, Aittala, and Aila]{lehtinen2018noise2noise}
Jaakko Lehtinen, Jacob Munkberg, Jon Hasselgren, Samuli Laine, Tero Karras, Miika Aittala, and Timo Aila.
\newblock Noise2noise: Learning image restoration without clean data.
\newblock In \emph{International Conference on Machine Learning}, pages 2965--2974. PMLR, 2018.

\bibitem[Xia and Chakrabarti(2019)]{NEURIPS2019_0ed94223}
Zhihao Xia and Ayan Chakrabarti.
\newblock Training image estimators without image ground truth.
\newblock In H.~Wallach, H.~Larochelle, A.~Beygelzimer, F.~d\textquotesingle Alch\'{e}-Buc, E.~Fox, and R.~Garnett, editors, \emph{Advances in Neural Information Processing Systems}, volume~32. Curran Associates, Inc., 2019.
\newblock URL \url{https://proceedings.neurips.cc/paper_files/paper/2019/file/0ed9422357395a0d4879191c66f4faa2-Paper.pdf}.

\bibitem[Liu et~al.(2020)Liu, Sun, Eldeniz, Gan, An, and Kamilov]{9103213}
Jiaming Liu, Yu~Sun, Cihat Eldeniz, Weijie Gan, Hongyu An, and Ulugbek~S. Kamilov.
\newblock Rare: Image reconstruction using deep priors learned without groundtruth.
\newblock \emph{IEEE Journal of Selected Topics in Signal Processing}, 14\penalty0 (6):\penalty0 1088--1099, 2020.
\newblock \doi{10.1109/JSTSP.2020.2998402}.

\bibitem[Liang et~al.(2025)Liang, Bell, Qu, Wang, and Ravishankar]{liang2024analysis}
Shijun Liang, Evan Bell, Qing Qu, Rongrong Wang, and Saiprasad Ravishankar.
\newblock Analysis of deep image prior and exploiting self-guidance for image reconstruction.
\newblock \emph{IEEE Transactions on Computational Imaging}, 11:\penalty0 435--451, 2025.
\newblock ISSN 2333-9403.
\newblock \doi{10.1109/TCI.2025.3540706}.

\bibitem[Wang et~al.(2023)Wang, Li, Zhuang, Chen, Liang, and Sun]{wang2021early}
Hengkang Wang, Taihui Li, Zhong Zhuang, Tiancong Chen, Hengyue Liang, and Ju~Sun.
\newblock Early stopping for deep image prior.
\newblock \emph{Transactions on Machine Learning Research}, 2023.

\bibitem[Heckel et~al.(2019)]{heckel2019deep}
R~Heckel et~al.
\newblock Deep decoder: Concise image representations from untrained non-convolutional networks.
\newblock In \emph{International Conference on Learning Representations}, 2019.

\bibitem[Alkhouri et~al.(2025{\natexlab{b}})Alkhouri, Bell, Ghosh, Liang, Wang, and Ravishankar]{alkhouri2025understanding}
Ismail Alkhouri, Evan Bell, Avrajit Ghosh, Shijun Liang, Rongrong Wang, and Saiprasad Ravishankar.
\newblock Understanding untrained deep models for inverse problems: Algorithms and theory.
\newblock \emph{arXiv preprint arXiv:2502.18612}, 2025{\natexlab{b}}.

\bibitem[Vyas et~al.(2024)Vyas, Humayun, Dashpute, Baraniuk, Veeraraghavan, and Balakrishnan]{vyas2024learning}
Kushal~Kardam Vyas, Imtiaz Humayun, Aniket Dashpute, Richard Baraniuk, Ashok Veeraraghavan, and Guha Balakrishnan.
\newblock Learning transferable features for implicit neural representations.
\newblock \emph{Advances in Neural Information Processing Systems}, 37:\penalty0 42268--42291, 2024.

\bibitem[Sitzmann et~al.(2020)Sitzmann, Martel, Bergman, Lindell, and Wetzstein]{sitzmann2020implicit}
Vincent Sitzmann, Julien Martel, Alexander Bergman, David Lindell, and Gordon Wetzstein.
\newblock Implicit neural representations with periodic activation functions.
\newblock \emph{Advances in neural information processing systems}, 33:\penalty0 7462--7473, 2020.

\bibitem[Chen and Zhang(2019)]{chen2019learning}
Zhiqin Chen and Hao Zhang.
\newblock Learning implicit fields for generative shape modeling.
\newblock In \emph{Proceedings of the IEEE/CVF conference on computer vision and pattern recognition}, pages 5939--5948, 2019.

\bibitem[Bhattacharjee et~al.(2022)Bhattacharjee, Zhang, S{\"u}sstrunk, and Salzmann]{bhattacharjee2022mult}
Deblina Bhattacharjee, Tong Zhang, Sabine S{\"u}sstrunk, and Mathieu Salzmann.
\newblock Mult: An end-to-end multitask learning transformer.
\newblock In \emph{Proceedings of the IEEE/CVF conference on computer vision and pattern recognition}, pages 12031--12041, 2022.

\bibitem[Zhou et~al.(2025)Zhou, Bao, Wang, Zhang, Zhang, and Zhang]{zhou2025multi}
Jingxuan Zhou, Weidong Bao, Ji~Wang, Dayu Zhang, Xiongtao Zhang, and Yaohong Zhang.
\newblock Multi-task federated learning with encoder-decoder structure: Enabling collaborative learning across different tasks.
\newblock \emph{arXiv preprint arXiv:2504.09800}, 2025.

\bibitem[Zhang et~al.(2024)Zhang, Lu, Alkhouri, Ravishankar, Song, and Qu]{Zhang_2024_CVPR}
Huijie Zhang, Yifu Lu, Ismail Alkhouri, Saiprasad Ravishankar, Dogyoon Song, and Qing Qu.
\newblock Improving training efficiency of diffusion models via multi-stage framework and tailored multi-decoder architecture.
\newblock In \emph{Proceedings of the IEEE/CVF Conference on Computer Vision and Pattern Recognition (CVPR)}, pages 7372--7381, June 2024.

\bibitem[Zhao et~al.(2020)Zhao, Zhao, and Gan]{zhao2020reference}
Di~Zhao, Feng Zhao, and Yongjin Gan.
\newblock Reference-driven compressed sensing mr image reconstruction using deep convolutional neural networks without pre-training.
\newblock \emph{Sensors}, 20\penalty0 (1):\penalty0 308, 2020.

\bibitem[Kingma and Ba(2014)]{kingma2014adam}
Diederik~P. Kingma and Jimmy Ba.
\newblock Adam: A method for stochastic optimization.
\newblock \emph{arXiv preprint arXiv:1412.6980}, 2014.

\bibitem[Tamir et~al.(2016)Tamir, Ong, Cheng, Uecker, and Lustig]{tamir2016generalized}
J.~I. Tamir, F.~Ong, J.~Y. Cheng, M.~Uecker, and M.~Lustig.
\newblock Generalized magnetic resonance image reconstruction using the berkeley advanced reconstruction toolbox.
\newblock In \emph{ISMRM Workshop on Data Sampling \& Image Reconstruction, Sedona, AZ}, 2016.

\bibitem[Karras et~al.(2019)Karras, Laine, and Aila]{karras2019style}
Tero Karras, Samuli Laine, and Timo Aila.
\newblock A style-based generator architecture for generative adversarial networks.
\newblock In \emph{Proceedings of the IEEE/CVF conference on computer vision and pattern recognition}, pages 4401--4410, 2019.

\bibitem[Newell et~al.(2016)Newell, Yang, and Deng]{newell2016stacked}
Alejandro Newell, Kaiyu Yang, and Jia Deng.
\newblock Stacked hourglass networks for human pose estimation.
\newblock In \emph{Computer Vision--ECCV 2016: 14th European Conference, Amsterdam, The Netherlands, October 11-14, 2016, Proceedings, Part VIII 14}, pages 483--499. Springer, 2016.

\bibitem[Wang et~al.(2004)Wang, Bovik, Sheikh, and Simoncelli]{wang2004image}
Zhou Wang, Alan~C Bovik, Hamid~R Sheikh, and Eero~P Simoncelli.
\newblock Image quality assessment: from error visibility to structural similarity.
\newblock \emph{IEEE transactions on image processing}, 13\penalty0 (4):\penalty0 600--612, 2004.

\bibitem[Zhang et~al.(2018)Zhang, Isola, Efros, Shechtman, and Wang]{zhang2018unreasonable}
Richard Zhang, Phillip Isola, Alexei~A Efros, Eli Shechtman, and Oliver Wang.
\newblock The unreasonable effectiveness of deep features as a perceptual metric.
\newblock In \emph{Proceedings of the IEEE conference on computer vision and pattern recognition}, pages 586--595, 2018.

\bibitem[Deng et~al.(2009)Deng, Dong, Socher, Li, Li, and Fei-Fei]{deng2009imagenet}
Jia Deng, Wei Dong, Richard Socher, Li-Jia Li, Kai Li, and Li~Fei-Fei.
\newblock Imagenet: A large-scale hierarchical image database.
\newblock In \emph{2009 IEEE conference on computer vision and pattern recognition}, pages 248--255. Ieee, 2009.

\end{thebibliography}

\newpage
\appendix
\onecolumn
\par\noindent\rule{\textwidth}{1pt}
\begin{center}
{\Large \bf Appendix}
\end{center}
\vspace{-0.1in}
\par\noindent\rule{\textwidth}{1pt}
\appendix

In the Appendix, we first present two experiments to evaluate the out-of-distribution performance of \our (Appendix~\ref{sec: appen ood eval}). Appendix~\ref{sec: ablation frzen encoder vs. initialized one} examines the impact of freezing the encoder weights at test time, followed by a study on robustness to noise overfitting (Appendix~\ref{sec: ablation lambda}). Appendix~\ref{sec: appen learned features for uGoDiT} investigates the learned features in an MRI-trained \our model. This is followed by ablation studies on the number of convolutional layers (Appendix~\ref{sec: append different arch}), the number of training images $M$ (Appendix~\ref{sec: append different number of training M}), and the choice of $N$ and $K$ (Appendix~\ref{sec: appen NK}). Limitations and future directions are discussed in Appendix~\ref{sec: limitation and future}. Appendix~\ref{sec: append imp details} provides implementation details of the baselines, followed by a set of figures for additional visualizations (Appendix~\ref{sec: appen more visual}).

\section{Out-of-distribution evaluation}\label{sec: appen ood eval}

\subsection{MRI}

In this section, we investigate the out-of-distribution (OOD) capabilities of \our for the task of MRI. Specifically, we evaluate the encoder of \our-6 (which we will refer to as \our-OOD) trained on knee scans with $20$ brain MRI test scans randomly selected from the fastMRI brain dataset. 

\begin{figure*}[htp]
\centering
\includegraphics[width=14.1cm]{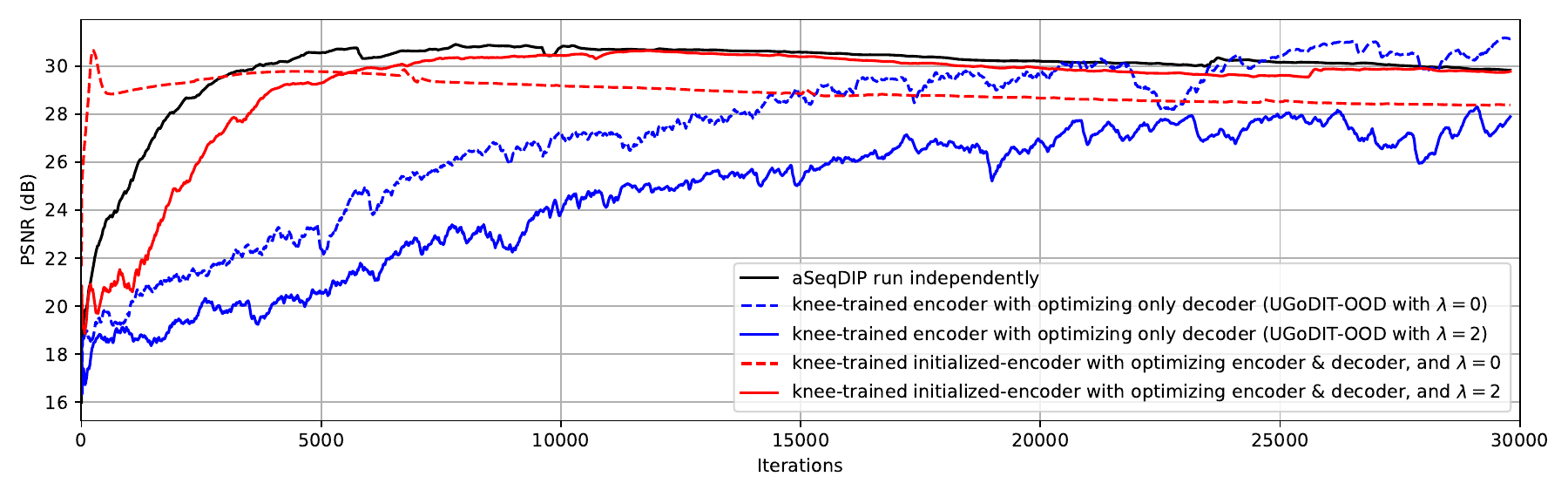}
\caption{\small{Average PSNR of $20$ MRI brain test scans using a knee-trained (with $M=6$) \our-OOD (blue) vs. the case where at test-time, we use the parameters of the pre-trained encoder to only initialize the test-time encoder (red). Then, we optimize over both the encoder and decoder. Running aSeqDIP independently is also included (black). Iterations in the x-axis correspond to $NK=30000$. }}
\label{fig: ood psnr curves}
\end{figure*}
\begin{figure*}[htp]
\centering
\includegraphics[width=14cm]{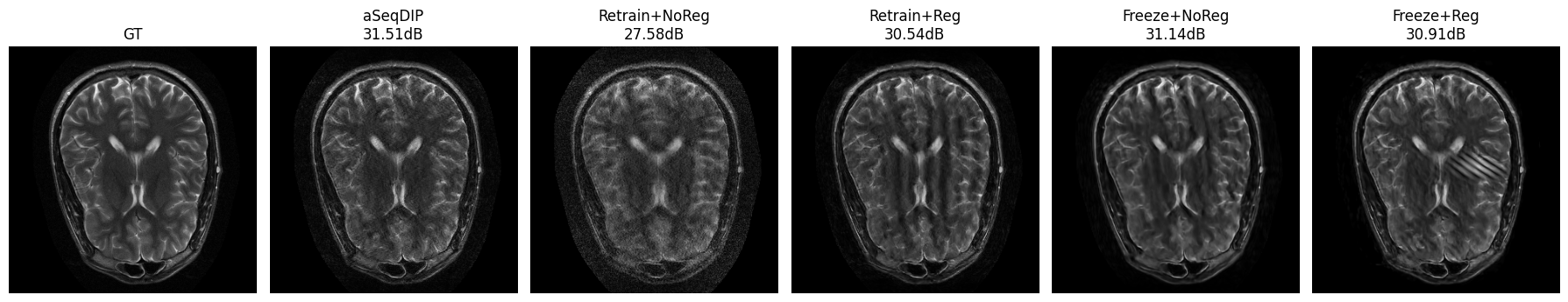}
\caption{\small{Brain reconstructed images for the OOD experiment. The titles of the third to fifth reconstructed images correspond to the dashed blue (\our-OOD with $\lambda = 0$), solid blue (\our-OOD with $\lambda = 2$), dashed red, and solid red curves, respectively, in the legend of Figure~\ref{fig: ood psnr curves}. PSNR values are given on the top of each reconstructed image.}}
\label{fig: visual brain OOD}
\end{figure*}
The average PSNR curves are shown in Figure~\ref{fig: ood psnr curves}. As observed, when compared to aSeqDIP—which is run independently from scratch on each test scan (black curve)—the encoder trained on knee images under-performs and converges very slowly on brain scans. 

We hypothesize that the autoencoding term and the input updates in \our—which implicitly enforce consistency between encoder-decoder input/output pairs—lead to the learning of features that are overly tailored to the training distribution. As a result, when the test distribution differs significantly, these learned features make the convergence slower.

More specifically, when training and test images come from the same distribution (i.e., anatomy), the input updates and autoencoding term help reconstruction, as they promote alignment of shared low-frequency structures. However, under distribution shift (i.e., when the training and testing anatomies are different), these mechanisms degrade performance, since training and test images no longer share similar low-frequency content.

To support this hypothesis, Figure~\ref{fig: ood psnr curves} includes two additional cases:
(\textit{i}) using the knee-trained encoder while optimizing only the decoder and setting $\lambda = 0$ (i.e., no autoencoding regularization in the first term of Eq.~\eqref{eqn: ours testing updates}; shown as the dashed blue curve); and
(\textit{ii}) initializing the encoder with knee-trained weights and optimizing both the encoder and decoder, with $\lambda \in {0, 2}$ (shown in red curves). 

As observed, when $\lambda=0$ (i.e., dashed curves), the results could be better than when $\lambda \neq 0$ as the input will have less impact. Furthermore, when we compare \our-OOD with only initializing the encoder (the red curves), the results are further improved which indicate that pre-trained weights on knees have less impact when only initialization is used. This is even more evident in the case when $\lambda = 0$ with initialization only (i.e., red dashed curve). 

The visualizations in Figure~\ref{fig: visual brain OOD} (which are recorded at iteration 30000) show that the all the considered cases would converge to between 27.58 dB to 31.51 dB. 

\subsection{Super Resolution}\label{sec: append ood SR}

In this section, we investigate the OOD capabilities of \our on natural images for the task of super resolution (SR). To this end, we test the encoder of \our (trained with $M=6$ images from the FFHQ dataset) with $20$ degraded images randomly selected from the ImageNet dataset \citep{deng2009imagenet}. 

\begin{figure*}[htp]
\centering
\includegraphics[width=14.1cm]{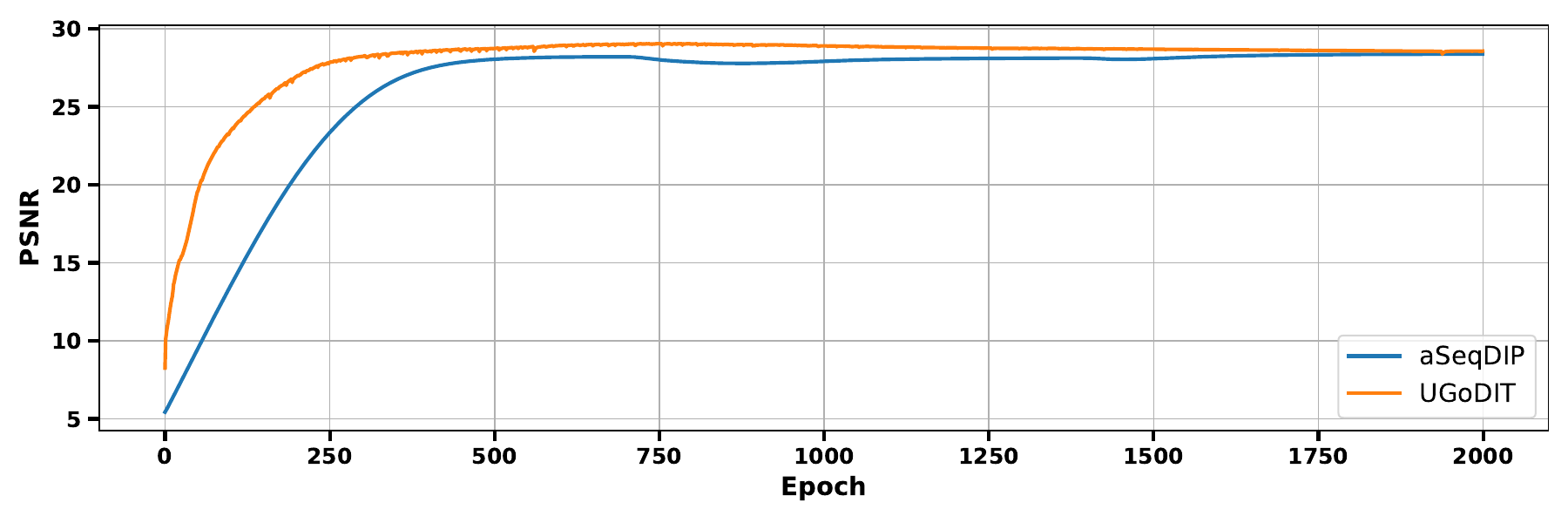}
\caption{\small{Average PSNR of $20$ test images from the \textbf{ImageNet} dataset of \our trained with $M=6$ \textbf{FFHQ} degraded images. Iterations in the x-axis correspond to $NK=20000$. }}
\label{fig: good SR}
\end{figure*}

\begin{figure*}[htbp]
\centering

\begin{tabular}{cccc}
    \textbf{Ground Truth} & 
    \textbf{Input}
    &\textbf{aSeq DIP} & \textbf{\our} \\
    
    \includegraphics[width=.22\linewidth,valign=t]{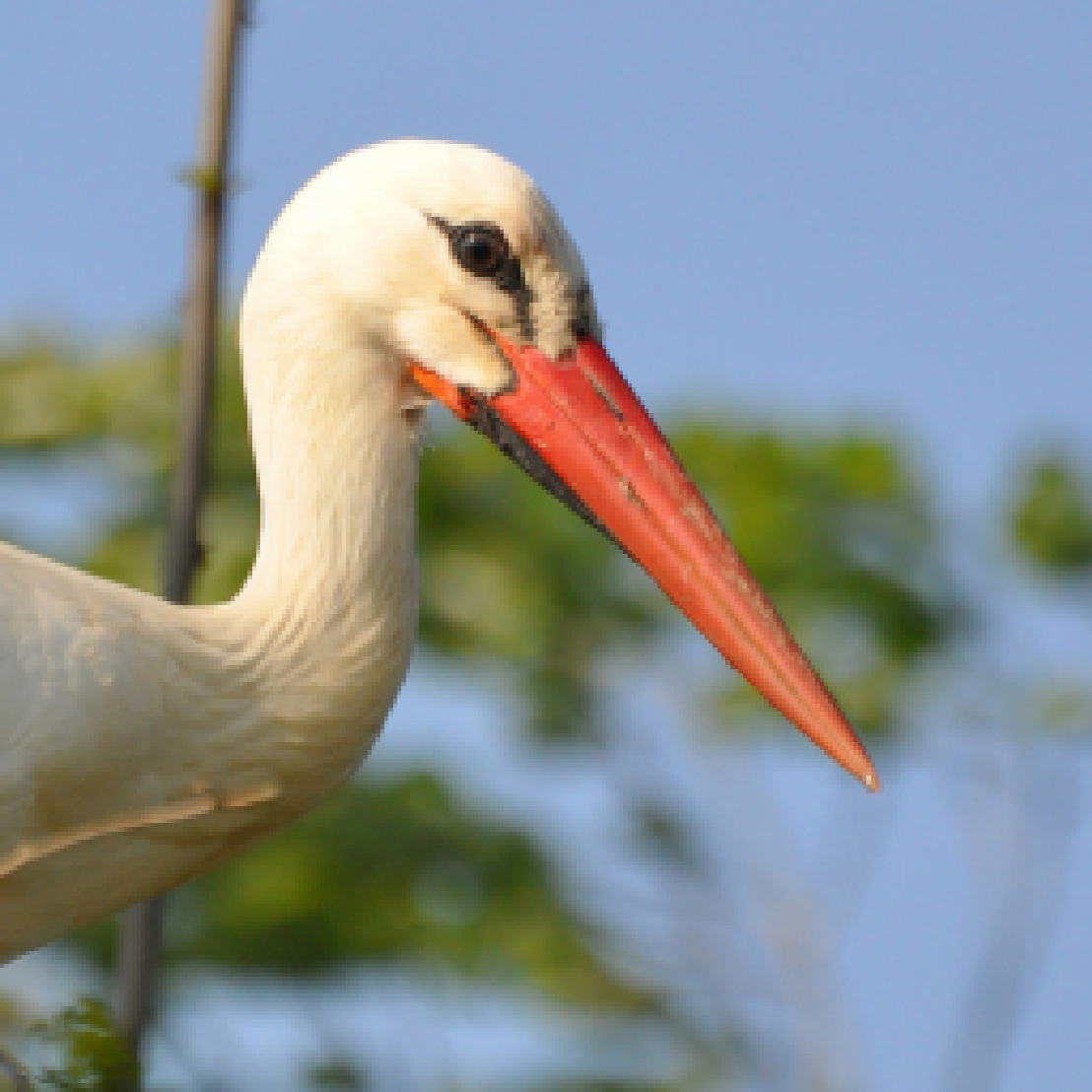} &
    \includegraphics[width=.22\linewidth,valign=t]{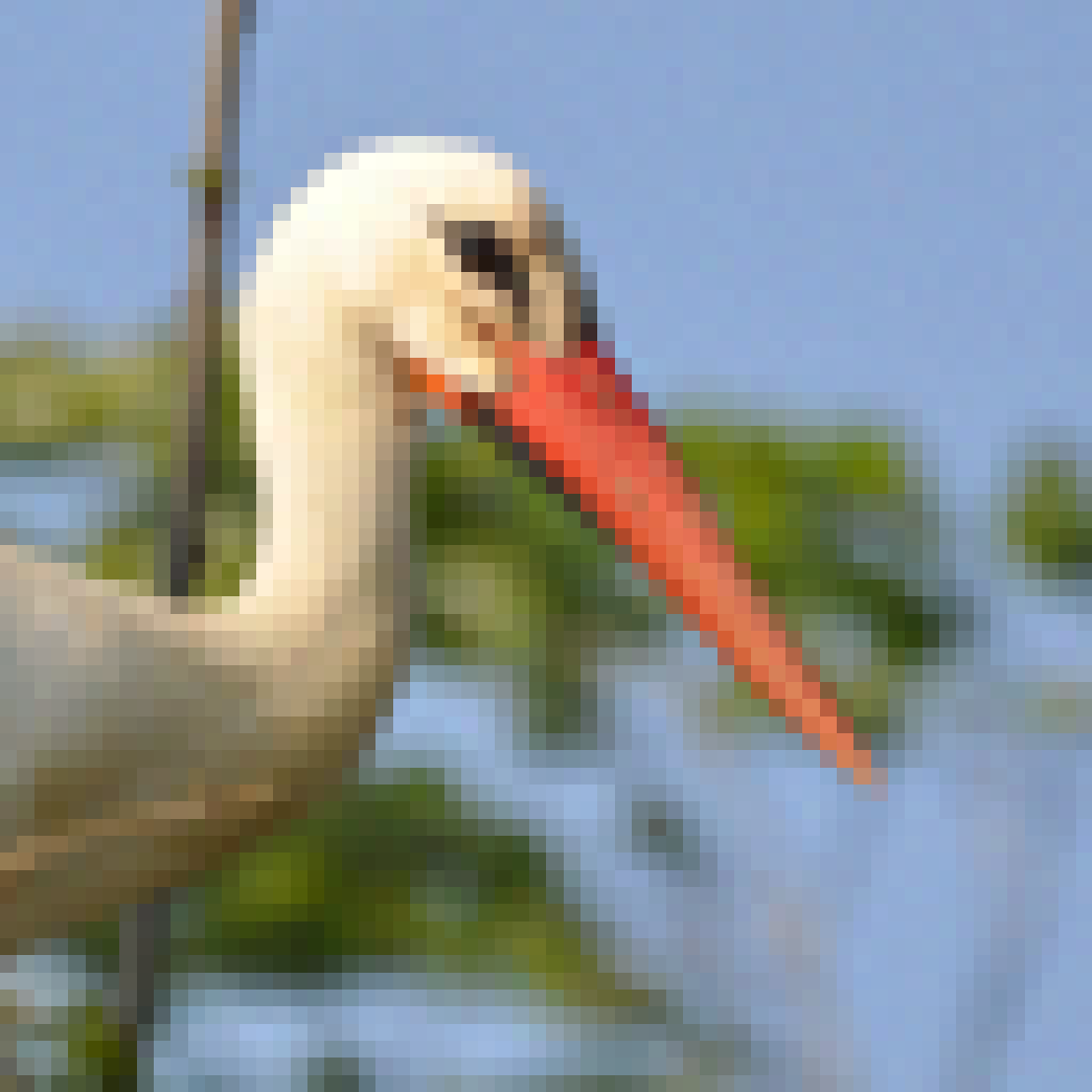} &
    \includegraphics[width=.22\linewidth,valign=t]{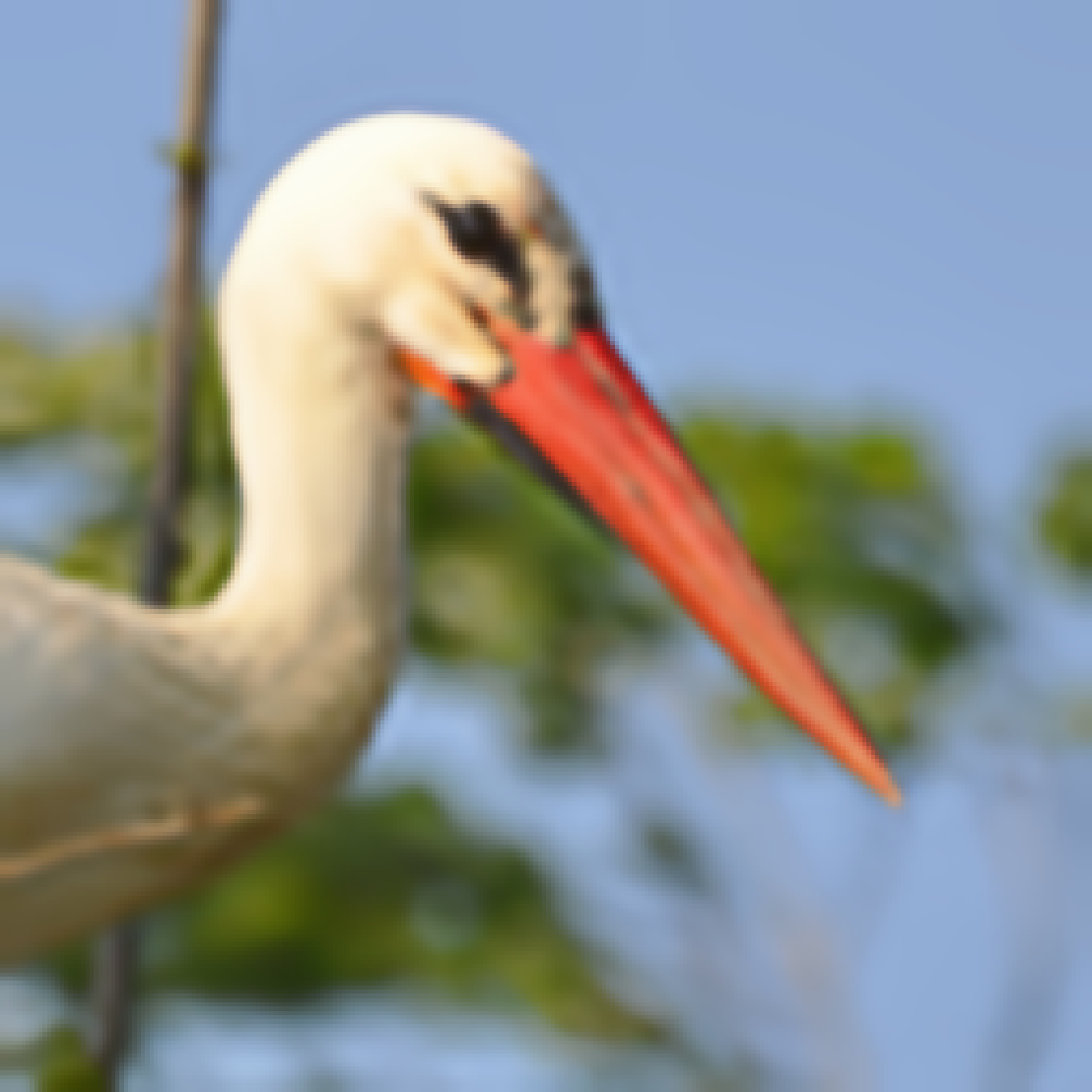} &
    \includegraphics[width=.22\linewidth,valign=t]{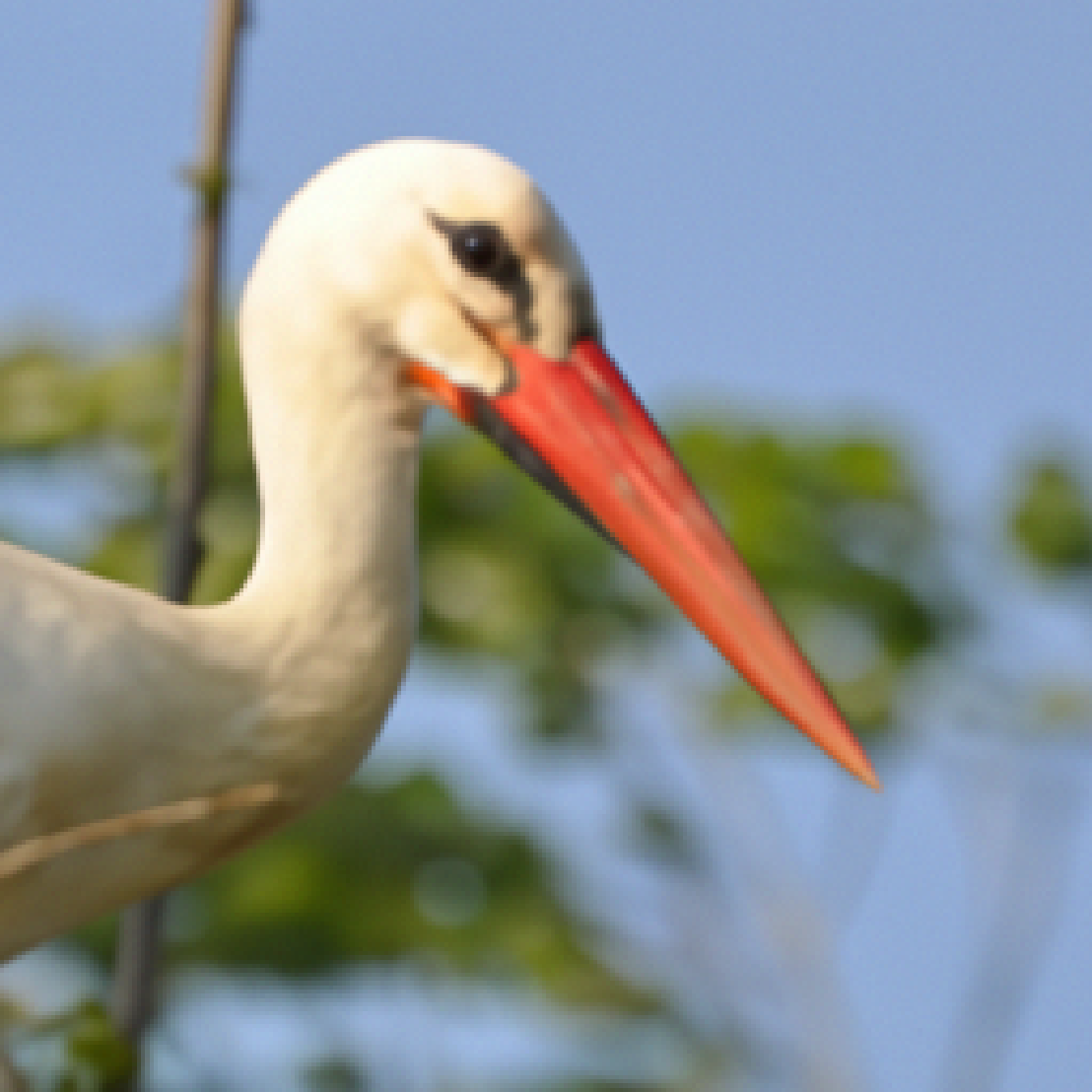} \\
    
    \scriptsize{PSNR = $\infty$ dB} &
    \scriptsize{PSNR = 21.12 dB} &\scriptsize{PSNR = 29.78 dB} & \scriptsize{PSNR = 29.95 dB} \
\end{tabular}
\caption{\small{Ground-truth, degraded, and restored images for the OOD experiment on the task of super resolution, where \our was trained on 6 images of faces from the FFHQ dataset and tested on the ImageNet degraded image in the second column. As observed, the testing set is not a human face as is the case for the training images in FFHQ.  }}
\label{fig:denoised_imgs_imagenet}
\end{figure*}
\begin{figure*}
    
\begin{tabular}{cccc}
    \textbf{Ground Truth} & 
    \textbf{Input}
    &\textbf{aSeq DIP} & \textbf{\our} \\
    
    \includegraphics[width=.22\linewidth,valign=t]{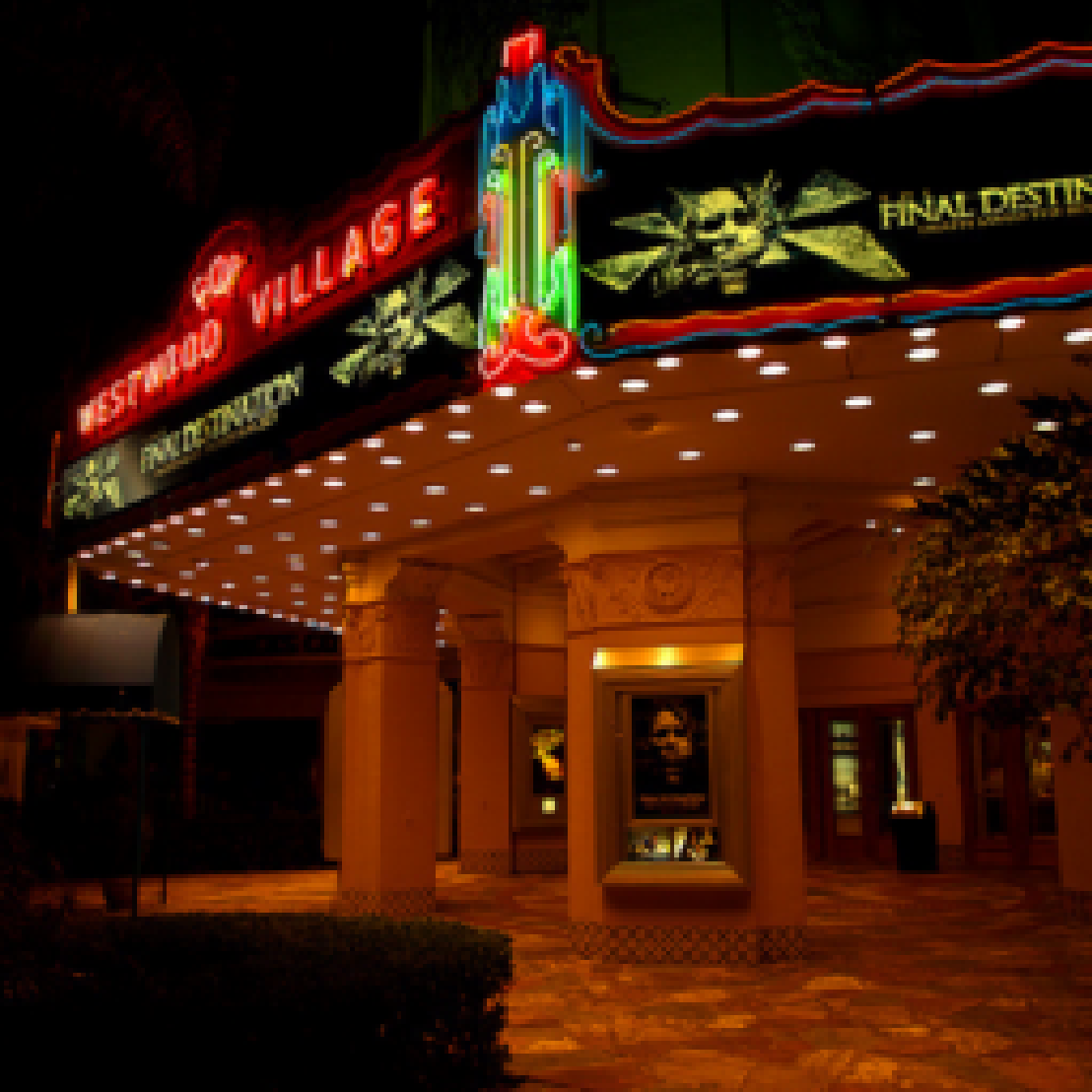} &
    \includegraphics[width=.22\linewidth,valign=t]{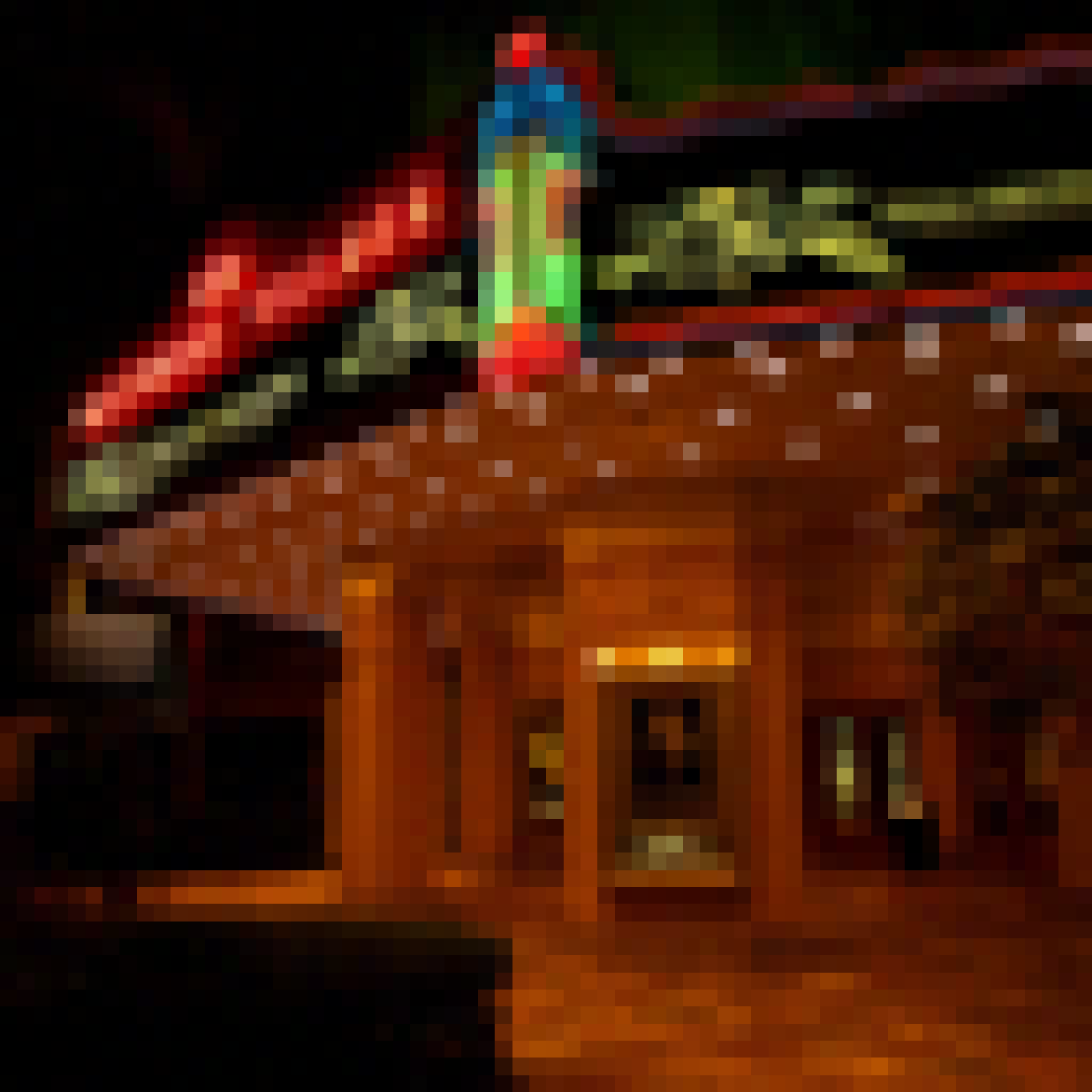} &
    \includegraphics[width=.22\linewidth,valign=t]{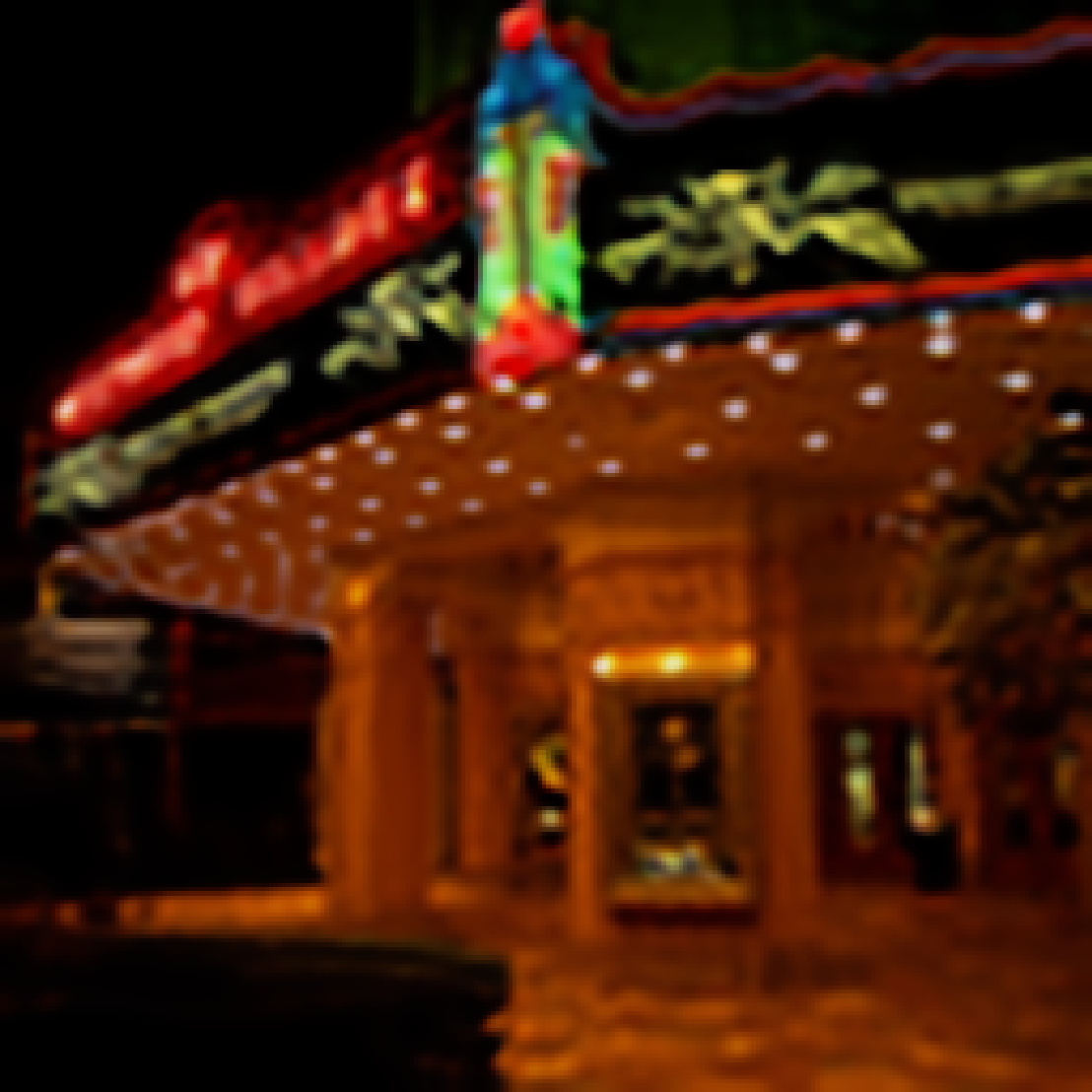} &
    \includegraphics[width=.22\linewidth,valign=t]{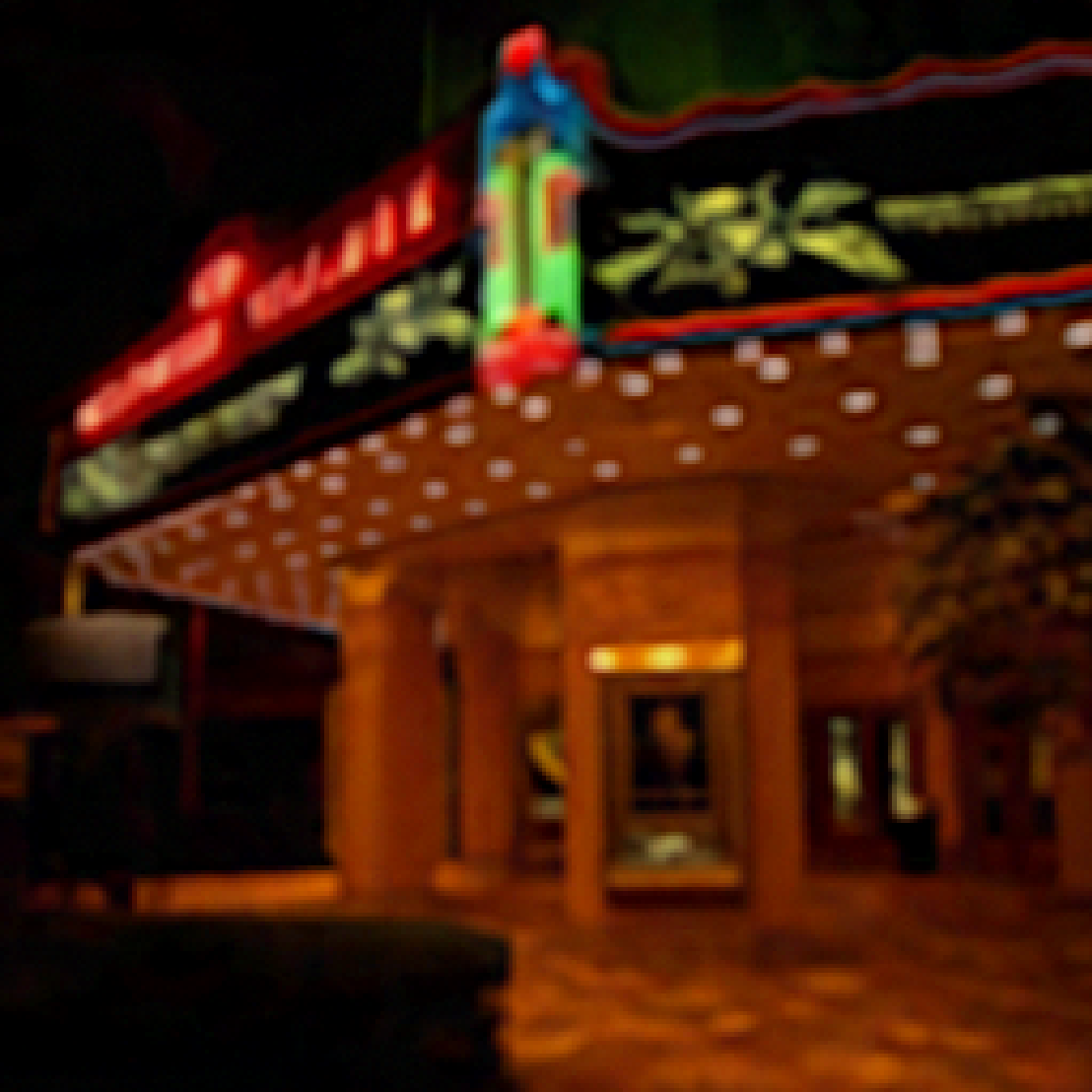} \\
    
    \scriptsize{PSNR = $\infty$ dB} &
    \scriptsize{PSNR = 19.32 dB} &\scriptsize{PSNR = 24.38 dB} & \scriptsize{PSNR = 24.89 dB} \
\end{tabular}
\caption{\small{Ground-truth, degraded, and restored images for the OOD experiment on the task of super resolution, where \our was trained on 6 images of faces from the FFHQ dataset and tested on the ImageNet degraded image in the second column. As observed, the testing set is not a human face as is the case for the training images in FFHQ.}}
\label{fig:denoised_imgs_imagenet2}
\end{figure*}

For SR, We observe that \our-OOD slightly achieves better PSNR that eventually converges to similar PSNR values. However, unlike the MRI case, \our-OOD converges faster than aSeqDIP. This can also be observed in the restored images of Figure~\ref{fig:denoised_imgs_imagenet} and Figure~\ref{fig:denoised_imgs_imagenet2}. We attribute \our’s ability to generalize well to learning multi-frequency features under OOD scenario for the natural image. 


\section{Impact of freezing the test-time encoder in \our}
\label{sec: ablation frzen encoder vs. initialized one}

Here, we show the impact of freezing the weights of the pre-trained encoder at test time. More specifically, given a pre-trained encoder with weight $\hat{\phi}$, we compare \our at test time with an alternative approach that initializes the test-time encoder with $\hat{\phi}$ and then optimizes both the encoder and decoder. We note that this is a similar setting to how the pre-trained MLP weights were used in STRAINER \cite{vyas2024learning}. 
\begin{figure*}[htp]
\centering
\includegraphics[width=14.1cm]{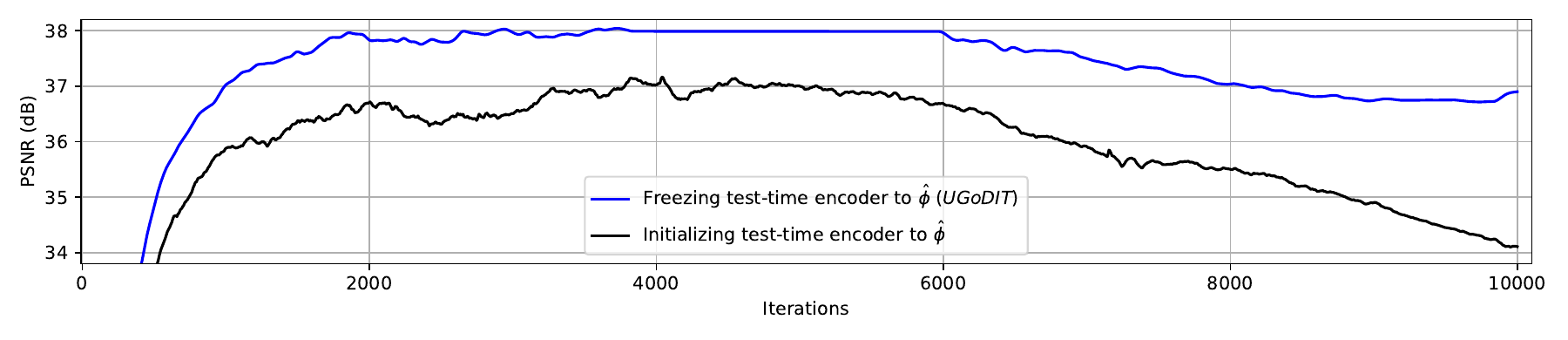}
\caption{\small{Average PSNR of $20$ MRI test scans of \our-6 vs. the case where we use the parameters of the pre-trained encoder to only initialize the test-time encoder. Then, the optimization takes place over both the encoder and decoder. Iterations in the x-axis correspond to $NK=10000$. For both cases, we use $\lambda=2$.}}
\label{fig: frozen_vs_initialized encoder}
\end{figure*}

Figure~\ref{fig: frozen_vs_initialized encoder} presents the average PSNR over $20$ MRI scans. As shown, freezing the encoder weights yields better results in terms of both PSNR (around 1dB difference) and convergence speed. Specifically, \our approaches the peak performance in approximately 2000 iterations. Moreover, \our requires optimizing fewer test-time parameters compared to the approach that initializes the encoder and optimizes both the encoder and decoder.

\section{Robustness to noise overfitting and the impact of the regularization parameter $\lambda$}\label{sec: ablation lambda}

In this section, we demonstrate how \our is robust to noise overfitting, typically encountered in DIP-based methods. To this end, for the task of MRI reconstruction, we train \our with $M=6$ knee scans across different values of the regularization parameter: $\lambda\in \{0.1, 1, 2, 10\}$. During testing, we use $20$ scans. 

\begin{figure*}[htp]
\centering
\includegraphics[width=14.1cm]{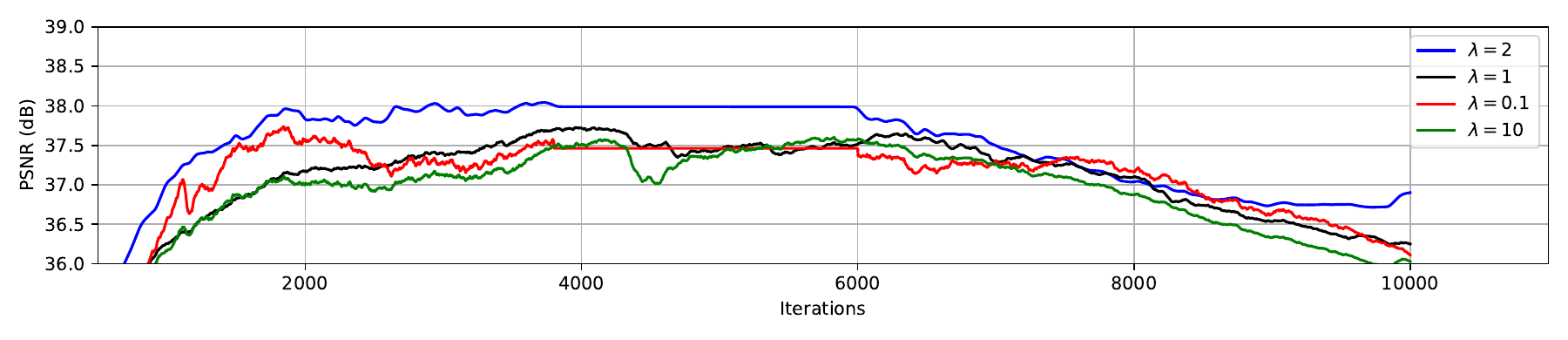}
\caption{\small{Average PSNR of $20$ MRI test scans of \our-6 using different values of the regularization parameter $\lambda$ at training and testing. Iterations in the x-axis correspond to $NK=10000$. }}
\label{fig: impact of lambda}
\end{figure*}
%

Figure~\ref{fig: impact of lambda} presents the average PSNR during testing. As observed, even when we run our method for 10000 iterations, the PSNR drops by approximately less than 1.5 dB. This indicates that noise overfitting takes place but is minimal given the large number of iterations. 

Additionally, the case of $\lambda=2$ returns the highest PSNR and has the least noise overfitting (as indicated by the drop in PSNR). We also observe that the difference in PSNR across different $\lambda$'s is approximately 1 dB. This indicates that the performance of \our is not highly sensitive to the choice of the regularization parameter. In other words, if different values of $\lambda$ are used (between 0.1 and 10), the PSNR value does not change significantly.


\section{A study on the learned features in \our}\label{sec: appen learned features for uGoDiT}

To better understand the representations learned by our shared encoder, we examine the feature maps from its first few convolutional layers across two different approaches - one with the proposed \our framework and the other with the shared decoder (called single E-D) framework. To analyze the frequency content of the learned features, we applied a 2D Fourier transform to each channel of the encoder’s layer and visualized the corresponding magnitude spectra. We computed a low-frequency (LF) magnitude ratio, which is the proportion of spectral energy contained within a central region of the frequency domain. This helps us to quantify, to how much extent, low frequency features dominate the learned representation. The LF magnitude ratio can be defined as

\begin{equation}
\text{LF ratio} = \frac{\sum_{c} \sum_{(u,v) \in \mathcal{C}} |\hat{x}_c[u,v]|}{\sum_{c} \sum_{(u,v)} |\hat{x}_c[u,v]|}
\end{equation}

where $\hat{x}_c[u,v]$ is the 2D Fourier transform of the feature map at channel $c$, and
$\mathcal{C}$ is a centered square region in frequency space corresponding to low frequencies.

\begin{table}[h]
\centering
\begin{tabular}{l|cc|cc}
\toprule
\textbf{Metric} & \multicolumn{2}{c|}{\textbf{Slice 1}} & \multicolumn{2}{c}{\textbf{Slice 2}} \\
                & \textbf{UGoDIT} & \textbf{Single E-D} & \textbf{UGoDIT} & \textbf{Single E-D} \\
\midrule
First Layer     & 0.463           & \textbf{0.525}      & 0.363           & \textbf{0.417} \\
Second Layer    & \textbf{0.341}  & 0.318               & 0.302           & \textbf{0.325} \\
Third Layer     & 0.235             & \textbf{0.337}                  & 0.233  & \textbf{0.357} \\
Fourth Layer  & 0.166             & \textbf{0.232}                  & 0.207  & \textbf{0.303} \\
Decoder Output  & 0.475           & \textbf{0.500}      & 0.503           & \textbf{0.534} \\
PSNR (dB)       & \textbf{36.61}  & 33.98    & \textbf{35.92}         & 34.82 \\
\bottomrule
\end{tabular}
\caption{Comparison of low-frequency (LF) magnitude ratios and PSNR for UGoDIT and the Single E-D approach for MRI reconstruction at 4$\times$ acceleration factor. The first five rows report LF magnitude ratios from encoder layers and the decoder output; the last row reports PSNR (in dB). 
For reference, the LF magnitude ratios for the ground truth images are 0.412 (Slice~1) and 0.439 (Slice~2).
}

\label{tab:lf_energy_ratios}
\end{table}

Table~\ref{tab:lf_energy_ratios} summarizes the low-frequency (LF) magnitude ratios computed for 2 images across four encoder layers and the decoder output, along with PSNR (in dB). These results are reported for both the \our and Single E-D (one encoder, one decoder) approaches that were used for MRI reconstruction at 4$\times$ acceleration factor. Across both test slices, Single E-D exhibits higher LF magnitude ratios in most layers, including the decoder output, indicating a stronger low-frequency bias. \our achieves higher PSNR values for both slices that have only moderate low-frequency content. Its feature representations better capture high-frequency details in the image and yield better final reconstructed images than the single E-D case, which shows a more pronounced low-frequency bias from shared training.

\section{Impact of the number of convolutional layers in the architecture of \our}\label{sec: append different arch}

In this section, we analyze the impact of varying the depth of the convolutional blocks in \our. Specifically, we evaluate five different configurations of the Skip-net architecture \citep{newell2016stacked} on the super-resolution task performance, each with a different number of upsampling and downsampling convolutional layers, and report the resulting PSNR values. 

\begin{figure*}[htbp]
    \centering
    \includegraphics[width=0.45\textwidth]{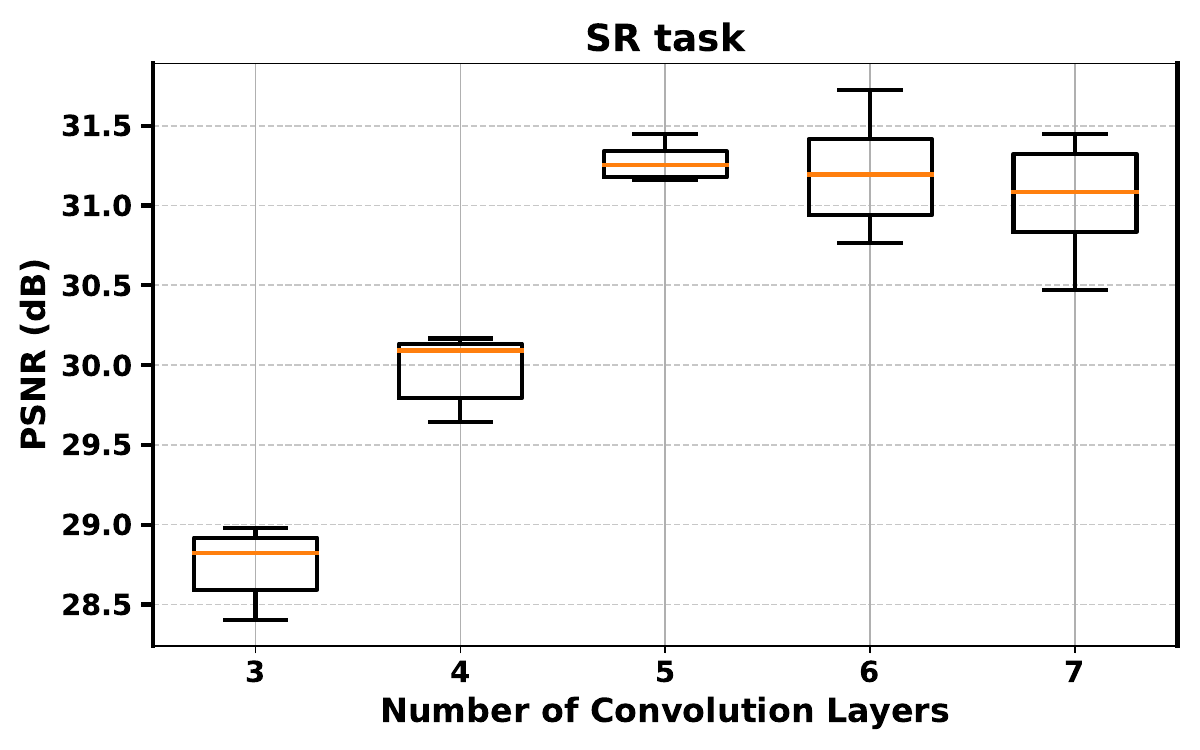}
    \caption{\small{Ablation study in terms of PSNR as a function of the number of Convolution layers in the Skip Net for the Super Resolution task, averaged over $20$ test images. The x-axis indicates the number of convolutional layers in each of the upsampling and downsampling parts of the network architecture.}}
    \label{fig: Ablation on convolution layer}
\end{figure*}

As shown in Figure~\ref{fig: Ablation on convolution layer}, the configuration with five convolutional layers in both the encoder and decoder produces the highest and most stable PSNR. Increasing the number of layers beyond five yields no further improvement, while reducing the layer count leads to a noticeable degradation in PSNR.

%

\section{Impact of the number of training images ($M$) in \our}\label{sec: append different number of training M}

\begin{figure*}[htbp]
    \centering
    \includegraphics[width=0.5\textwidth]{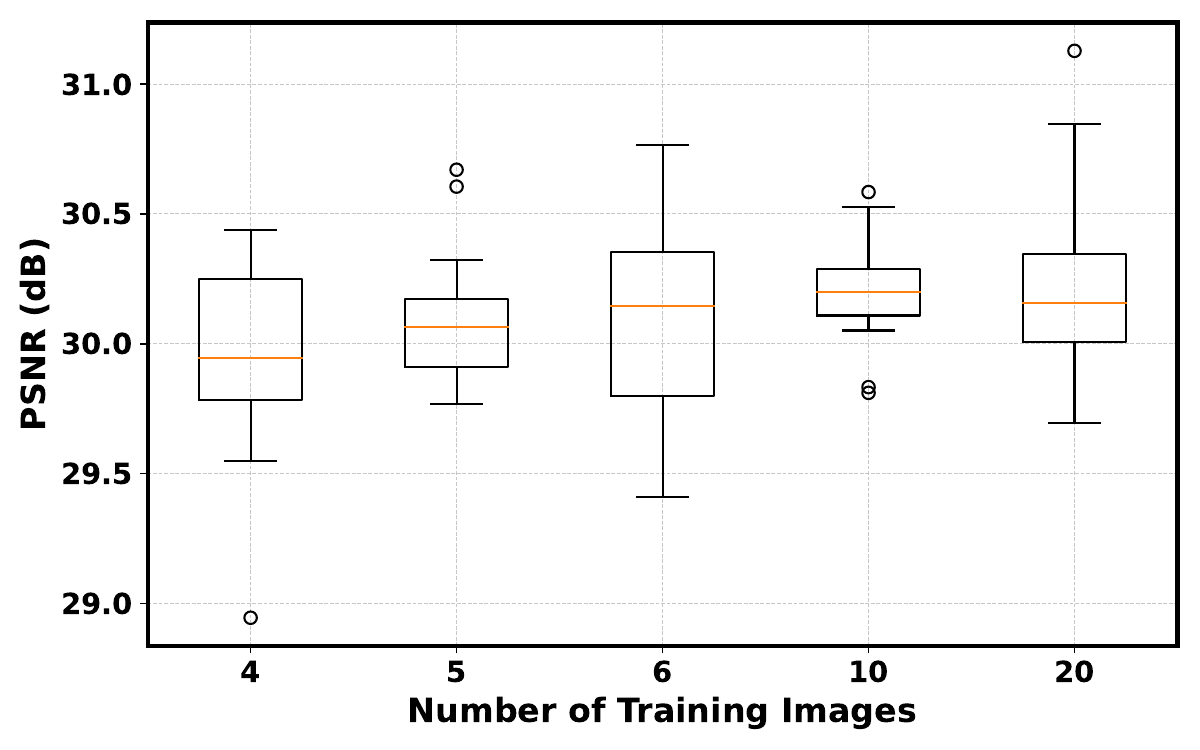}
    \caption{\small{Ablation study on the number of training images, $M$, as given in Algorithm~\ref{alg: training} and Algorithm~\ref{alg: testing}. Results are averaged over $20$ images for the task of NDB.}}
    \label{fig: Ablation on M}
\end{figure*}

In this section, we investigate how the size of the training set influences \our’s performance. To this end, we consider the non-linear deblurring task and train the shared encoder using five different FFHQ-derived subsets \citep{karras2019style}, and evaluate each on the same deblurring benchmark. As shown in Figure~\ref{fig: Ablation on M}, performance steadily improves as we increase the number of training images, peaking at six images—beyond which additional data yields negligible gains.

\section{Ablation study on $N$ and $K$ in \our}
\label{sec: appen NK}

\begin{figure*}[htbp]
    \centering
    \includegraphics[width=\textwidth]{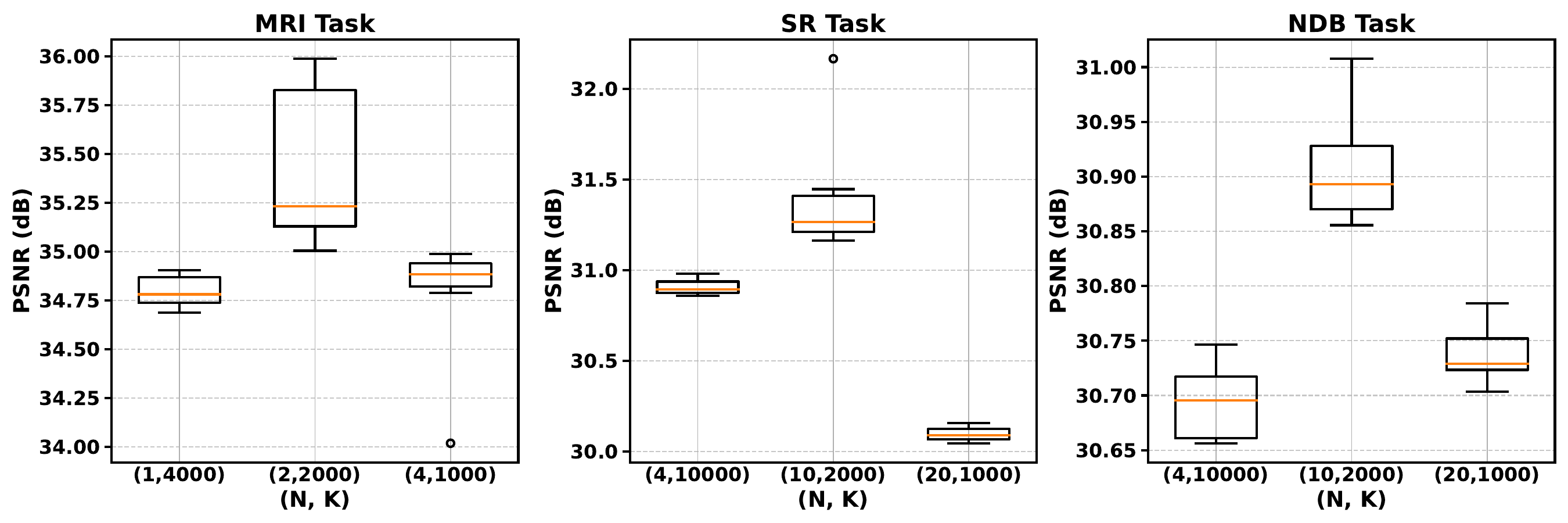}
    \caption{\small{Ablation study on the number of gradient updates, $N$, for every inputs updates, $K$, as given in Algorithm~\ref{alg: training} and Algorithm~\ref{alg: testing}. Results are given as the average of 20 images for the tasks of MRI (\textit{left}), SR (\textit{middle}), and NDB (\textit{right}).}}
    \label{fig: Ablation on NK}
\end{figure*}
Here, we conduct an ablation study on the number of gradient updates, $N$, for each inputs updates (i.e., $K$) in the training (Algorithm~\ref{alg: training}) and testing (Algorithm~\ref{alg: testing}) of \our. 

The results are given presented in Figure~\ref{fig: Ablation on NK}, where we report the average PSNR values across the tasks of MRI (considering combinations of $(N,K)$ as $(1,4000)$, $(2,2000)$, and $(4,1000)$), SP, and NDB (considering combinations of $(N,K)$ as $(4,5000)$, $(10,2000)$, and $(20,1000)$). 

The results show that the combination we selected (for each task), on average, returns the best results.

\section{Limitations and future work}\label{sec: limitation and future}

In this paper, we assume full access to the forward model and focus on 2D image recovery problems. In future work, we aim to extend \our to 3D image reconstruction and to the blind setting, where the goal is to jointly reconstruct the image and estimate the forward operator.

Additionally, our out-of-distribution (OOD) performance varies across tasks, as discussed in Appendix~\ref{sec: appen ood eval}. Specifically, for MRI, our OOD performance is lower than that of independently running aSeqDIP. In contrast, for super-resolution (SR), we observe improved acceleration while slightly outperforming aSeqDIP in terms of PSNR. As such, for future work, we will explore combining our method with test-time adaptation approaches to enhance performance on OOD test cases.

\section{Implementation details for baselines}\label{sec: append imp details}


\paragraph{Natural image restoration baselines:} For Diffusion-based approaches, we use the authors’ official code for each method: DPS with the DDPM solver using $1,000$ diffusion steps; SITCOM with \(N=20\) diffusion steps where \(K=20\) optimization steps are used for each. These methods use the pre-trained DM (trained on FFHQ) from DPS. For STRAINER, the Shared INR architecture consists of a single shared encoder—composed of six sine-activated layers—and \(k\) decoder heads, where \(k\) is the number of images to be learned. During training, we jointly train the encoder and all decoders by minimizing the sum of per-decoder mean squared error (MSE) losses using ADAM \citep{kingma2014adam} with a learning rate of $0.001$. At test time, we instantiate a new single-decoder INR initialized with the pre-trained encoder weights and train only both using a data-consistency loss tailored to the relevant forward operator.

\paragraph{MRI reconstruction baselines:}
For DM-based methods, the pre-trained model is from DDS \cite{ye2024decomposed}. In DDS, we set \(N=100\) diffusion steps, whereas SITCOM-MRI uses \(N=50\) diffusion steps, \(K=10\) optimization steps, and $\lambda = 0$ (no regularization). For the supervised mode baseline, we adopt the default VarNet configuration, comprising 12 cascades of unrolled reconstruction steps. Optimization is performed with Adam (learning rate \(3\times10^{-4}\)) on the FastMRI knee dataset \citep{zbontar2018fastmri}, which contains 973 volumes. From each volume, we discard the first and last five slices, then randomly sample 8,000 images from the remaining slices to form the training set.

\section{Visualizations} \label{sec: appen more visual}

\begin{figure*}[htbp]
  \centering
  \begin{minipage}{\textwidth}
    \centering
    \begin{subfigure}[b]{0.15\textwidth}
      \includegraphics[width=\textwidth]{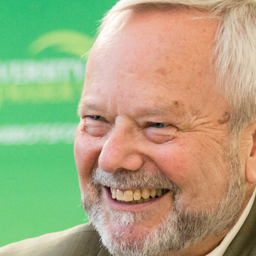}
    \end{subfigure}\hfill
    \begin{subfigure}[b]{0.15\textwidth}
      \includegraphics[width=\textwidth]{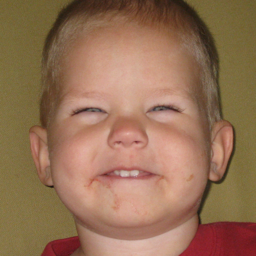}
    \end{subfigure}\hfill
    \begin{subfigure}[b]{0.15\textwidth}
      \includegraphics[width=\textwidth]{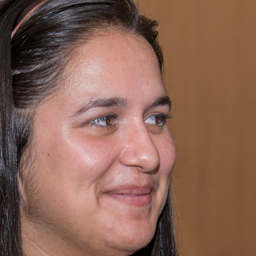}
    \end{subfigure}\hfill
    \begin{subfigure}[b]{0.15\textwidth}
      \includegraphics[width=\textwidth]{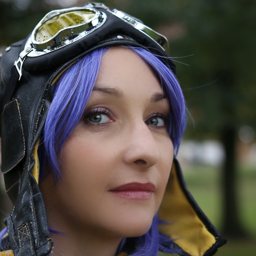}
    \end{subfigure}\hfill
    \begin{subfigure}[b]{0.15\textwidth}
      \includegraphics[width=\textwidth]{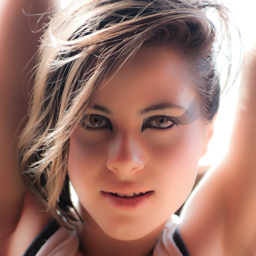}
    \end{subfigure}\hfill
    \begin{subfigure}[b]{0.15\textwidth}
      \includegraphics[width=\textwidth]{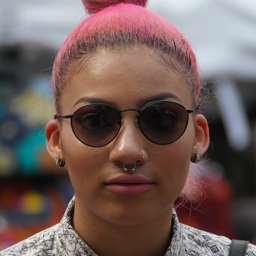}
    \end{subfigure}
  \end{minipage}

  \vspace{1em}

  \begin{tabular}{*{5}{c}}
    \includegraphics[width=0.12\textwidth]{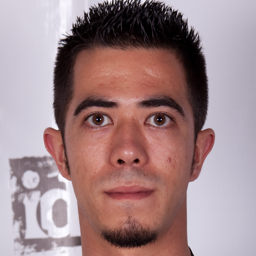}&
    \includegraphics[width=0.12\textwidth]{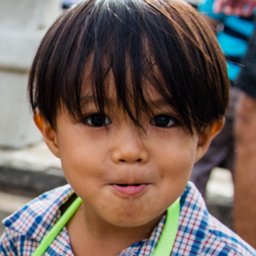}&
    \includegraphics[width=0.12\textwidth]{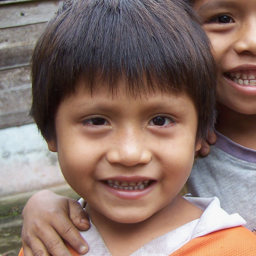}&
    \includegraphics[width=0.12\textwidth]{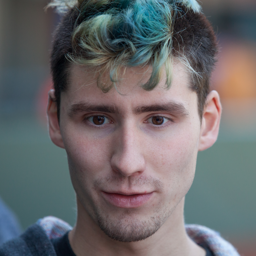}&
    \includegraphics[width=0.12\textwidth]{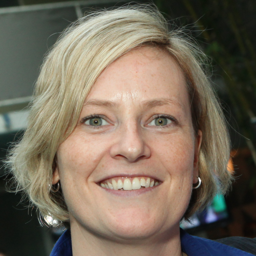} \\[-1pt]
    \includegraphics[width=0.12\textwidth]{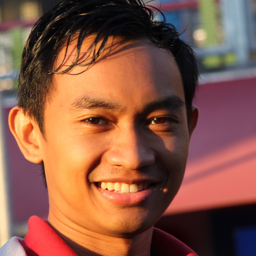}&
    \includegraphics[width=0.12\textwidth]{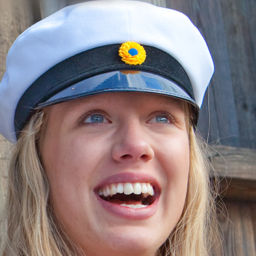}&
    \includegraphics[width=0.12\textwidth]{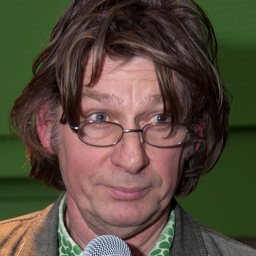}&
    \includegraphics[width=0.12\textwidth]{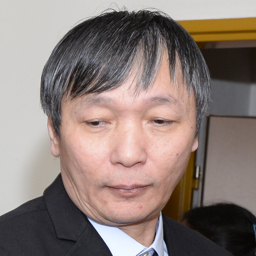}&
    \includegraphics[width=0.12\textwidth]{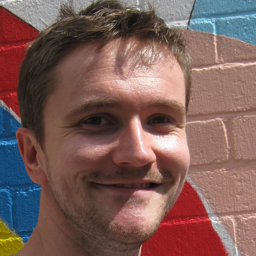}
  \end{tabular}
  \begin{tabular}{*{5}{c}}
    \includegraphics[width=0.12\textwidth]{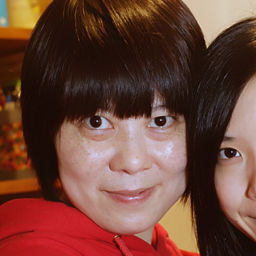}&
    \includegraphics[width=0.12\textwidth]{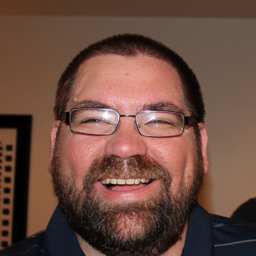}&
    \includegraphics[width=0.12\textwidth]{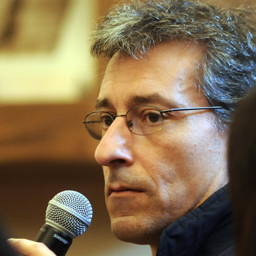}&
    \includegraphics[width=0.12\textwidth]{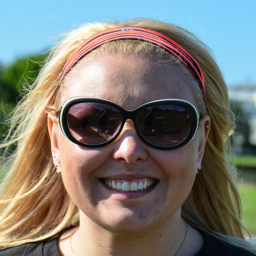}&
    \includegraphics[width=0.12\textwidth]{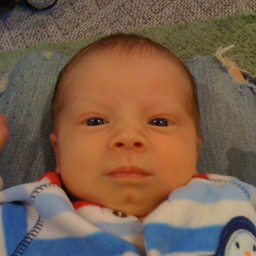} \\[-1pt]
    \includegraphics[width=0.12\textwidth]{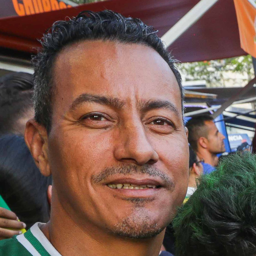}&
    \includegraphics[width=0.12\textwidth]{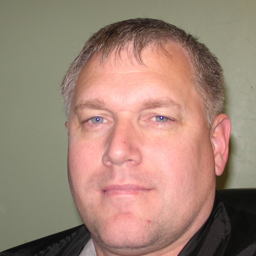}&
    \includegraphics[width=0.12\textwidth]{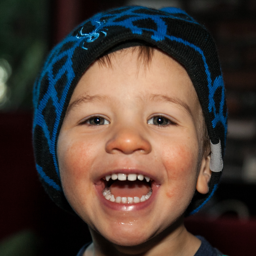}&
    \includegraphics[width=0.12\textwidth]{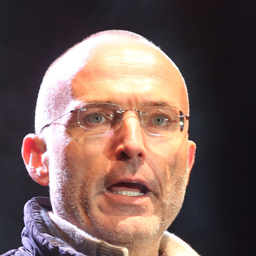}&
    \includegraphics[width=0.12\textwidth]{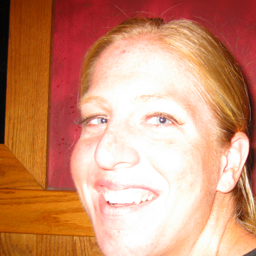}
  \end{tabular}

  \caption{%
    Top: the 6  training images. 
    Bottom: the 20 test images.
  }
  \label{fig:yourgrid}
\end{figure*}

\begin{figure*}[htbp]
\centering

\begin{tabular}{cccc}
    \textbf{Ground Truth} & 
    \textbf{Input}
    &\textbf{Self-Guided} & \textbf{DDS} \\
    
    \includegraphics[width=.22\linewidth,valign=t]{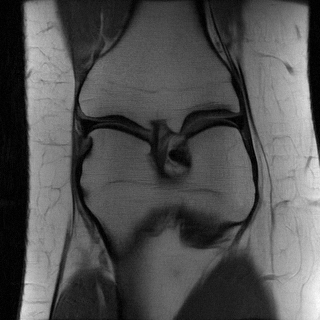} &
    \includegraphics[width=.22\linewidth,valign=t]{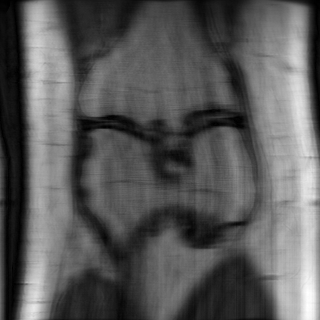} &
    \includegraphics[width=.22\linewidth,valign=t]{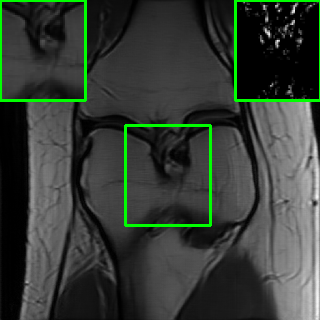} &
    \includegraphics[width=.22\linewidth,valign=t]{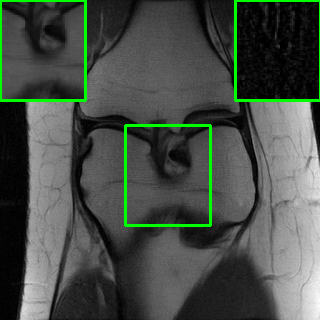} \\
    
    \scriptsize{PSNR = $\infty$ dB} &
    \scriptsize{PSNR = 21.12 dB} &\scriptsize{PSNR = 32.78 dB} & \scriptsize{PSNR = 34.55 dB} \\
    
    \textbf{SITCOM MRI} & \textbf{aSeqDIP} & \textbf{\our (Ours)} \\
    
    \includegraphics[width=.22\linewidth,valign=t]{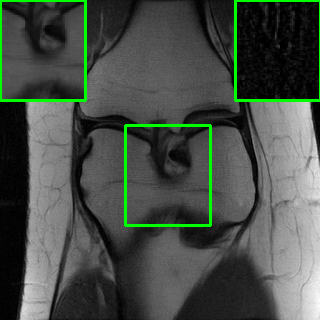} &
    \includegraphics[width=.22\linewidth,valign=t]{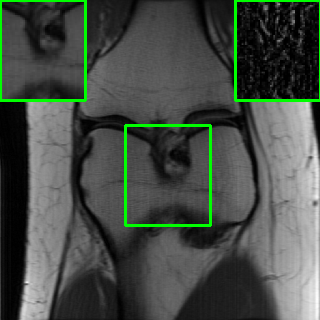} &
    \includegraphics[width=.22\linewidth,valign=t]{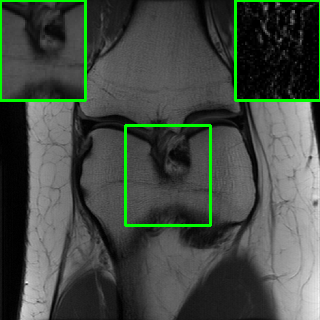} \\
    
    \scriptsize{PSNR = 35.67 dB} & \scriptsize{PSNR = 34.48 dB} & \scriptsize{PSNR = 35.56 dB} \\
\end{tabular}
\caption{\small{Visualization of ground-truth and reconstructed images using different methods of a knee image from the fastMRI dataset with 4x k-space undersampling. A region of interest is shown with a green box and its error (magnitude) is shown in the panel on the top right. }}
\label{fig:denoised_imgs_zoomed mri 1}
\end{figure*}

\begin{figure*}[htbp]
\centering

\begin{tabular}{cccc}
    \textbf{Ground Truth} & 
    \textbf{Input}
    &\textbf{Self-Guided} & \textbf{DDS} \\
    
    \includegraphics[width=.22\linewidth,valign=t]{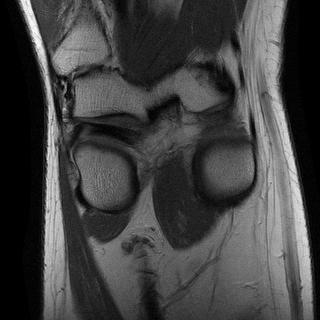} &
    \includegraphics[width=.22\linewidth,valign=t]{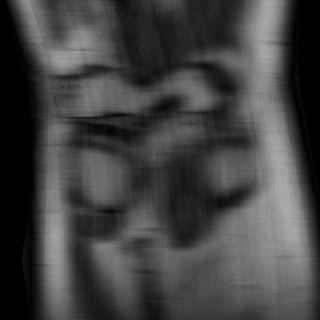} &
    \includegraphics[width=.22\linewidth,valign=t]{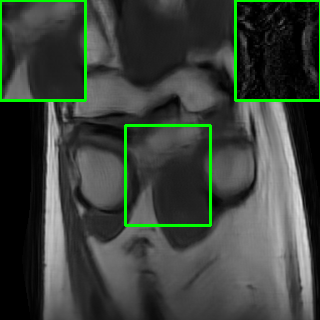} &
    \includegraphics[width=.22\linewidth,valign=t]{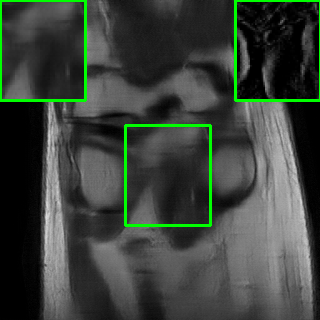} \\
    
    \scriptsize{PSNR = $\infty$ dB} &
    \scriptsize{PSNR = 16.45 dB} &\scriptsize{PSNR = 31.65 dB} & \scriptsize{PSNR = 32.15 dB} \\
    
    \textbf{SITCOM MRI} & \textbf{aSeqDIP} & \textbf{\our (Ours)} \\
    
    \includegraphics[width=.22\linewidth,valign=t]{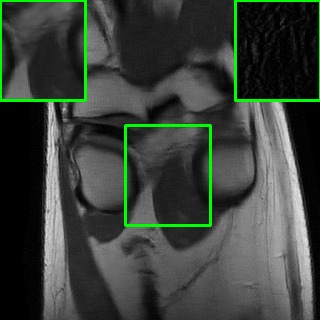} &
    \includegraphics[width=.22\linewidth,valign=t]{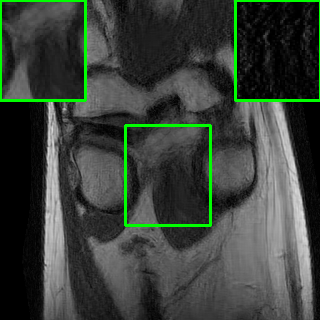} &
    \includegraphics[width=.22\linewidth,valign=t]{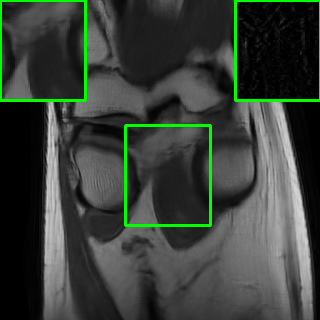} \\
    
    \scriptsize{PSNR = 32.57 dB} & \scriptsize{PSNR = 32.09 dB} & \scriptsize{PSNR = 32.66 dB} \\
\end{tabular}
\caption{\small{Visualization of ground-truth and reconstructed images using different methods of a knee image from the fastMRI dataset with 8x k-space undersampling. A region of interest is shown with a green box and its error (magnitude) is shown in the panel on the top right.}}
\label{fig:denoised_imgs_zoomed mri 2}
\end{figure*}

\begin{figure*}[htbp]
\centering
\begin{tabular}{cccc}
    \textbf{Ground Truth} &
    \textbf{Input} &
    \textbf{STRAINER} & \textbf{DPS} \\
    
    \includegraphics[width=.22\linewidth,valign=t]{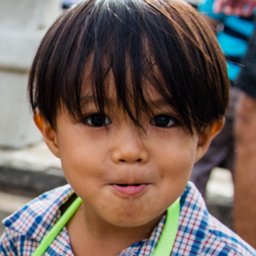} &
    \includegraphics[width=.22\linewidth,valign=t]{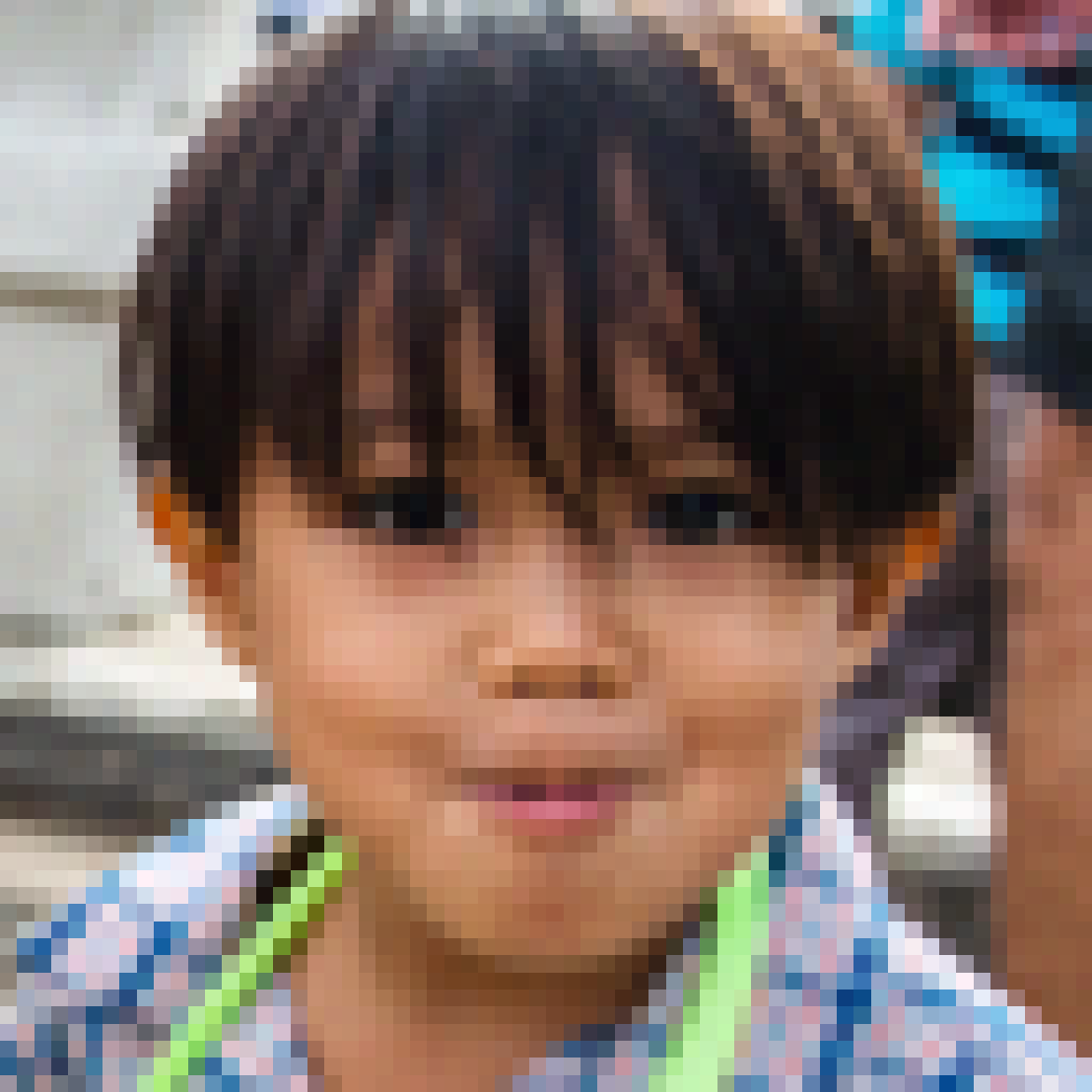} &
    \includegraphics[width=.22\linewidth,valign=t]{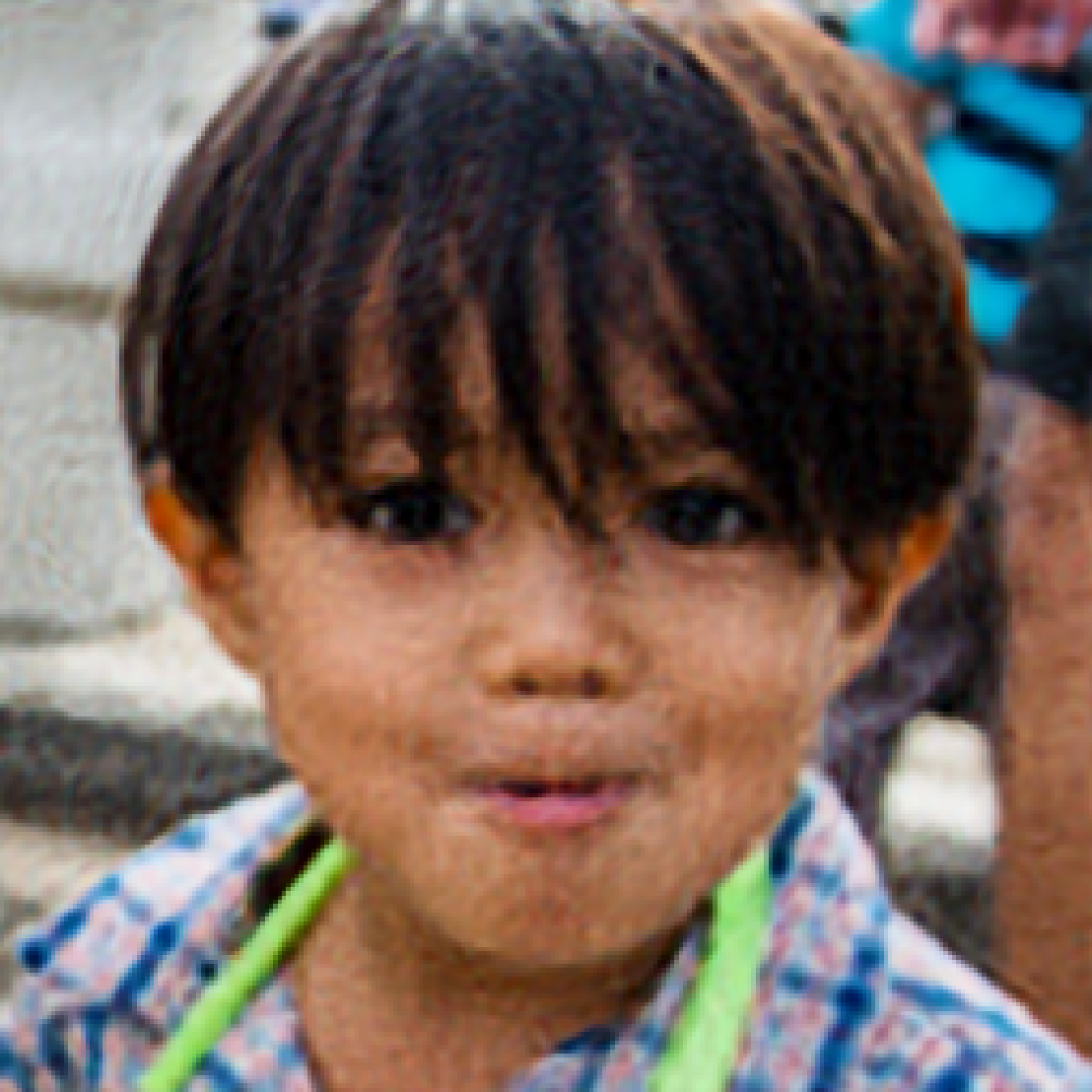} &
    \includegraphics[width=.22\linewidth,valign=t]{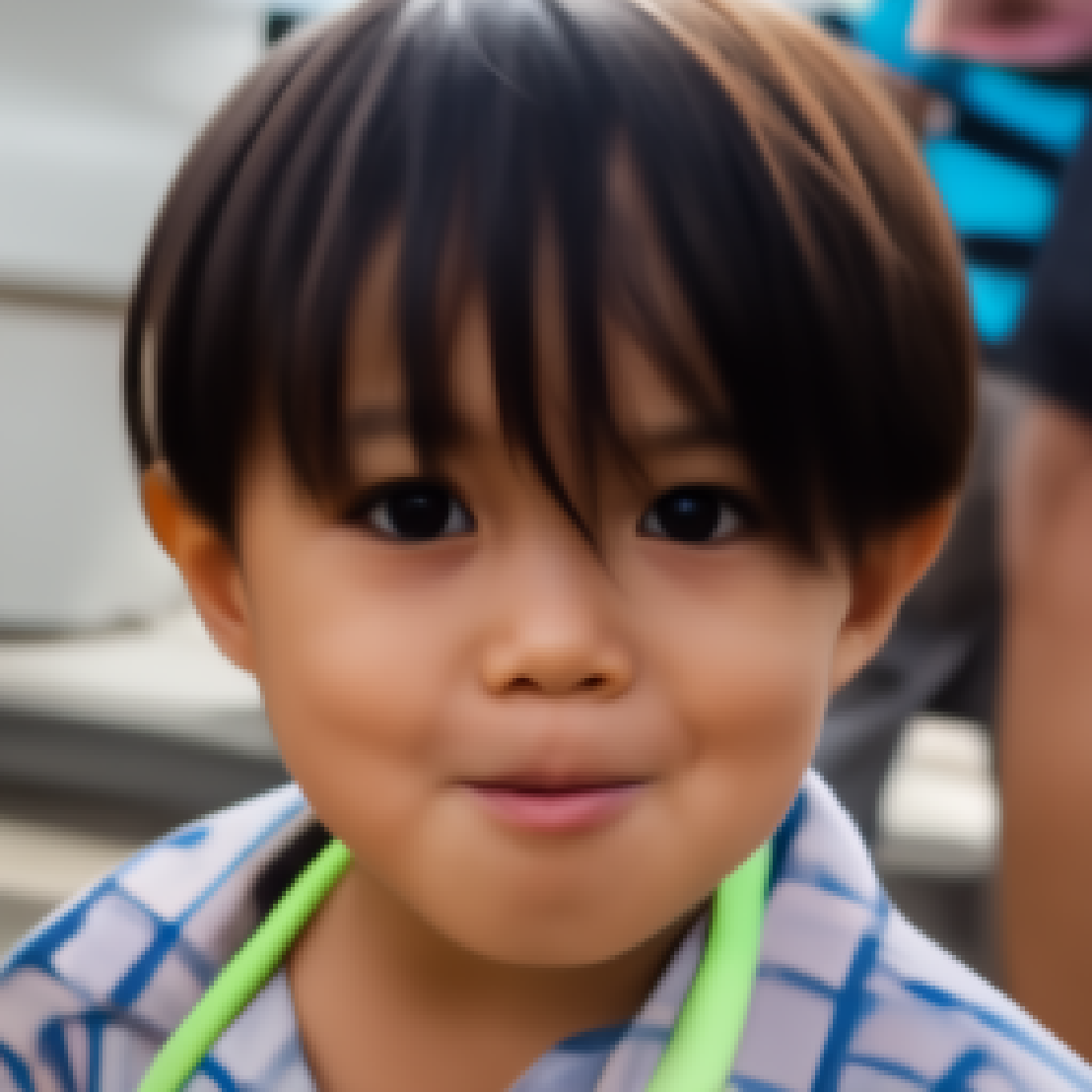} \\
    
   \scriptsize{PSNR = $\infty$ dB} &
    \scriptsize{PSNR = 22.45 dB} &\scriptsize{PSNR = 28.65 dB} & \scriptsize{PSNR = 27.35 dB} \\

    \textbf{SITCOM} & \textbf{aSeqDIP} & \textbf{\our (Ours)} \\
    
    \includegraphics[width=.22\linewidth,valign=t]{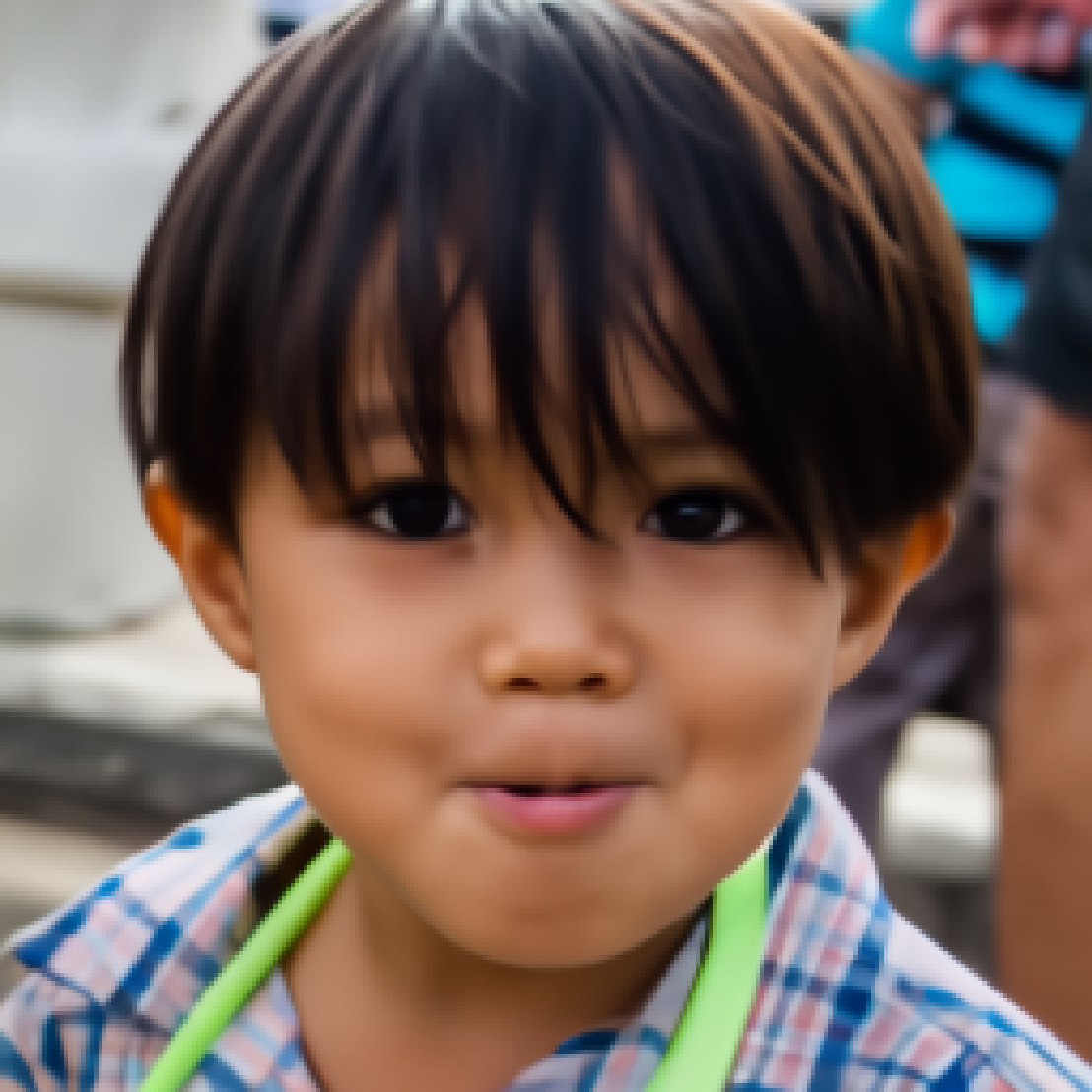} &
    \includegraphics[width=.22\linewidth,valign=t]{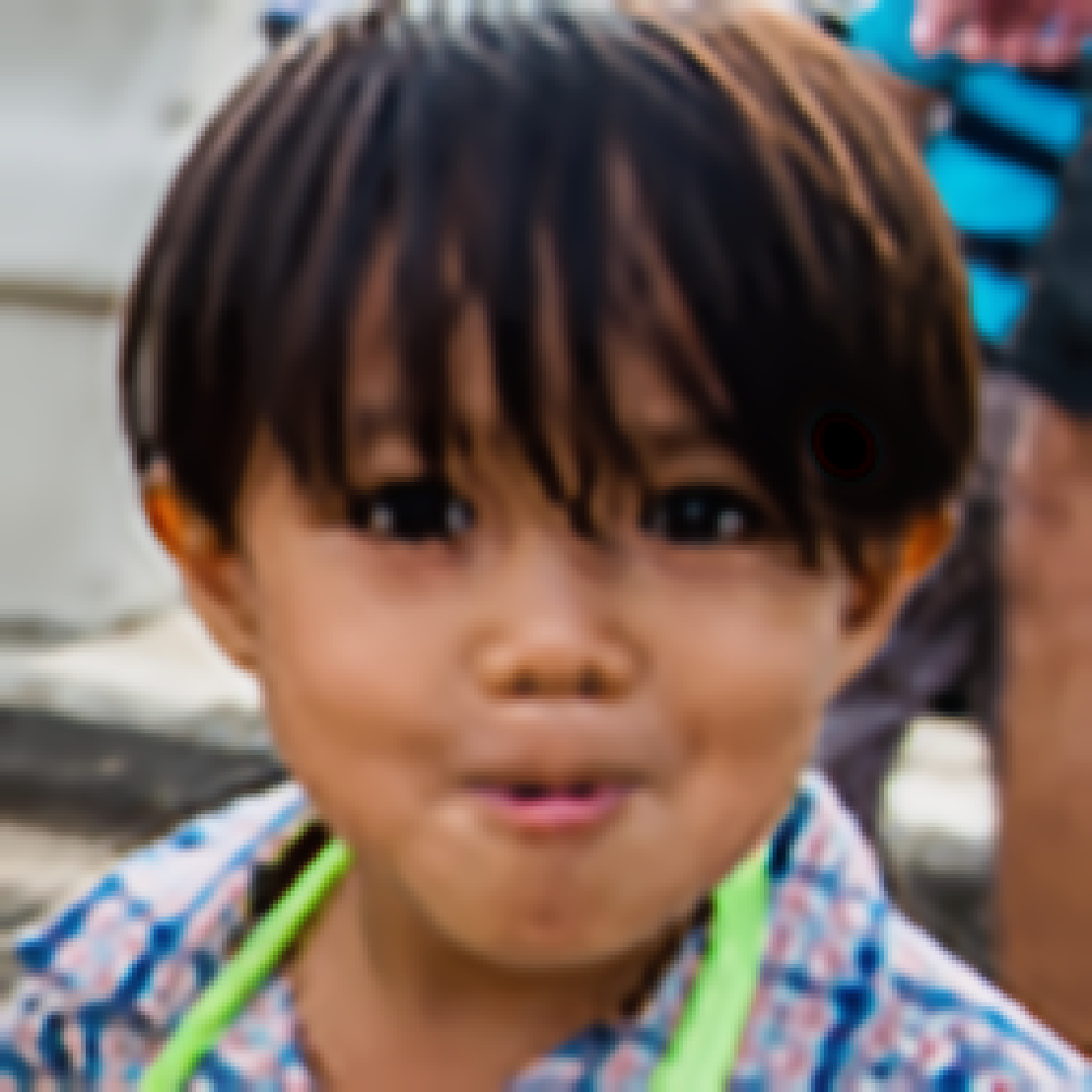} &
    \includegraphics[width=.22\linewidth,valign=t]{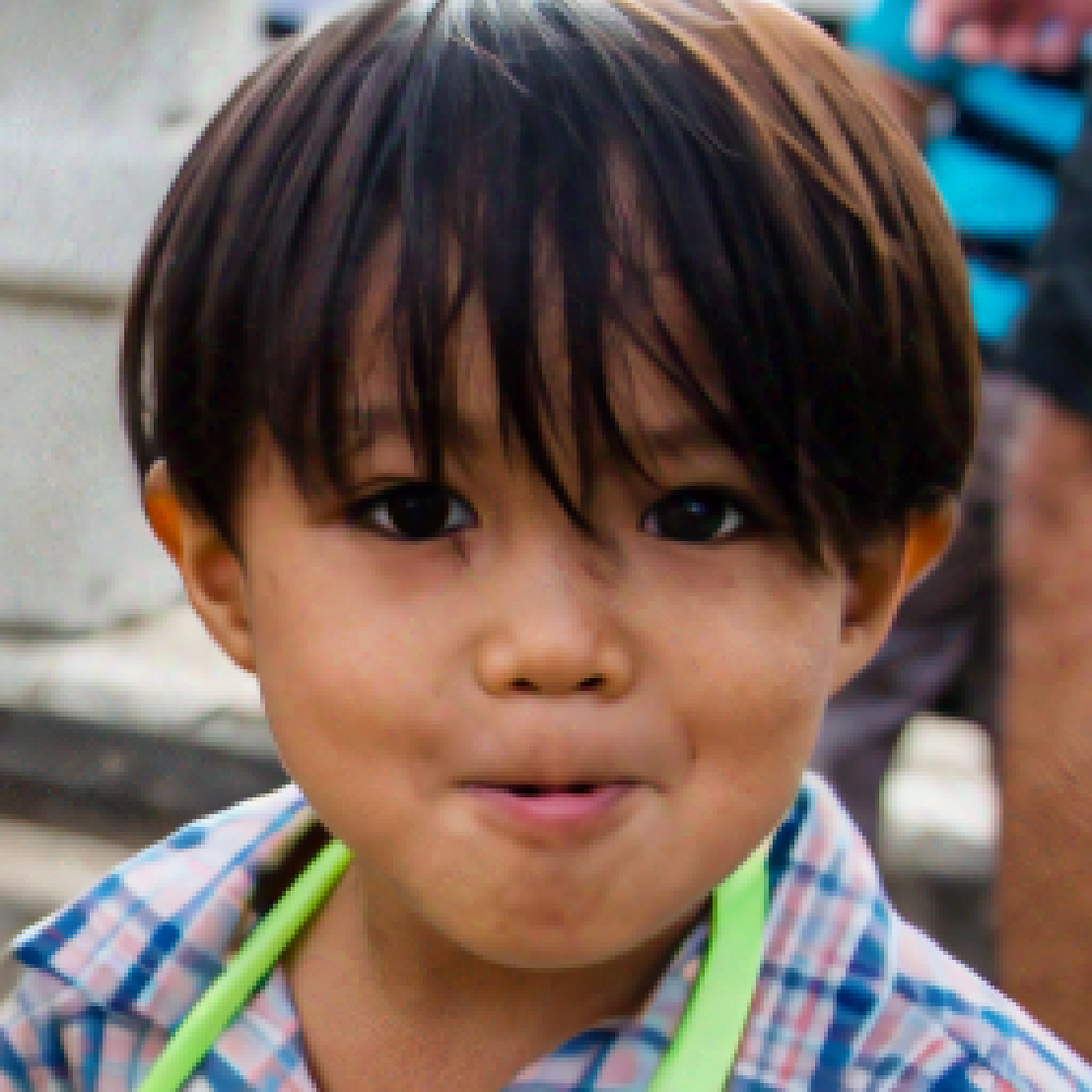} \\
    
    \scriptsize{PSNR = 31.78 dB} & \scriptsize{PSNR = 31.24 dB} & \scriptsize{PSNR = 31.96 dB} \\
\end{tabular}
\caption{\small{
Super Resolution example from the FFHQ dataset.
}}
\label{fig:SR}
\vspace{-0.1in}
\end{figure*}

\begin{figure*}[htbp]
\centering
\begin{tabular}{cccc}
    \textbf{Ground Truth} &
    \textbf{Input} &
    \textbf{STRAINER} & \textbf{DPS} \\
    
    \includegraphics[width=.22\linewidth,valign=t]{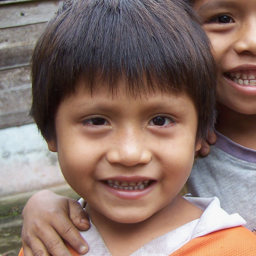} &
    \includegraphics[width=.22\linewidth,valign=t]{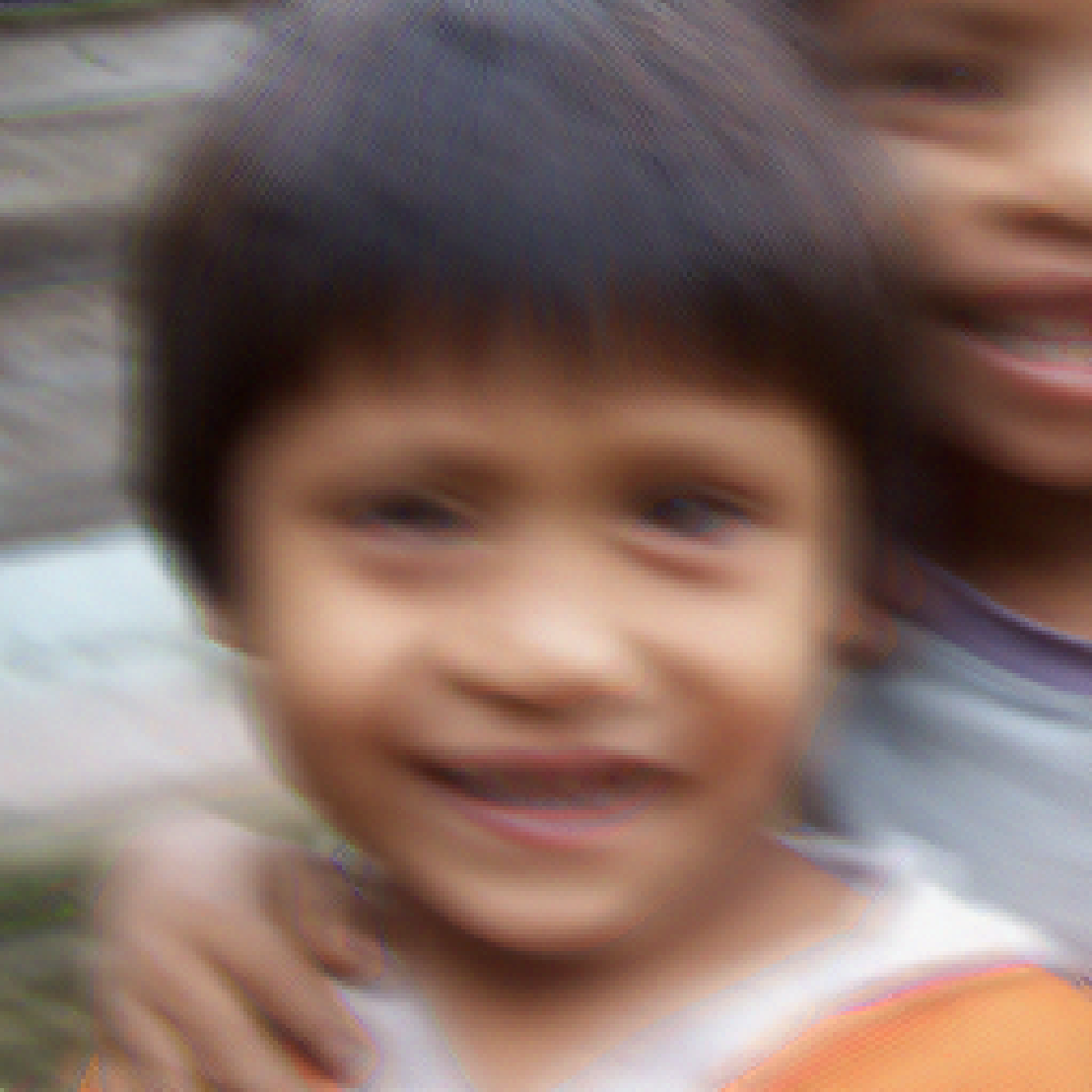} &
    \includegraphics[width=.22\linewidth,valign=t]{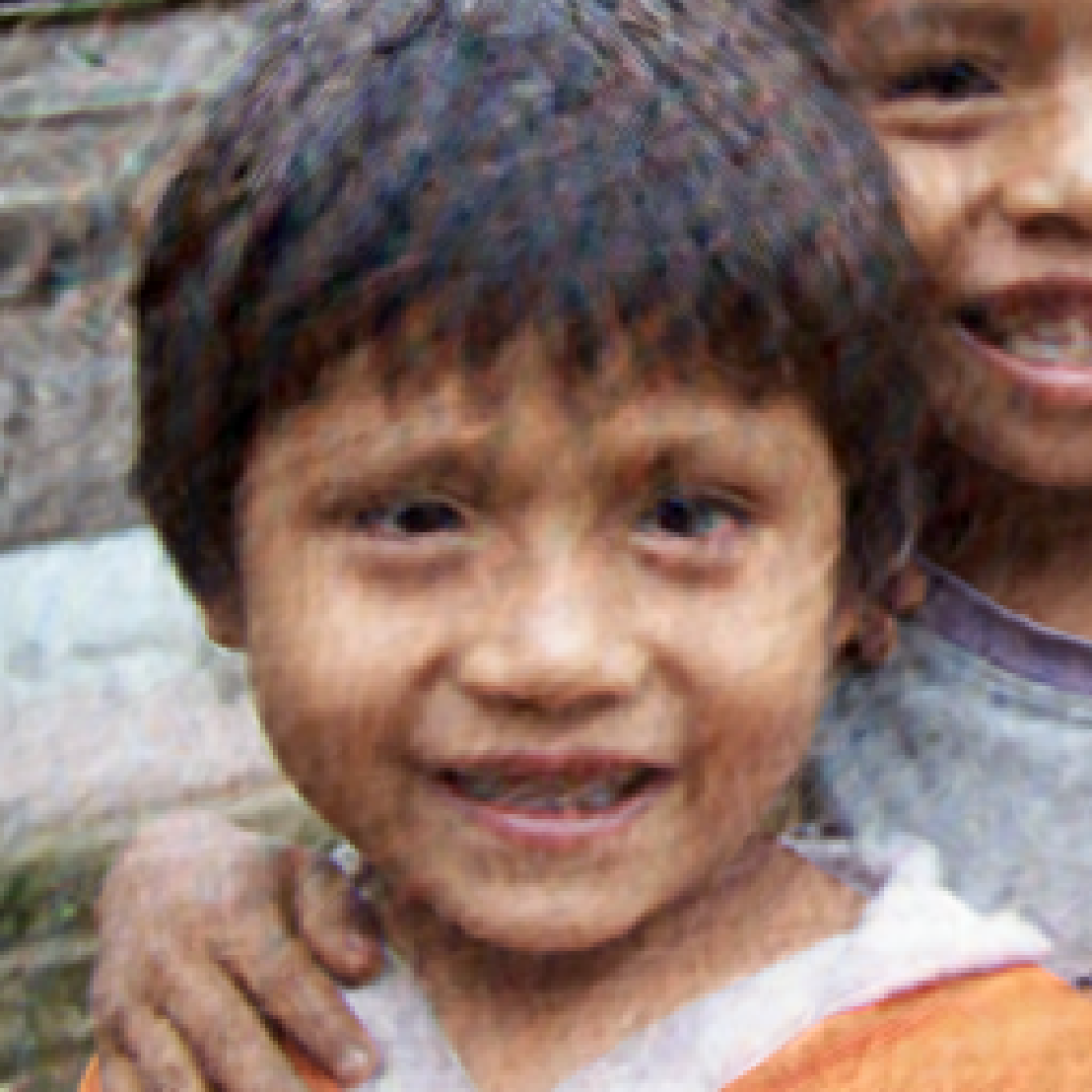} &
    \includegraphics[width=.22\linewidth,valign=t]{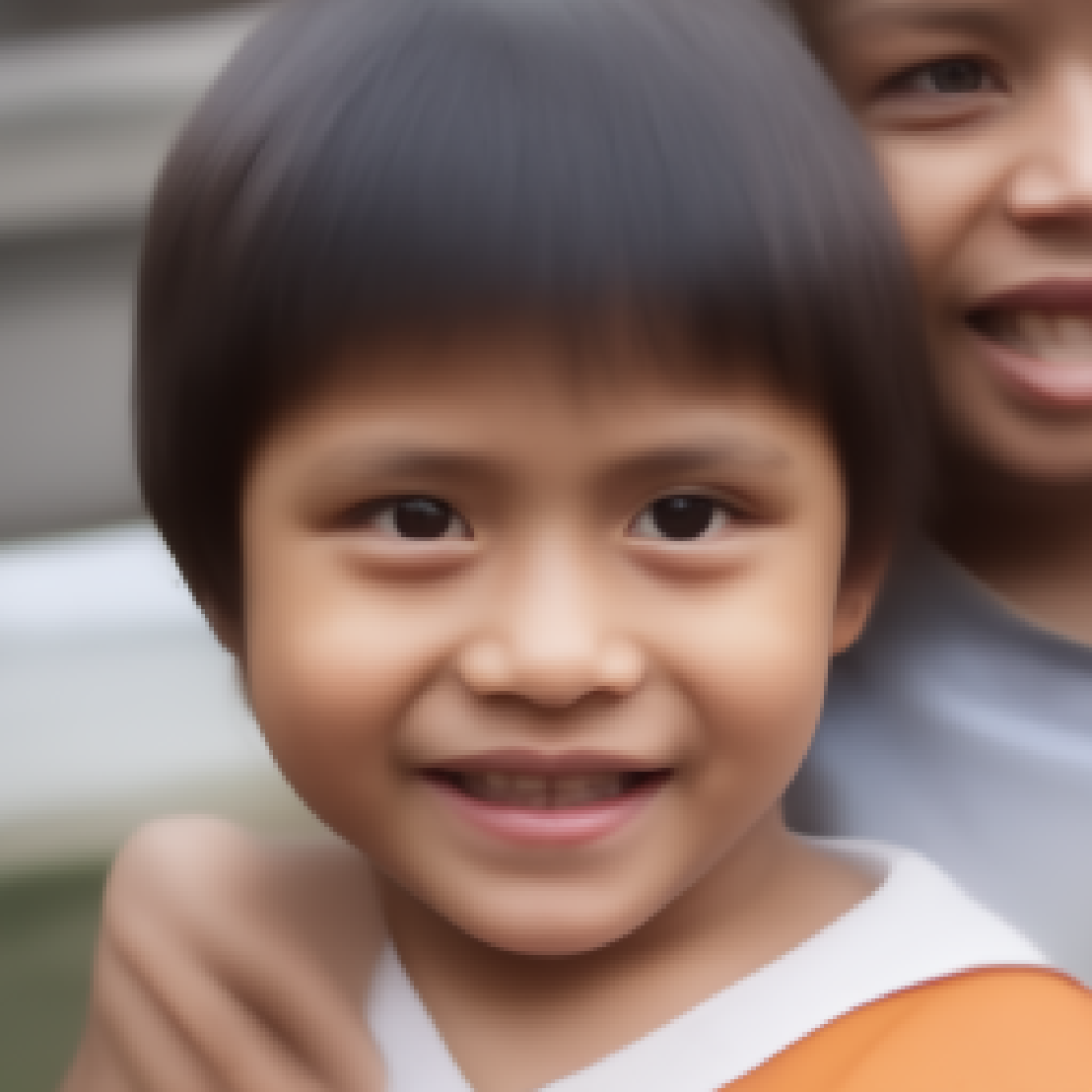} \\
    
    \scriptsize{PSNR = $\infty$ dB} &
    \scriptsize{PSNR = 23.24 dB} &\scriptsize{PSNR = 24.82 dB} & \scriptsize{PSNR = 25.52 dB} \\
    
    \textbf{SITCOM} & \textbf{aSeqDIP} & \textbf{\our (Ours)} \\
    
    \includegraphics[width=.22\linewidth,valign=t]{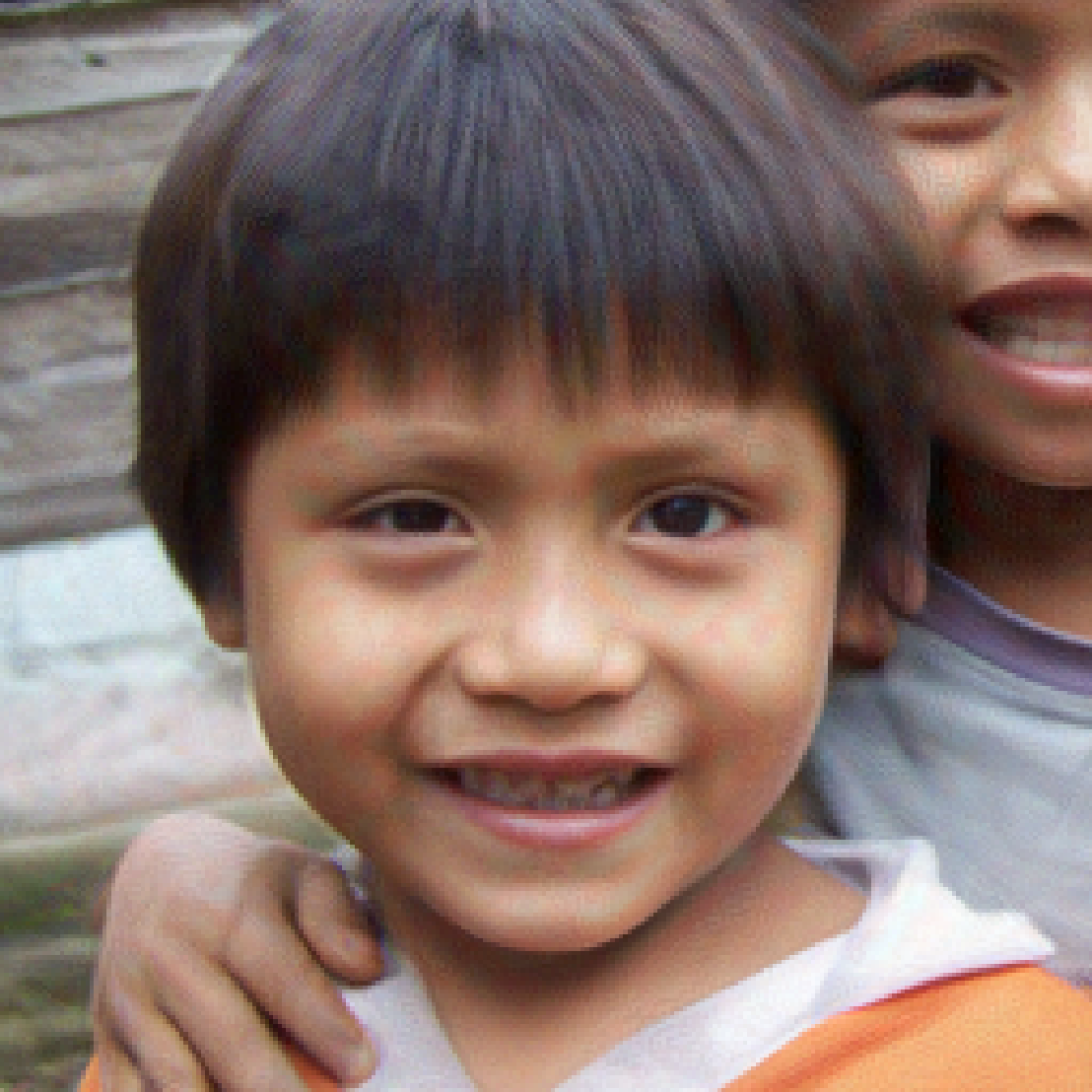} &
    \includegraphics[width=.22\linewidth,valign=t]{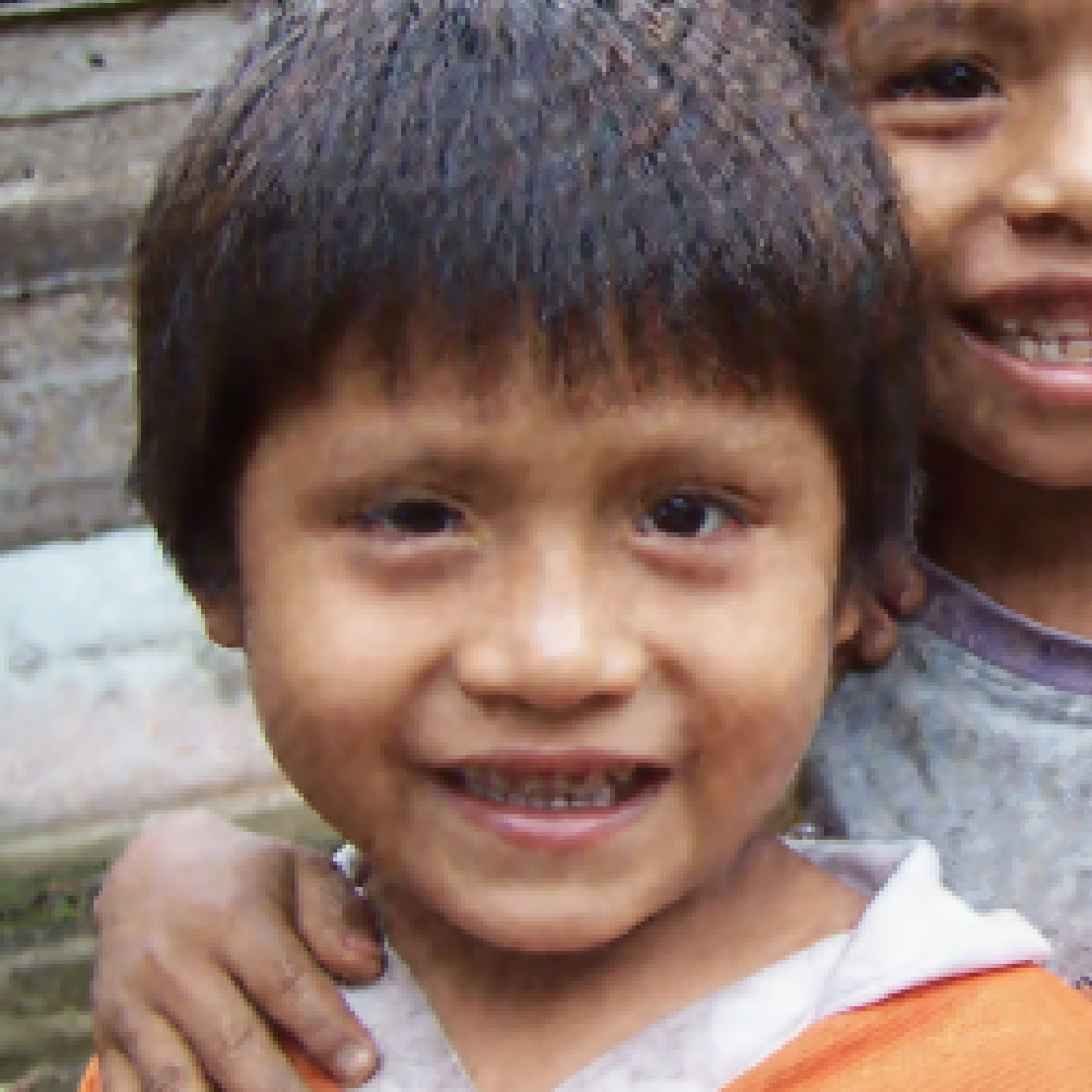} &
    \includegraphics[width=.22\linewidth,valign=t]{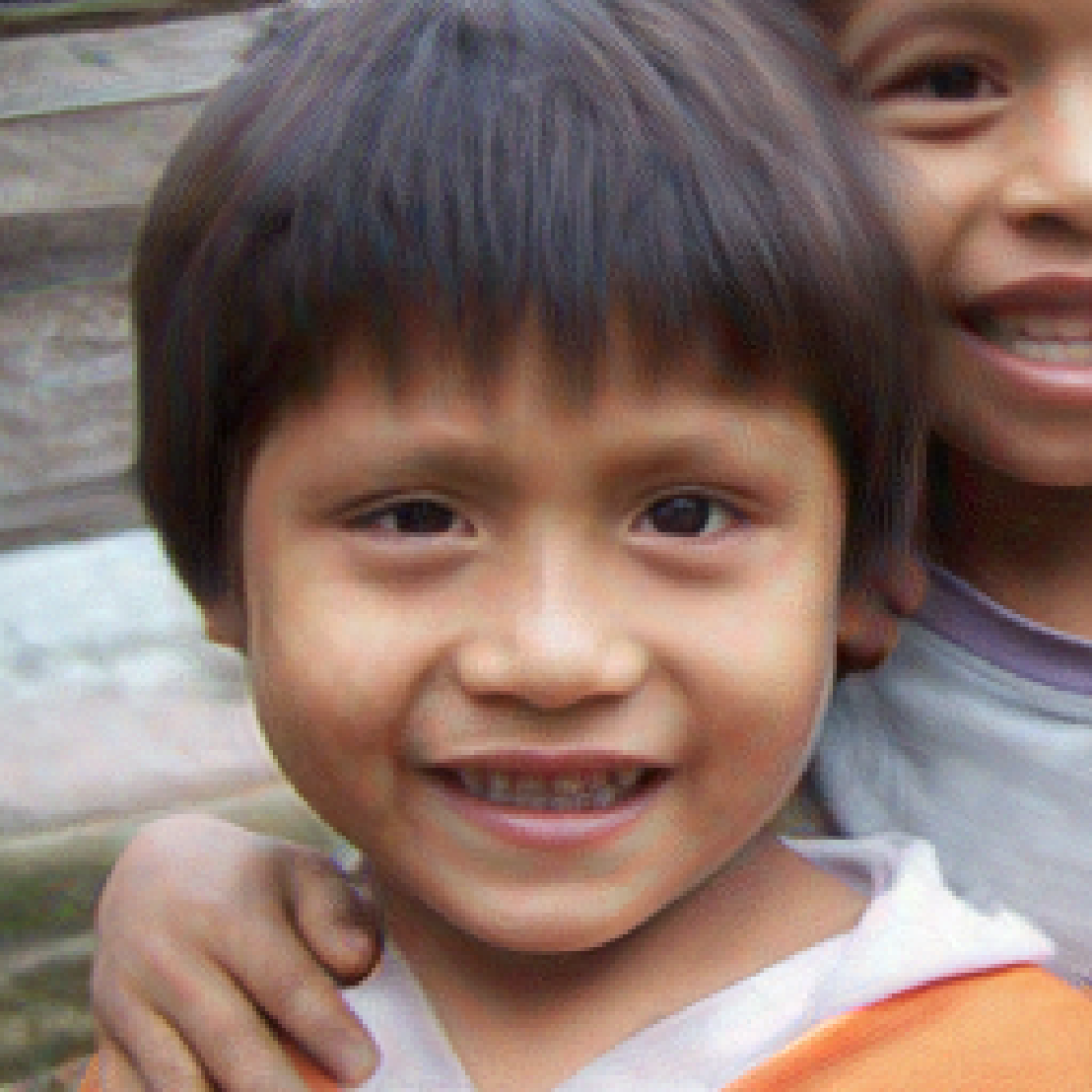} \\
    
    \scriptsize{PSNR = 30.82 dB} & \scriptsize{PSNR = 30.17 dB} & \scriptsize{PSNR = 31.06 dB} \\
\end{tabular}
\caption{\small{
Non-linear deblurring example from the FFHQ dataset. 
}}
\label{fig:DEBLURRING}
\end{figure*}

Figure~\ref{fig:yourgrid} shows the clean versions of $6$ training images (top row) and the $20$ testing images sampled from the FFHQ dataset that we used in \our. Figures~\ref{fig:denoised_imgs_zoomed mri 1} and \ref{fig:denoised_imgs_zoomed mri 2} present additional MRI visualizations. Samples from the natural image restoration tasks are given in Figure~\ref{fig:SR} and Figure~\ref{fig:DEBLURRING} super resolution and non-linear deblurring, respectively.


\end{document}